\documentclass[runningheads]{llncs}

\usepackage{eccv}
\usepackage{eccvabbrv}

\usepackage{graphicx}
\usepackage{booktabs}
\usepackage{multirow}
\usepackage{bbm}
\usepackage[font=footnotesize,skip=0pt]{caption}

\usepackage{xspace}
\def\ours{\texttt{MED-DSLC}\@\xspace}
\def\dsours{\texttt{MED-DS}\@\xspace}
\def\med{\texttt{MED}\@\xspace}

\usepackage[accsupp]{axessibility}  

\usepackage{hyperref}

\usepackage{orcidlink}

\begin{document}

\title{MED-DSLC: Multi-Expert-Domain Classification via Domain Supervision and Logit Calibration} 

\titlerunning{MED-DSLC}

\author{Zheng Zeng\inst{*1} \and
Deepak Sridhar\inst{*1} \and
Nuno Vasconcelos\inst{1}}

\authorrunning{Z. Zeng, D. Sridhar et al.}

\institute{University of California, San Diego, USA \\
\email{\{zhz396,desridha\}@ucsd.edu}\\
\textsuperscript{*}Equal contribution
}

\maketitle

\begin{abstract}
Vision-language models (VLMs) such as CLIP enable zero-shot classification by comparing image features with text prompts in a shared embedding space. A fundamental property underlying this capability is the global comparability of logits across arbitrary candidate classes. However, VLMs are often adapted to fine-grained domains using techniques such as LoRA. While this improves in-domain accuracy, out-of-domain accuracy degrades. This leads to a higly fragmented model ecosystem, with thousands of specialized models. {\it Multi-Expert-Domain} (\med{}) classification seeks to address this problem, by merging LoRAs trained independently on specialized domains.  However, due to the independent training, the various domain experts no longer produce globally calibrated logits. As a result, when evaluating over the union of multiple domain-specific class sets, heterogeneous logit scales induce cross-domain interference and artificially high confidence for out-of-domain classes, inducing prediction errors.
In this work, we identify domain supervision and cross-domain logit miscalibration as the key issue to scalable multi-domain zero-shot recognition. We propose a mixture-of-experts \med{} architecture, \ours{}, combining  domain supervised training and domain-wise logit scaling, to explicitly restore global logit comparability. 
\ours{} is a lightweight solution for \med classification, which is shown to preserve within-domain discrimination while reducing cross-domain logit interference with minimal data. Extensive experiments across diverse fine-grained benchmarks demonstrate that it substantially improves mean accuracy (+15\%), cross-domain robustness, and scalability in the size of \med{} classification problem. Our results show that restoring output-level calibration is essential under highly data imbalanced settings for achieving a truly zero-shot VLM under multi-domain specialization. Code is publicly available at \href{https://github.com/Leonard-Zeng/MED-DSLC}{MED-DSLC}.

\end{abstract}

\section{Introduction}
\label{sec:intro}

\begin{figure*}[t]
\centering

\begin{minipage}{0.56\linewidth}
    \centering
    \includegraphics[width=\linewidth, trim=40 130 40 110, clip]{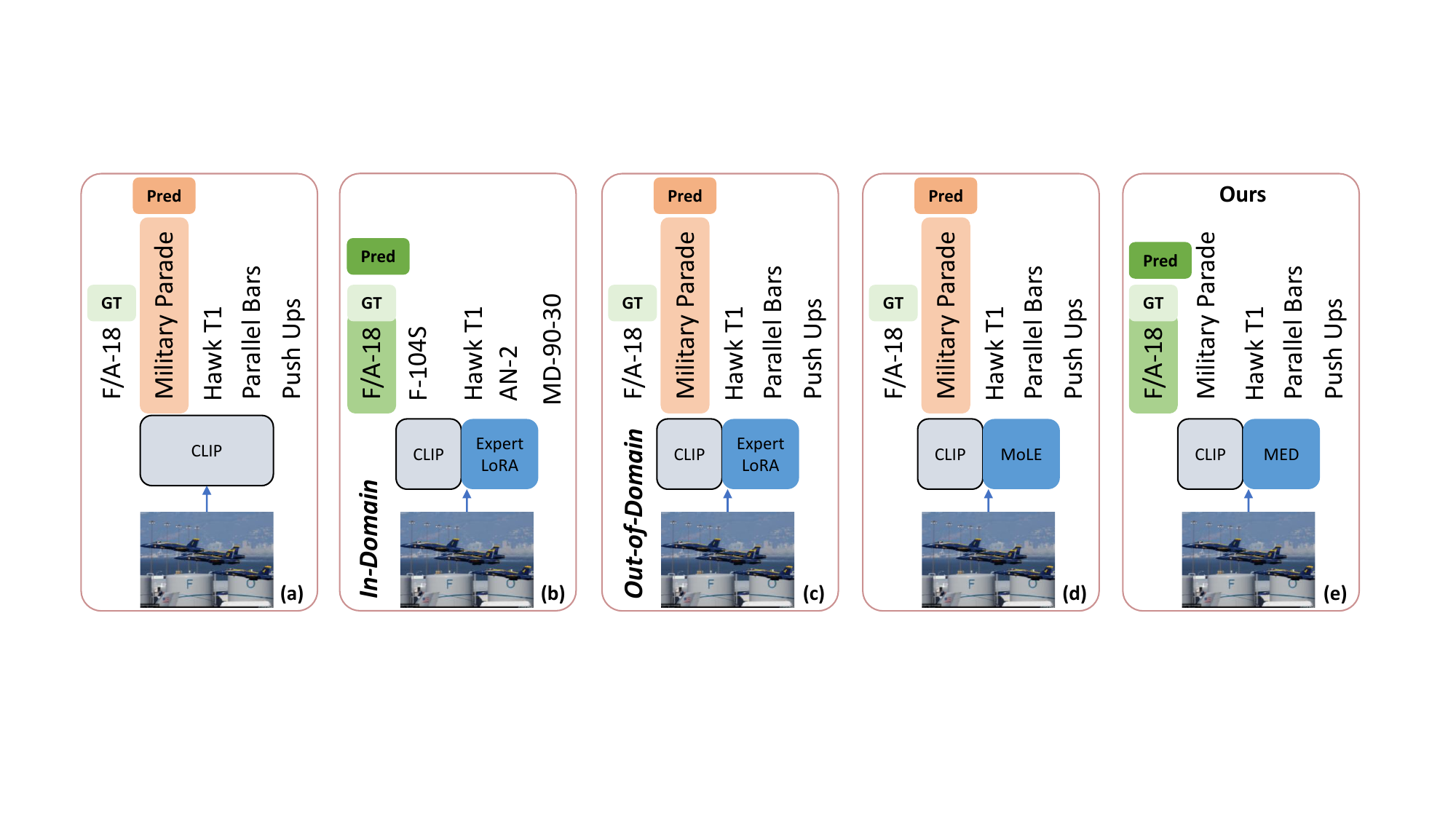}
    \caption{Logits for classification of an image of the class `FA-18'. Only largest logit value is shaded. (a) CLIP logits. (b) Logits of LoRA adapted domain expert trained on the image domain. (c) Logits of an expert trained on another domain. (d) Logit miscalibration across domains of an existing \med{} model (MoLE). (e) Logits of the proposed \ours{} classifier.}
    \label{fig:archteaser}
\end{minipage}
\hfill
\begin{minipage}{0.42\linewidth}
    \centering
    \includegraphics[width=\linewidth, trim=40 70 40 60, clip]{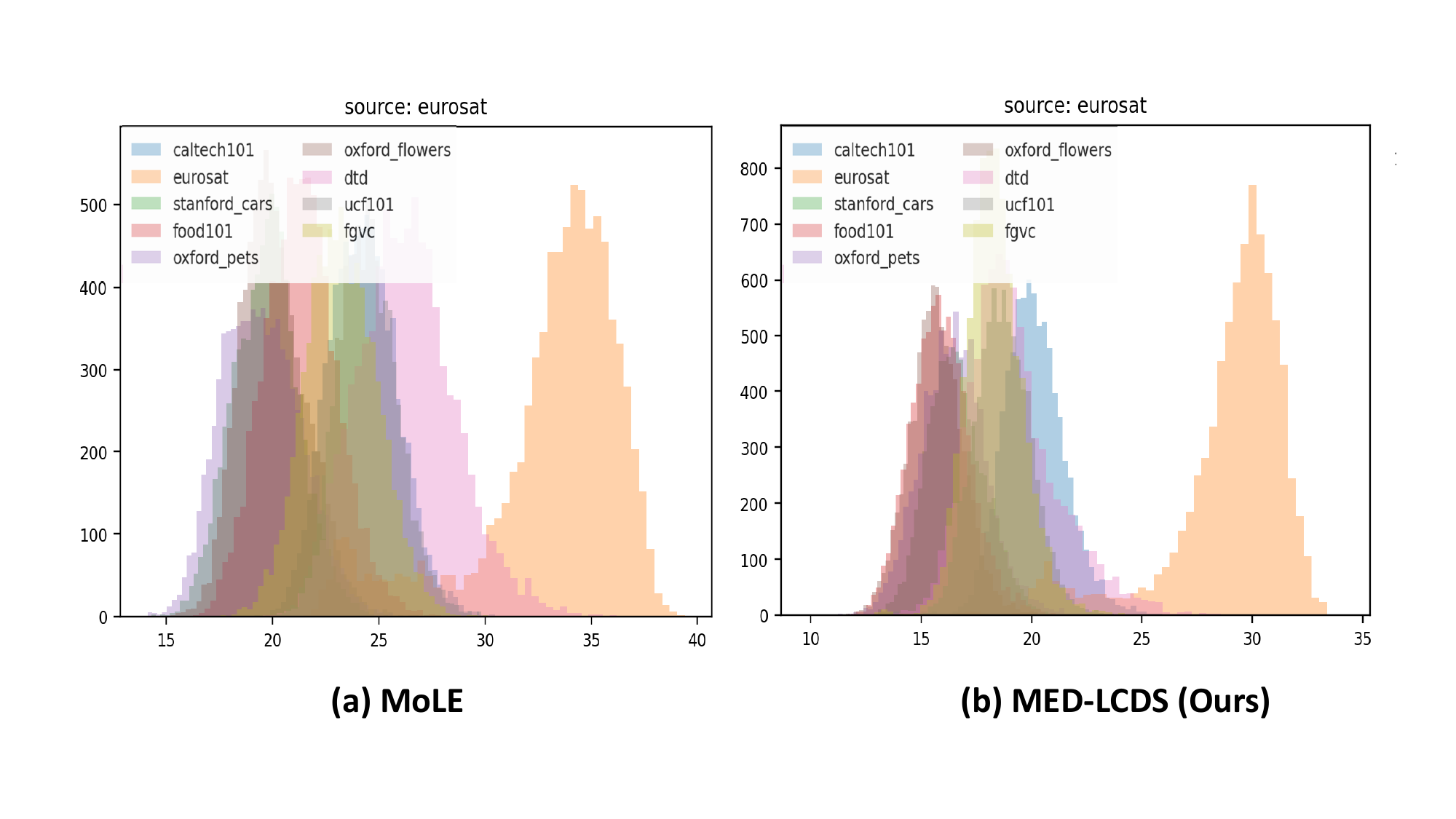}
    \caption{Logit histograms for images of the EuroSAT dataset and classes of a \med{} classifier with 10 domain experts (identified by color). (a) Logits produced by MoLE exhibit overlap between domains (EuroSAT and DTD). (b) The overlap is significantly reduced by \ours.}
    \label{fig:histteaser}
\end{minipage}

\end{figure*}

Vision-language models (VLMs), such as CLIP~\cite{radford2021learning}, have transformed visual recognition by enabling zero-shot classification with open class vocabularies. By aligning images and text in a shared embedding space, these models can classify images with respect to arbitrary class names without retraining. Given feature vectors for image and class labels, a similarity score (logit) is computed per label and a softmax function used to output class probabilities. A central property underlying the zero shot capability is that the logits of all classes are globally comparable, allowing softmax normalization over any set of labels. Despite this strong zero-shot generalization, performance often degrades on fine-grained or domain-specific datasets involving image classes not commonly found in the public domain. This is illustrated in Fig.~\ref{fig:archteaser}(a), which visualizes the logits (only largest one shaded) of the CLIP model for the image shown and multiple-domain class labels. While CLIP maintains a single globally calibrated logit space, it performs poorly due to lack of domain knowledge. In this case, it fails to classify the image into the fine-grained class of `F-18 jets', instead assigning it to the coarser and only loosely related class of `Military Parade'. 

To address this limitation, parameter-efficient finetuning (PEFT) methods, such as prompt tuning~\cite{zhou2022coop} and low-rank adaptation (LoRA)~\cite{hu2022lora}, are widely used to specialize VLMs to individual domains. In practice, however, this results in a collection of independently adapted domain experts, each achieving strong in-domain accuracy but typically exhibiting weak performance outside that domain. This is illustrated In Fig.~\ref{fig:archteaser}(b-c), where domain-experts achieve strong in-domain performance but poor out-of-domain generalization. While in Figure~\ref{fig:archteaser} (b) a domain-expert LoRA, trained on an airplane dataset, assigns the image to the correct airplane class, in Figure~\ref{fig:archteaser} (c) an out-of domain expert (LoRA trained on another dataset) produces an even larger logit for the original CLIP prediction. This creates an extremely fragmented model ecosystem where, for each VLM, many specialized adapters must be maintained. Furthermore, for many applications, the appropriate domain is typically unknown at test time or there is a need to combine classes from many domains. In this case, it is not even clear which domain expert to use.

This problem has created interest in the multi-expert-domain (\med{}) task, where the goal is to merge adapter parameters or their outputs~\cite{wu2024mixture,feng2024mixtureofloras,gou2023mocle} to consolidate several domain-specific experts into a single unified model. While merging alleviates model fragmentation, it is not trivial to guarantee that the \med{} model preserves the performance of the individual experts across domains. There are two major difficulties. The first is the routing problem, namely to decide which expert to use for each image at test time. A recent trend in the literature~\cite{wu2024mixture,muqeeth2024learning}, which we follow, is to rely on a mixture of experts (MoE) architecture~\cite{shazeer2017outrageously}, where a gating network selects the expert LoRA. However, these approaches follow a literal MoE implementation, where domain assignments are learned without supervision. Hence, they inherit the MoE problem of expert load balancing~\cite{fedus2021switch, lepikhin2020gshard}, which is not trivial to solve. We note that, for \med{} classifiers, domain labels can be easily derived from class labels, which are always available for training. We thus propose an alternative MoE training scheme, based on supervised domain learning.

A second problem is that independently adapted experts implicitly reshape both the geometry and magnitude of their output logits. When candidate labels from multiple domains are evaluated jointly, either through naïve merging or mixture-based aggregation, logits originating from different domain specialists directly compete under softmax normalization. Because these logits are not calibrated across domains, systematic scale discrepancies arise. Domains producing higher-magnitude logits can dominate predictions even for out-of-domain inputs, leading to artificial overconfidence and degraded cross-domain accuracy. This is shown in Fig.~\ref{fig:archteaser}(d), for a model (MoLE) that combines logits from different experts, including those of Fig.~\ref{fig:archteaser}(b)-(c), by mixture-based aggregation~\cite{wu2024mixture}. While each expert has independently reshaped its logit geometry and scale, the expert logits  compete directly under a softmax that spans all classes. Since the expert logits are no longer comparable, this results in cross-domain interference: domains producing larger-magnitude logits dominate predictions, even for out-of-domain inputs. Fig.~\ref{fig:histteaser} (a) illustrates the problem by showing the histogram of logits produced for the various classes of a \med{} problem by the MoE classifier, for images of the EuroSAT dataset. Logit miscalibration produces significant overlap between logits of EuroSAT classes and those of other domains. Importantly, this phenomenon intensifies as the class imbalance or the number of specialized domains increases, fundamentally undermining scalability.

While domain supervision mitigates this problem, there can still be interference from out-of-domain experts that produce unusually high logits. To overcome this problem, we augment domain supervision with a simple and efficient domain-wise logit scaling. This consists of learning a temperature parameter per domain to normalize its logits before joint softmax evaluation. This domain-wise logit scaling preserves within-domain discrimination while aligning cross-domain magnitudes, preventing any single domain from dominating predictions. We denote the combination of domain supervised routing and logit calibration {\it \med{} by Domain Supervision and Logit Calibration} (\ours). As shown in Fig.~\ref{fig:archteaser}(e), this combination restores cross-domain logit alignment, enabling the correct predictions. Fig.~\ref{fig:histteaser} (b) shows that \ours{} learns to better separate the logits of classes from different domains. 
Extensive experiments across diverse fine-grained recognition benchmarks demonstrate that this combination substantially improves cross-domain robustness and stabilizes performance as the number of domains grows. Compared to existing \med{} strategies, our approach achieves higher mean accuracy over union label spaces and significantly reduces overconfidence on out-of-domain classes.

Overall this paper makes the following contributions :
\begin{enumerate}
    \item We identify unsupervised domain routing and cross-domain logit miscalibration as primary barriers to the \med{} problem.
    \item We propose \ours{}, a combination of domain-wise logit calibration and  domain-aware expert routing to restore global logit comparability.
    \item We demonstrate consistent improvements in cross-domain and union-class evaluation, enabling robust truly zero-shot recognition across specialized domains.
\end{enumerate}

\section{Related Work}

\noindent{\bf Vision--Language Pretraining and Parameter-Efficient Finetuning.}  
Contrastive trained VLMs, such as CLIP~\cite{radford2021learning}, have become a foundation for open-vocabulary recognition. These models learn aligned image-text embeddings from large-scale web data, enabling zero-shot classification across diverse concepts. However, they often underperform on fine-grained or domain-specific benchmarks, motivating parameter-efficient finetuning (PEFT) strategies such as prompt learning ~\cite{zhou2022coop,zhou2022cocoop,jia2022vpt} and low-rank adaptation (LoRA)~\cite{hu2022lora, zanella2024low}. LoRA adds low-rank updates to frozen weights, enabling efficient specialization to fine-grained domains. However, these approaches typically produce one set of adapted parameters per domain, resulting in a collection of specialized experts rather than a unified model capable of handling multiple domains jointly. This is the starting point for the \med{} classification problem , which seeks to produce that model.

\noindent{\bf LoRA Composition and Mixture-of-Experts.} 
Recent works explore combining multiple LoRA adapters through dynamic composition~\cite{huang2023lorahub} or mixture-of-experts (MoE)~\cite{jacobs1991adaptive,jordan1993hierarchical,shazeer2017outrageously} mechanisms. 
LoraHub~\cite{huang2023lorahub} proposes dynamic LoRA composition to improve cross-task generalization. Model-based Clustering~\cite{pmlr-v235-ostapenko24a} proposes maintaining a library of LoRA modules that can be composed to extend model capabilities through dynamic adapter routing. Several works investigate merging independently trained LoRAs into a single model: while \cite{prabhakar2025lorasoups} extends model soups to LoRA via weight averaging, ~\cite{zhao2025loralego} studies rank-wise clustering to enhance modularity, and~\cite{shah2023ZipLoRA} demonstrates effective composition of subject and style LoRAs through structured merging.
KnOTS~\cite{stoica2025knots} method uses singular value decomposition to map LoRA updates into a shared representation space, improving merging quality and generality without using any task data. Other works~\cite{panariello2025core_space,apgd2024merging,zheng2025do_merging,alipour2025reversible_merging} explore similar themes, such as data-free merging LoRA modules in a common low-rank core space that preserves efficient adaptation while improving merge accuracy. While these approaches highlight the flexibility of low-rank updates, they primarily focus on skill or style composition, they typically target language tasks and do not explicitly address cross-domain logit calibration, which is central to the \med{} problem.

MoE methods such as Mixture-of-LoRA-Experts (MoLE)~\cite{wu2024mixture} introduces load-balanced expert routing over LoRA modules while
\cite{muqeeth2024learning} studies routing strategies with pretrained LoRAs and gating vectors to improve unseen-domain transfer. 
These methods typically rely on unsupervised gating or learned routing without explicitly leveraging domain priors or availability of specialized fine-tuned LoRAs. They do not explicitly address domain routing or cross-domain interference when multiple domain-specialized experts are jointly evaluated. These problems are crucial to \med{} and the focus of our work.

\section{Method}
\label{sec:method}

In multi-expert-domain (\med{}) classification, a VLM has been adapted to multiple expert domains, using PEFT techniques like LoRA, and must perform classification on a label set drawn from the union of all classes, including novel class compositions that span several domains, without domain knowledge at test-time. The \med{} classifier should leverage the VLM and a set of adaptation parameters (e.g. LoRAs) independently trained on the different domains. Ideally, both of these should be kept frozen during \med{} training. The independent domain adaptation induces cross-domain logit miscalibration, which compromises global zero-shot consistency. To address this problem, we propose \ours{}, a \med classifier with domain-supervised routing and domain-aware logit scaling.

\subsection{Domain adapted foundation models}

In the era of foundation models and open set classification, image classification is usually done  by a VLM, such as CLIP~\cite{radford2021learning}, composed by a pair $v,g$ of vision $v(x)$ and text $g(t)$ encoders, trained to map pairs $(x,t)$ of image $x$ and text $t$ into a semantic embedding $(v(x), g(t))$ where the two modalities are aligned. Given an image $x$ and a label set $\cal Y$, the VLM produces a similarity logit
\begin{equation}
z_c(x) = \langle f(x), g(t_c) \rangle.
\end{equation}
for each class name $t_c \in {\cal Y}$. A posterior class distribution is then evaluated with a softmax 
\begin{equation}
p(c \mid x) = \frac{\exp(z_c(x))}{\sum_{c' \in \mathcal{Y}} \exp(z_{c'}(x))},
\label{eq:softmax}
\end{equation}
and the class of largest probability assigned to $x$.

While this allows zero-shot classification for any label set $\cal Y$, the VLM performance is not uniform over label sets. While classification accuracies tend to be high for coarser-grained problems involving relatively non-similar classes containing images popular on the internet, performance can degrade substantially when $\cal Y$ includes fine-gained classes, or classes from image domains for which data is not widely available in the public domain. Figure~\ref{fig:archteaser} (a) gives the example of fine-grained classification of airplanes. We refer to these as {\it expert domains}\footnote{Here, ``expert'' refers to domains that challenge CLIP, not necessarily humans; e.g., fine-grained face recognition.} in this work. To overcome this problem, several PEFT techniques~\cite{zhou2022coop,zhou2022cocoop,hu2022lora,jia2022vpt} have been developed. These augment the VLM with a small set of {\it domain-specific} parameters $W_d$, and train them with expert domain data ${\cal D}_d$, while keeping $f$ and $g$ frozen, usually in the low-shot regime.
Given expert domains datasets $\{\mathcal{D}_d\}_{d=1}^D$ and associated label sets $\{\mathcal{Y}_d\}_{d=1}^D$, the VLM is adapted independently to each domain, producing $D$ specialized encoders $\{f_d = f \oplus W_d\}_{d=1}^D$. While these techniques are very effective, the adapted models tend to loose the open set robustness of the VLM.
In general,  $f_d$ achieves substantially improved performance for labels in ${\cal Y}^d$ but degrades classification outside of this set, as illustrated in Figures~\ref{fig:archteaser} (b)-(c). This results in a highly fragmented model ecosystem, where practitioners develop thousands of individual models for specific domains. 

\subsection{Multi-expert-domain classification}

In this work, we consider the problem of how to leverage these domain-specific experts to implement a {\it multi-expert-domain} (\med{}) classifier. This is a classifier of images $x$ over a label set ${\cal Y} = \{{\cal Y}_1, \ldots {\cal Y}_D\}$ composed of several expert domain label sets. PEFT parameters $W_d$ are available for each of the domains, and  the task is to combine domain parameters $W_d$ and VLM encoder $f$ into an encoder 
\begin{equation}
    f^{\text{\med{}}} = f \oplus_{d=1}^D W_d 
\end{equation}
 of good performance for any label set ${\cal C} \subseteq {\cal Y}$, while having no access to domain labels at inference time. While \med{} classification can be developed for any PEFT technique, we focus on LoRA~\cite{hu2022lora}, due to is large popularity and demonstrated success for computer vision problems.

\begin{figure*}[t] 
  \centering
   \includegraphics[width=0.6\linewidth, trim= 90 168 220 80, clip]{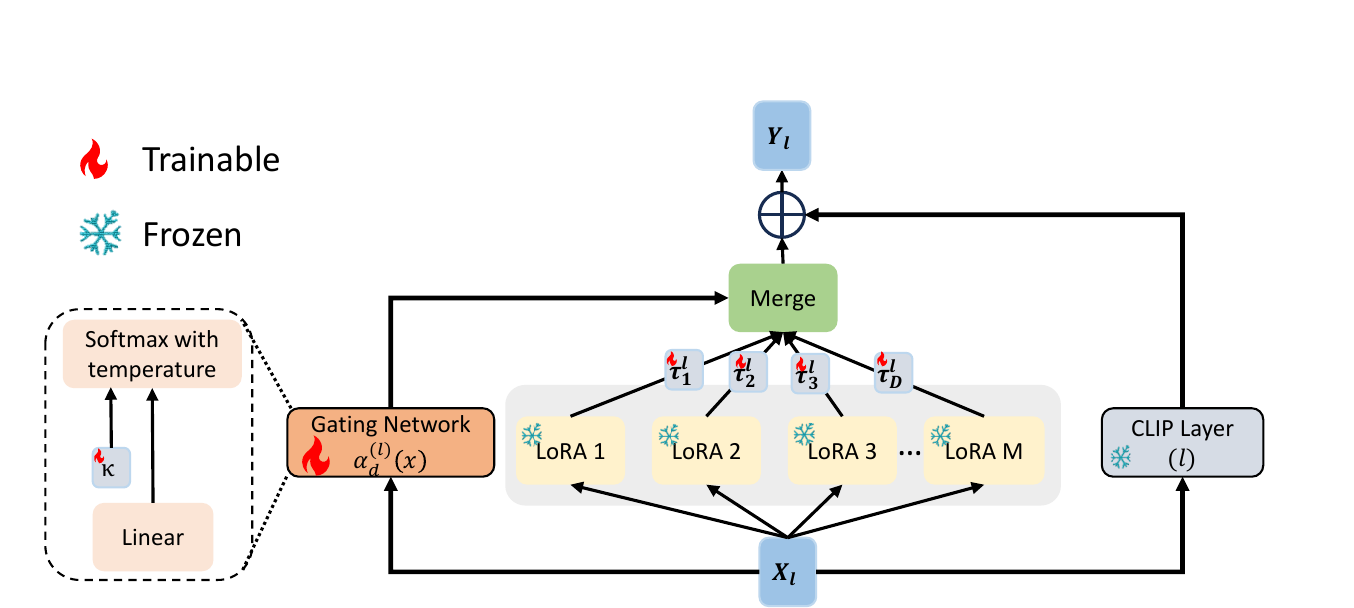}
\caption{The \ours{} network combines a VLM (such as CLIP) and a set of LoRAs learned independently for different image domains, which are kept frozen. \ours is an MoE architecture that relies on a combination of 1) domain routing by a gating network $\alpha^{(l)}(x)$ trained with domain supervision and 2) logit rescaling using a set of learned domain-specific temperature parameters $\tau_d^{(l)}$ in the last layer before softmax.}
   \label{fig:archIntro}
\end{figure*}

LoRA-based PEFT consists of adding to a parameter matrix $W \in \mathbb{R}^{p\times q}$ of the VLM encoder $f(x)$ an adaptation matrix $\Delta W = B A$, where $B \in  \mathbb{R}^{p \times k}$ and $A \in  \mathbb{R}^{k \times q}$ are two low-rank matrices, of shared intermediate dimensionality $k$, typically much smaller than $p$ or $q$.  The  residual parameters $\Delta W$ are learned with a cross-entropy loss, while leaving $W$ fixed. This allows adaptation with a  small number of parameters and few training examples. To implement \med{}, we adopt the Mixture-of-Experts (MoE)~\cite{shazeer2017outrageously} architecture of Figure~\ref{fig:archIntro}. At each adapted layer $\ell \in \{1, \dots, L\}$, the LoRA residual from expert $d$ is denoted as $\Delta h_d^{(\ell)}(x)=B^{(\ell)}_dA^{(\ell)}_d x$, where $B^{(\ell)}_d, A^{(\ell)}_d$ are the LoRA matrices available for domain $d$. Given a set of input-dependent domain mixing weights $\alpha_d^{(\ell)}(x) \ge 0 $ such that
$\sum_{d=1}^D \alpha_d^{(\ell)}(x) = 1,$ 
the \med{}-LoRA residual is computed with
\begin{equation}
\Delta h^{(\ell)}(x)
=
\sum_{d=1}^D
\alpha_d^{(\ell)}(x)\,
\Delta h_d^{(\ell)}(x).
\label{eq:med-lora}
\end{equation}
As usual for MoEs~\cite{shazeer2017outrageously}, the weights $\alpha_d^{(\ell)}(x)$ are computed by a gating network $\alpha (x)$  learned from a dataset of examples from all domains. This network learns to route the features derived from image $x$ to the associated domain expert. This produces a unified image encoder $f^{\text{\med{}}}(\cdot)$ that dynamically combines domain-specific expertise without requiring domain labels at inference.
In our implementation, the gating network combines linear layer $W_g$ and non-linearity $\rho$,
\begin{equation}
    g(x) = \rho(W_gx). 
\end{equation}
This generalizes two architectures previously proposed for LoRA mixing: MoLE~\cite{wu2024mixture} if $\rho$ is the softmax function 
\begin{equation}
\rho_d(x) = \frac{\exp(W_d x)}{\sum_{c = 1}^D \exp(W_cx)},
\label{eq:mole}
\end{equation}
and Phatgoose~\cite{muqeeth2024learning} if $\rho$ consists of a vector of sigmoid functions
\begin{equation}
\rho_d(x) = \frac{1}{1+\exp(-W_d x)}.
\label{eq:phat}
\end{equation}
However, there are significant differences to these works that we discuss next.

\subsection{Training}

Like most MoE methods, MoLE and Phatgoose assume that domain labels are unavailable during training. A significant challenge, in this case, is to guarantee that the training produces well-balance experts, e.g. avoiding the scenario where one expert is used for most of the data and the remaining only infrequently. This balancing is usually encouraged by introducing a balancing loss during training. Several such losses have  been proposed in the literature~\cite{mu2025comprehensive}. MoLE relies on
\begin{equation}
    {\cal L}_b = -\log(\prod_{d=1}^D \alpha_d), \quad \alpha_d = \frac{1}{L} \sum_{l=1}^l \alpha_d^{(l)},
    \label{eq:MoELoss}
\end{equation}
However, in the \med{} problem, the domain of each example $x$ can be trivially derived from its class label $y$. It suffices to search for the label in each of the domain label sets ${\cal Y}_d$, i.e.
\begin{equation}
    d^* = \arg\max_d \mathbbm{1}(y \in {\cal Y}_d)
\end{equation}
where $\mathbbm{1}(.)$ is equal to one when the argument holds and 0 otherwise. This motivates us to use a training objective consisting of two losses.
The first is the image {\it classification loss}, a cross-entropy loss over the complete label set $\cal Y$,
\begin{equation}
\mathcal{L}_{\text{cls}} = - \log p(y \mid x),
\end{equation}
where $y$ is the ground-truth class of $x$. The second is the {\it domain classification loss} that supervises the gating function $g$. Let
$\alpha_d(x)$ be as in (\ref{eq:MoELoss}). Given ground-truth domain $d^*$ for image $x$, 
\begin{equation}
\mathcal{L}_{\text{domain}} = - \log \alpha_{d^*}(x).
\end{equation}
This loss promotes domain-discriminative expert selection, reducing cross-domain interference. Its domain supervised nature guarantees much stronger discrimination between domains than (\ref{eq:MoELoss}), therefore inducing much more accurate domain selections by the gating network $\alpha$. Furthermore, if the number of training examples per domain is balanced, i.e. the same for each domain, the experts are automatically balanced. The overall training loss is
\begin{equation}
\mathcal{L}
=
\mathcal{L}_{\text{cls}}
+
\lambda \mathcal{L}_{\text{domain}},
\end{equation}
where $\lambda$ balances classification and routing supervision.

\subsection{Domain-sensitive logit calibration}

The main difficulty of \med{} classification is to combine LoRAs trained on non-overlaping datasets ${\cal D}_d$ from different domains $d$. This induces two challenges. The first is to guarantee that each example $x$ is routed to the LoRA $h_{d^*}(x)$ of the corresponding domain $d^*$. The supervised training of the previous section is a strong solution  to this problem. The second is to guarantee that the logits $z_c(x)$ can be effectively used for classification with (\ref{eq:softmax}). This is not guaranteed when the logits are produced by independently trained networks, as in \med{}, because the probabilities $p(c|x)$ are invariant to many rigid logit transformations. For example, due to the monotonicity of the exponential function
\begin{eqnarray}
\arg \max{(p(a|x), p(b|x))} &=& \arg \max (e^{z_a(x)}, e^{z_b(x)}) \nonumber \\
& = & \arg\max (e^{\gamma z_a(x)}, e^{\gamma z_b(x)}), \forall \gamma > 0 \label{eq:scaling}
\end{eqnarray}
and the relative ordering of the probabilities remains constant under the scaling of the logits by a common non-negative constant $\gamma$. This is commonly exploited to control the smoothness of output probability distributions of machine learning models using a temperature parameter. Since logit rescaling only affects the relative magnitudes of the probabilities, not their ranking, the logits of the optimal classifier are only defined up to a scaling by $\gamma$. While this is not a problem when all network parameters are learned simultaneously, it can derail a \med{} classifier, where the logits of each domain are produced by an independently trained LoRA. In this case, logits of domain $d$ are likely to be scaled by a value $\gamma_d \in \mathbb{R}^{|{\cal Y}_d|}$ different from the scaling $\gamma_m$ of domain $m$. It follows that, if class $a$ belongs to domain $d$ and class $b$ to domain $m$, the guarantee of (\ref{eq:scaling}) no longer holds, i.e
\begin{equation}
\exists \gamma_a \neq \gamma_b, \text{such that} \arg \max{(p(a|x), p(b|x))} \neq  \arg\max (e^{\gamma_a z_a(x)}, e^{\gamma_b z_b(x)}). \nonumber
\end{equation}
Since, when evaluating over the entire label set $\mathcal{Y}$, logits from heterogeneous domains ${\cal Y}_d$ compete directly under softmax normalization, a  domain that systematically produces larger-magnitude logits will induce the classifier to predict its classes even for out-of-domain examples. For example, in Figure~\ref{fig:archteaser}, the logit produced for the groundtruth-class of the image ('F-18 jet') by the expert of its domain (Figure~\ref{fig:archteaser} (b)) is smaller than the logit produced for another class ('Military Parade') by an out-of-domain expert (Figure~\ref{fig:archteaser} (c)). This breaks the fundamental assumption underlying VLM-based zero-shot inference: that similarity scores are globally calibrated across candidate classes. Hence, the \ours residual of (\ref{eq:med-lora}) can be dominated by the logit of the incorrect class, making it the class predicted by the \med{}-classifier, as shown in Figure~\ref{fig:archteaser} (d). As the number of domains increases, the probability of spurious high-confidence predictions grows, leading to degraded cross-domain performance.

To restore global logit comparability, we introduce a learnable temperature parameter $\tau_d > 0$ for each domain. For any class $c \in \mathcal{Y}_d$, logits are rescaled with
\begin{equation}
\tilde{z}_c(x) = \frac{z_c(x)}{\tau_d}.
\end{equation}
This is equivalent to applying domain-specific temperature scaling prior to softmax evaluation over $\mathcal{Y}$,
$
p(c \mid x)
=
\frac{
\exp(\tilde{z}_c(x))
}{
\sum_{c' \in \mathcal{Y}}
\exp(\tilde{z}_{c'}(x))
}.
$
This normalization preserves within-domain ranking while aligning cross-domain magnitudes. Learning the temperature parameters $\{\tau_d\}_{d=1}^D$ compensates for domain-specific amplification effects introduced by independent LoRA adaptation. Importantly, this mechanism introduces only $D$ scalar parameters and does not modify the VLM.

By combining domain-supervised routing with domain-sensitive logit scaling, \ours{} restores the key property required for zero-shot inference: logits across arbitrary candidate classes are globally comparable. This enables robust classification over any union of domain-specific classes without requiring domain labels at test time, thereby recovering scalable, truly zero-shot CLIP behavior under multi-domain specialization. Figure~\ref{fig:histteaser} illustrates the benefits of domain supervision and domain-sensitive logit scaling for \med{}. The figure shows histograms of the logits for classes of the EuroSAT dataset with (\ours{}) and  without (MoLE) these two components. While the MoLE histogram exhibits a significant overlap between the logits of DTD and EuroSAT classes, this overlap is drastically reduced for \ours{}.

\section{Experimental Results}
\label{sec:results}

We evaluate \ours{} on a diverse suite of nine fine-grained recognition benchmarks spanning object categories, textures, scenes, and actions. Following prior work~\cite{zanella2024low}, we train domain-specific LoRA experts and evaluate merged models under both \textbf{in-domain} and \textbf{cross-domain} protocols. 
The in-domain setting evaluates each dataset independently, while the cross-domain protocol forms a unified label space over the union of all classes, requiring models to discriminate across domains without domain labels at test time. This setting explicitly measures robustness to cross-domain interference. 
For all experiments, we split each dataset into two halves, \emph{base} (seen during training) and \emph{remaining} (unseen during training) classes, and report results on \emph{base} and \emph{all} (seen+unseen) splits. The \emph{all} split is the most comprehensive and challenging evaluation, as it measures joint generalization across both seen and unseen classes within a unified label space.

\begin{table*}[t]
\centering
\small
\setlength{\tabcolsep}{4pt}
\resizebox{\textwidth}{!}{
\begin{tabular}{lcccccccccccc}
\toprule
Method & Mean (\%) & $\Delta$ (\%) & Caltech & EuroSAT & Cars & Food & Pets & Flowers & DTD & UCF & FGVC \\
\midrule
\midrule
\multicolumn{12}{c}{\textbf{Base Split}} \\
\midrule Expert-LoRA  & \underline{85.48} & \underline{+15.36} & \underline{96.67} & \underline{94.81} & \underline{81.71} & 88.56 & \underline{95.96} & \textbf{97.91} & \textbf{81.02} & \textbf{83.61} & \textbf{49.10} \\
\midrule
\ours{} & \textbf{85.51} & \textbf{+15.39} & \textbf{97.40} & \textbf{94.90} & \textbf{81.73} & 88.70 & \textbf{96.17} & \textbf{97.91} & \underline{80.67} & \underline{83.20} & \underline{48.92} \\
KnOTS-DARE-TIES~\cite{stoica2025knots} & 78.17 & +8.05 & 93.76 & 78.81 & 74.74 & 89.62 & 94.21 & 88.51 & 67.94 & 77.51 & 38.42 \\
KnOTS-TIES~\cite{stoica2025knots} & 74.91 & +4.79 & 93.51 & 71.76 & 70.84 & 89.86 & 94.15 & 81.67 & 61.81 & 74.82 & 35.77 \\
PHATGOOSE~\cite{muqeeth2024learning} & 72.97 & +2.85 & 92.62 & 66.12 & 70.01 & 89.68 & 93.67 & 79.20 & 57.18 & 74.04 & 34.21 \\
MoLE & 71.80 & +1.69 & 93.11 & 72.90 & 63.47 & \underline{89.16} & 92.40 & 72.74 & 60.07 & 72.08 & 30.31 \\
LoRA-Mean & 70.12 & +0.00 & 89.94 & 60.29 & 66.42 & \textbf{89.70} & 93.04 & \underline{75.97} & 53.59 & 71.15 & 30.97 \\
\midrule
Single-LoRA & 42.35 & -27.77 & 83.54 & 44.17 & 26.39 & 60.56 & 53.32 & 32.10 & 33.68 & 45.40 & 1.98 \\
CLIP & 66.22 & -3.90 & 84.43 & 52.69 & 63.07 & 89.46 & 89.79 & 72.17 & 49.07 & 67.27 & 28.03 \\
\midrule
\multicolumn{12}{c}{\textbf{All Split}} \\
\midrule
Expert-LoRA & \underline{71.60} & \underline{+6.63} & \underline{92.56} & \underline{67.58} & \underline{70.10} & 83.29 & \underline{91.03} & \textbf{78.44} & \textbf{56.50} & \textbf{72.56} & \textbf{32.34} \\
\midrule
\ours{} & \textbf{71.80} & \textbf{+6.83} & \textbf{92.95} & \textbf{68.43} & \textbf{70.33} & 83.73 & \textbf{91.55} & \underline{78.32} & \underline{56.26} & \underline{72.43} & \underline{32.19} \\
KnOTS-DARE-TIES~\cite{stoica2025knots} & 68.66 & +3.69 & 90.64 & 58.28 & \textbf{70.51} & 84.79 & 90.57 & 76.00 & 49.88 & 69.73 & 27.51 \\
KnOTS-TIES~\cite{stoica2025knots} & 67.70 & +2.73 & 90.69 & 54.19 & 69.22 & 85.72 & 90.81 & 74.95 & 46.10 & 69.18 & 28.47 \\
PHATGOOSE~\cite{muqeeth2024learning} & 59.77 & -5.20 & 82.85 & 23.10 & 65.86 & 78.33 & 88.33 & 67.48 & 47.10 & 61.49 & 23.37 \\
MoLE & 65.85 & +0.88 & 90.19 & 56.25 & 65.03 & 85.02 & 88.66 & 70.85 & 44.86 & 66.56 & 25.20 \\
LoRA-Mean & 64.97 & +0.00 & 87.68 & 45.81 & 66.98 & \textbf{85.82} & 89.78 & 73.12 & 42.67 & 66.72 & 26.13 \\
\midrule
Single-LoRA & 36.28 & -28.69 & 73.63 & 28.10 & 27.27 & 57.44 & 54.18 & 22.25 & 25.65 & 36.98 & 0.99 \\
CLIP & 62.52 & -2.45 & 85.26 & 40.21 & 65.13 & \underline{85.04} & 87.38 & 70.60 & 40.54 & 63.94 & 24.57 \\
\bottomrule
\end{tabular}
}
\caption{Cross-domain results. Rows grouped by base/all splits. Best per column in bold, second-best underlined. $\Delta$ is improvement over LoRA-Mean within each split.}
\label{tab:cross_domain_results}
\end{table*}

\paragraph{\bf Datasets and training.} We use a suite of fine-grained recognition datasets commonly used to evaluate few-shot learning~\cite{zhou2022coop}. This includes Caltech101, EuroSAT, FGVC Aircraft, Cars, Pets, Flowers, DTD, UCF101 and Food101. We note that certain domain classes conflict with other domains. For example, ``water\_lilly'' in Caltech101 overlaps with classes in Oxford Flowers. To avoid this duplication, we exclude the following classes from Caltech101: water\_lilly, lotus, pagoda, pizza, sunflower, airplane, barrel, ice\_cream, car\_side, lobster, helicopter, soccer\_ball. In total, we obtain 783 classes across all datasets after filtering. For training, we follow the standard setup of prompt learning~\cite{zhou2022coop}. We train rank-2 LoRA experts with 16-shot images per class of the Base split. The AdamW optimizer is used with a batch size of 16 and a learning rate of 1e-4. For more details, refer to supplementary section 1.

\paragraph{\bf Baselines.} The proposed \ours{} approach was compared to several baselines. \ours is implemented with domain supervision during training, logit rescaling, and the softmax gating function of (\ref{eq:mole}). CLIP is the original CLIP model, without adaptation. Single-LoRA is the standard LoRA adaptation over the entire set of 10 datasets, using a LoRA with 10x the number of parameters of the individual expert LoRAs of \ours{}. LoRA-mean is a representative of methods that combine LoRA parameters through arithmetic operations, in this case the average of the 10 domain LoRAs. Expert-LoRA is the performance achieved with a domain oracle, i.e. when the domain of the test image is known and the corresponding LoRA always chosen. Intuitively, this is an upper bound for performance, since the "right"expert is always selected. However, we will see that a combination of LoRAs can usually slightly outperform this result. MoLE is identical to \ours{} without domain supervision and logit scaling.

\section{Results}

\paragraph{\bf Cross-Domain Evaluation.} Table~\ref{tab:cross_domain_results} reports performance under the cross-domain protocol, where class labels from multiple datasets are unified and logits compete under a single softmax. The table is organized in three sections. The top section shows the expected upper bound performance of Expert-LoRA. The middle section compares different \med{} approaches. Finally, the bottom section includes Single-Lora approach and the original CLIP model.

\begin{table*}[t]
\centering
\small
\setlength{\tabcolsep}{4pt}
\resizebox{\textwidth}{!}{
\begin{tabular}{lcccccccccccc}
\toprule
Method & Mean (\%) & $\Delta$ (\%) & Caltech & EuroSAT & Cars & Food & Pets & Flowers & DTD & UCF & FGVC \\
\midrule
\midrule
\multicolumn{12}{c}{\textbf{Base Split}} \\
\midrule
Expert-LoRA & \textbf{85.96} & \textbf{+14.10} & \underline{97.89} & \underline{95.02} & \underline{81.71} & 88.69 & \textbf{96.17} & \textbf{97.91} & \textbf{83.45} & \textbf{83.66} & \textbf{49.10} \\
\midrule
\ours{} & \underline{85.89} & \underline{+14.04} & \textbf{98.05} & \textbf{95.05} & \textbf{81.73} & 88.78 & \textbf{96.17} & \textbf{97.91} & \underline{82.99} & \underline{83.40} & \underline{48.92} \\
KnOTS-DARE-TIES~\cite{stoica2025knots} & 79.29 & +7.44 & 97.65 & 81.62 & 74.74 & 89.75 & 94.84 & 88.51 & 70.37 & 77.77 & 38.42 \\
KnOTS-TIES~\cite{stoica2025knots} & 76.26 & +4.41 & 97.40 & 74.60 & 70.84 & \underline{89.99} & 94.63 & 81.67 & 66.32 & 75.13 & 35.77 \\
PHATGOOSE~\cite{muqeeth2024learning} & 74.78 & +2.93 & 92.05 & 55.95 & 76.44 & 84.65 & 94.42 & 89.08 & 70.72 & 74.20 & 35.53 \\
MoLE & 72.90 & +1.05 & 96.92 & 73.95 & 63.47 & 89.27 & 92.72 & 72.74 & 64.47 & 72.29 & 30.31 \\
LoRA-Mean & 71.85 & +0.00 & 96.43 & 64.00 & 66.42 & \textbf{89.80} & \underline{93.46} & \underline{75.97} & 58.22 & 71.41 & 30.97 \\
\midrule
Single-LoRA & 43.40 & -28.45 & 85.16 & 44.33 & 26.61 & 61.21 & 54.70 & 32.67 & 35.42 & 48.55 & 1.98 \\
CLIP (Base) & 67.99 & -3.87 & 89.86 & 57.10 & 63.07 & \underline{89.57} & 90.11 & 72.17 & 54.40 & 67.58 & 28.03 \\
\midrule
\multicolumn{12}{c}{\textbf{All Split}} \\
\midrule
Expert-LoRA & \underline{72.27} & \underline{+5.77} & \textbf{94.43} & \underline{67.89} & \underline{70.19} & 83.79 & \underline{91.31} & \textbf{78.48} & \textbf{59.16} & \textbf{72.80} & \textbf{32.37} \\
\midrule
\ours{} & \textbf{72.33} & \textbf{+5.82} & \underline{94.38} & \textbf{68.63} & \textbf{70.35} & 83.90 & \textbf{91.63} & \underline{78.32} & \underline{58.69} & \textbf{72.80} & \underline{32.22} \\
KnOTS-DARE-TIES~\cite{stoica2025knots} & 69.78 & +3.27 & 93.69 & 60.78 & \textbf{70.53} & 85.18 & 91.06 & 76.00 & 53.31 & 69.92 & 27.51 \\
KnOTS-TIES~\cite{stoica2025knots} & 68.92 & +2.42 & 93.79 & 56.98 & 69.22 & 86.02 & 91.25 & 74.95 & 50.18 & 69.44 & 28.47 \\
PHATGOOSE~\cite{muqeeth2024learning} & 62.66 & -3.84 & 87.48 & 38.44 & 65.86 & 78.72 & 88.85 & 67.56 & 50.30 & 63.34 & 23.40 \\
MoLE & 66.97 & +0.46 & 93.54 & 57.79 & 65.03 & \underline{85.33} & 88.91 & 70.85 & 49.11 & 66.93 & 25.20 \\
LoRA-Mean & 66.50 & +0.00 & 92.46 & 48.57 & 66.98 & \textbf{86.05} & 90.16 & 73.12 & 47.87 & \underline{67.17} & 26.13 \\
\midrule
Single-LoRA & 37.47 & -29.03 & 76.34 & 28.58 & 27.50 & 58.07 & 55.14 & 22.53 & 27.60 & 40.52 & 0.99 \\
CLIP (Base) & 63.87 & -2.63 & 87.88 & 43.00 & 65.13 & 85.29 & 87.65 & 70.60 & 46.22 & 64.47 & 24.57 \\
\bottomrule
\end{tabular}
}
\caption{In-domain results. Rows grouped by base/all splits. Best per column in bold, second-best underlined. $\Delta$ is improvement over LoRA-Mean within each split.}
\label{tab:in_domain_results}
\end{table*}

\paragraph{Base Split.}
On the base split, \ours{} achieves the best mean accuracy of \textbf{85.51\%}, improving over LoRA-Mean (70.12\%) by \textbf{+15.39} points. Notably, \ours{} slightly surpasses the oracle Expert-LoRA (85.48\%). This can be explained by the LoRA combination of~(\ref{eq:med-lora}), which increases robustness to light overfitting by the expert LoRAs. It also achieves a large gain (+12.99) over MoLE, demonstrating the advantages of domain supervision and logit scaling. 
All models involving expert LoRAs improve dramatically on CLIP and Single-LoRA. The latter has significantly lower performance than even CLIP, perhaps due to overfitting of the large number of parameters being learned. Compared to existing parameter merging and model-editing approaches with  LoRAs, \ours{} outperforms \textsc{KnOTS-DARE-TIES} by more than \textbf{7} points and the gap is even larger relative to \textsc{PHATGOOSE}, which trails \ours{} by almost \textbf{12} points.

\paragraph{All Split.}
The \textbf{all split} has qualitatively similar results. All methods have weaker performance, due to the need to generalize to unseen classes, but the relative performances are similar. Again \ours{} slightly outperforms Expert-Lora and achieves large gains (close to $6$ points) over the other \med{} methods. 
The gains of this split are particularly meaningful, as it maximizes cross-domain interference: logits from independently trained experts must remain comparable across both domains and class partitions, and generalize to classes unseen at training. 

\paragraph{\bf In-Domain Evaluation}

Table~\ref{tab:in_domain_results} presents in-domain results, where each dataset is evaluated independently and no cross-domain competition occurs. The model  includes all LoRAs, and must ideally route images through the dataset expert.

\paragraph{Base Split.} \ours achieves 85.89\%, closely matching Expert-LoRA (85.96\%). 
The gains over the other \med{} methods are again significant ($7$-$13$ points). The significant gain over MoLE and PHATGOOSE illustrates the advantage of domain supervision. Because the weaker gating function of MoLE routes images through LoRAs of mismatched domains, performance degrades. On the other hand, the average LoRA of LoRA-mean is not very effective for most domains,  illustrating the limitations of adaptation by parameter arithmetic.

\paragraph{All Split.} Again, the results are qualitatively similar to those of the Base split.

\section{Ablations}

\paragraph{\bf \ours components.} Table \ref{tab:ablation_cross_domain_results} compares various implementations of \ours{} in the cross-domain setting, varying in terms of the inclusion, or not, of domain scaling and supervised learning, and the use of the  softmax of (\ref{eq:mole}) or sigmoid vector of (\ref{eq:phat}) as non-linearity of the gating network. The results are qualitatively the same for the Base and All splits. The weakest performance is achieved by the model without logit scaling and domain supervision. For the same non-linearity, the addition of domain supervision during training results in a large performance gain. Further adding logit scaling results in a smaller additional gain. We note that the gains of logit scaling are likely to increase for larger numbers of domains. Comparing the gating non-linearities, the softmax function clearly outperforms the sigmoid array, for comparable network configurations. Please refer to supplementary for ablation studies for the in-domain setting.

\begin{table*}[t]
\centering
\small
\setlength{\tabcolsep}{4pt}
\resizebox{\textwidth}{!}{
\begin{tabular}{cccc|cccccccccccc}
\toprule
Log & Dom & Soft & Sig & Mean (\%) & $\Delta$ (\%) & Caltech & EuroSAT & Cars & Food & Pets & Flowers & DTD & UCF & FGVC \\
Scal & Sup &  &  &  &  &  &  &  &  &  &  &  &  &  \\
\midrule
\midrule
\multicolumn{15}{c}{\textbf{Base Split}} \\
\midrule
\checkmark & \checkmark & \checkmark &  & \textbf{85.51} & \textbf{+15.39} & \textbf{97.40} & \underline{94.90} & \textbf{81.73} & 88.70 & \underline{96.17} & \textbf{97.91} & \underline{80.67} & \underline{83.20} & \underline{48.92} \\
 & \checkmark & \checkmark &  & \underline{85.43} & \underline{+15.31} & \underline{96.59} & \textbf{94.95} & \underline{81.41} & 88.48 & \textbf{96.28} & \textbf{97.91} & \textbf{80.90} & \textbf{83.25} & \textbf{49.10} \\
\checkmark & \checkmark &  & \checkmark & 79.21 & +9.09 & 96.27 & 89.55 & 73.46 & \underline{89.90} & 94.74 & \underline{87.75} & 68.98 & 76.73 & 35.53 \\
 & \checkmark &  & \checkmark & 78.55 & +8.43 & 95.46 & 89.38 & 73.69 & \textbf{89.95} & 94.26 & 84.14 & 66.44 & 76.63 & 36.97 \\
 &  & \checkmark &  & 71.80 & +1.69 & 93.11 & 72.90 & 63.47 & 89.16 & 92.40 & 72.74 & 60.07 & 72.08 & 30.31 \\
\midrule
\multicolumn{15}{c}{\textbf{All Split}} \\
\midrule
\checkmark & \checkmark & \checkmark &  & \textbf{71.80} & \textbf{+6.83} & \textbf{92.95} & 68.43 & \textbf{70.33} & 83.73 & \textbf{91.55} & \underline{78.32} & \textbf{56.26} & \underline{72.43} & \underline{32.19} \\
 & \checkmark & \checkmark &  & \underline{71.61} & \underline{+6.64} & \underline{91.77} & 67.47 & \underline{70.25} & 83.53 & \underline{91.20} & \textbf{78.60} & \textbf{56.26} & \textbf{73.09} & \textbf{32.34} \\
\checkmark & \checkmark &  & \checkmark & 69.97 & +5.00 & 92.02 & \textbf{69.04} & 69.39 & \textbf{85.94} & 91.17 & 76.13 & \underline{49.76} & 69.44 & 26.85 \\
 & \checkmark &  & \checkmark & 69.67 & +4.71 & 91.87 & \underline{68.94} & 69.08 & \underline{85.79} & 90.71 & 74.75 & 49.35 & 69.71 & 26.88 \\
 &  & \checkmark &  & 65.85 & +0.88 & 90.19 & 56.25 & 65.03 & 85.02 & 88.66 & 70.85 & 44.86 & 66.56 & 25.20 \\

\bottomrule
\end{tabular}
}
\caption{Cross-domain ablations. Rows grouped by base/all splits. Best per column in bold, second-best underlined. $\Delta$ is improvement over LoRA-Mean within each split.}
\label{tab:ablation_cross_domain_results}
\end{table*}

\begin{table}[t]
\centering
\scriptsize
\setlength{\tabcolsep}{6pt}
\caption{Ablation on LoRA ranks, SigLIP LoRA, and data efficiency in the cross-domain experiment under all split.}
\label{tab:rank_data_ablation}
\label{tab:data_eff_ablation}
\label{tab:rank_ablation}
\resizebox{\columnwidth}{!}{%
\begin{tabular}{lccc|c|cccc}
\toprule
\multirow{2}{*}{Method}
& \multicolumn{3}{c|}{LoRA Rank}
& \multicolumn{1}{c|}{SigLIP LoRA}
& \multicolumn{4}{c}{Data Efficiency} \\
\cmidrule(lr){2-4} \cmidrule(lr){5-5} \cmidrule(lr){6-9}
& r2 & r4 & r8 & r2 & 1 & 4 & 8 & 16 \\
\midrule
LoRA-Mean & 64.97 & 64.84 & 65.07 & 51.64 & 64.97 & 64.97 & 64.97 & 64.97 \\
MoLE & 65.85 & 65.82 & 65.51 & 50.81 & 65.26 & 65.63 & 65.88 & 65.85 \\
PHATGOOSE & 59.77 & 61.36 & 61.98 & 49.67 & 61.51 & 59.42 & 57.75 & 59.77 \\
KnOTS-TIES & 67.70 & 67.29 & 67.84 & 53.23 & 67.70 & 67.70 & 67.70 & 67.70 \\
\noindent\textbf{\ours} & \noindent\textbf{71.80} & \noindent\textbf{72.20} & \noindent\textbf{72.48}
& \noindent\textbf{62.27} & \noindent\textbf{71.70} & \noindent\textbf{71.50} & \noindent\textbf{71.53} & \noindent\textbf{71.80} \\
\bottomrule
\end{tabular}%
}
\end{table}

\begin{figure}[h]
    \centering
    \includegraphics[width=0.8\linewidth]{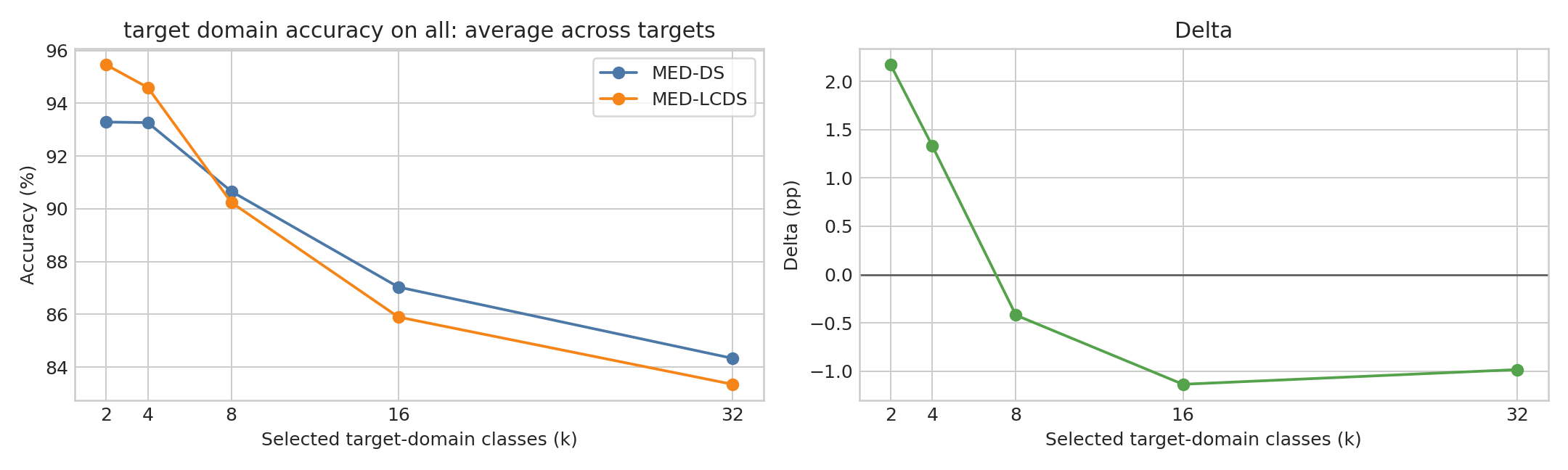}
    \caption{Effect of logit scaling on performance for a target-domain imbalance sweep. Left: accuracy vs number of target-domain classes ($k$) for \ours and MED-DS. Right: difference between the two approaches.}
    \label{fig:target-domain-k-sweep}
\end{figure}

\paragraph{\bf Sensitivity to class imbalance.} We conduct a target-domain imbalance sweep experiment to better understand the role of LS in Figure~\ref{fig:target-domain-k-sweep}. Here, one target domain is restricted to $k$ classes while the remaining domains retain their original class sets. Figure~\ref{fig:target-domain-k-sweep} (left) shows that LS is more useful in the highly imbalanced settings. In particular, at $k=2$, \ours improves target-domain accuracy over MED-DS by $+2.2$ points; at $k=4$, the gains are smaller, at $+1.3$ points. This indicates that LS compensates for domain-level score inflation rather than only fitting balanced priors.

\paragraph{\bf Comparisons to MED baselines with SigLIP and different LoRA ranks}  Table~\ref{tab:rank_ablation} shows the backbone architecture and LoRA rank ablation, for  all-split mean accuracies. The gains of \ours are almost constant across LoRA ranks, and even more significant for SigLIP than CLIP. It also shows that the method is model agnostic and does not rely on the biases of an architecture.

\paragraph{\bf Domain mixing.}
\label{sec:ablation_k}

In the \med{} problem, the classifier should maintain good performance over any label subset ${\cal C} \subseteq {\cal Y}$ combining classes from any of the domains. To test the robustness of to the size of the label set $\cal C$, we performed an experiment with increasing number of classes $k$ per domain. For a given $k$, the difficulty of the classification also depends on the class composition. To further analyze robustness to variations in label-space complexity, we consider two approaches for class selection: (1) \textbf{First-$k$}, which uses the first $k$ classes from each domain, and 
(2) \textbf{Random-$k$} (Top-$k$), where $k$ classes are randomly sampled per domain. The rationale is that datasets are usually organized by class similarity. So, while First-$k$ should create a more homogeneous classification problem, Random-$k$ has more heterogeneous coverage of the domain classes. We consider $k \in \{5, 10, 20, 40\}$ and report mean accuracy across domains for \textit{All Split}.

Tables~\ref{tab:ablation_firstk_all} and~\ref{tab:ablation_randomk_all} show that \med{} performance is indeed sensitive to the class composition.  Classification performance degrades as the number of classes increases and is higher for Random-$k$ than for First-$k$. This is expected, since problems with more classes and greater class similarity tend to be more challenging. However, in both cases, \ours{} is superior to LoRA-Mean and MoLE, and tracks the performance of Expert-LoRA, achieving gains similar to those of Table~\ref{tab:cross_domain_results} where $k$ is the size of the entire label set $\cal Y$. The gains ($\Delta$) of \ours over MoLE are also larger for the more difficult problem and grow with the total number of classes. This shows that \ours{} is quite robust to the class composition of the \med{} problem.

\begin{table*}[t]
\centering
\small
\setlength{\tabcolsep}{6pt}

\begin{minipage}{0.48\textwidth}
\centering
\resizebox{\textwidth}{!}{
\begin{tabular}{lcccc}
\toprule
Method 
& $k=5$ 
& $k=10$ 
& $k=20$ 
& $k=40$ \\
\midrule

Expert-LoRA
& 91.14
& 85.37
& 83.58
& 79.15 \\
\midrule

LoRA-Mean
& 81.34
& 74.88
& 72.43
& 66.63 \\

MoLE
& 82.94
& 76.80
& 73.58
& 66.99 \\

\ours{}
& \textbf{91.37}
& \textbf{85.34}
& \textbf{83.52}
& \textbf{79.39} \\

\midrule
$\Delta$ & 8.43 & 8.54 & 9.94 & 12.4 \\

\bottomrule
\end{tabular}
}
\caption{First-$k$ class selection (All split). 
$\Delta$ is the difference between \ours and  MoLE. Mean accuracy (\%) across domains. }
\label{tab:ablation_firstk_all}
\end{minipage}
\hfill
\begin{minipage}{0.48\textwidth}
\centering
\resizebox{\textwidth}{!}{
\begin{tabular}{lcccc}
\toprule
Method 
& $k=5$ 
& $k=10$ 
& $k=20$ 
& $k=40$ \\
\midrule

Expert-LoRA
& 92.33
& 90.07
& 83.49
& 78.00 \\
\midrule

LoRA-Mean
& 85.48
& 86.39
& 79.57
& 71.50 \\

MoLE
& 87.47
& 86.92
& 79.72
& 72.25 \\

\ours{}
& \textbf{92.30}
& \textbf{90.21}
& \textbf{83.53}
& \textbf{77.97} \\
\midrule
$\Delta$ & 4.83 & 3.29 & 3.81 & 5.72 \\

\bottomrule
\end{tabular}
}
\caption{Random-$k$ (Top-$k$) class selection (All split). 
$\Delta$ is the difference between \ours and  MoLE. Mean accuracy (\%) across domains. }
\label{tab:ablation_randomk_all}
\end{minipage}

\end{table*}

\paragraph{\bf Data Efficiency.}
To evaluate the data efficiency of \ours{}, we consider a \textbf{one-shot training} setting where the expert LoRAs remain fixed and only the gating network and temperature scaling are trained using a single labeled example per class. This experiment measures how much supervision is required to learn effective routing between pretrained experts. Table~\ref{tab:cross_domain_results_one_shot} shows that \ours{} achieves a mean accuracy of \textbf{71.71\%} on the \textit{All Split}, outperforming LoRA-Mean (64.97\%) by \textbf{+6.75} points and MoLE (65.26\%) by \textbf{+6.46} points. 
Table~\ref{tab:rank_ablation} additionally shows the results for 4, 8 and 16 shots. The gains are consistent across most domains as compared to the 16-shot setting in Table~\ref{tab:cross_domain_results}. This demonstrates that \ours{} is highly data-efficient, maintaining strong cross-domain performance even when the routing mechanism is trained with minimal data.

\begin{figure*}[t]
\centering

\begin{minipage}{0.48\linewidth}
    \centering
    \includegraphics[width=\linewidth, trim=120 150 120 155, clip]{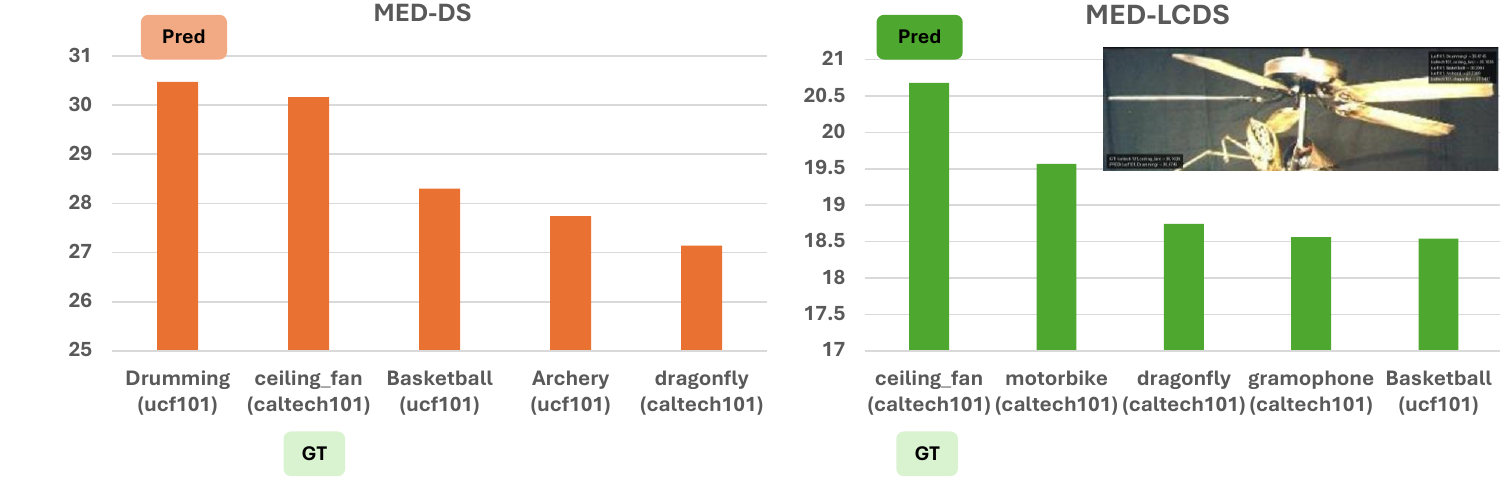}
\end{minipage}
\hfill
\begin{minipage}{0.48\linewidth}
    \centering
    \includegraphics[width=\linewidth, trim=140 170 120 140, clip]{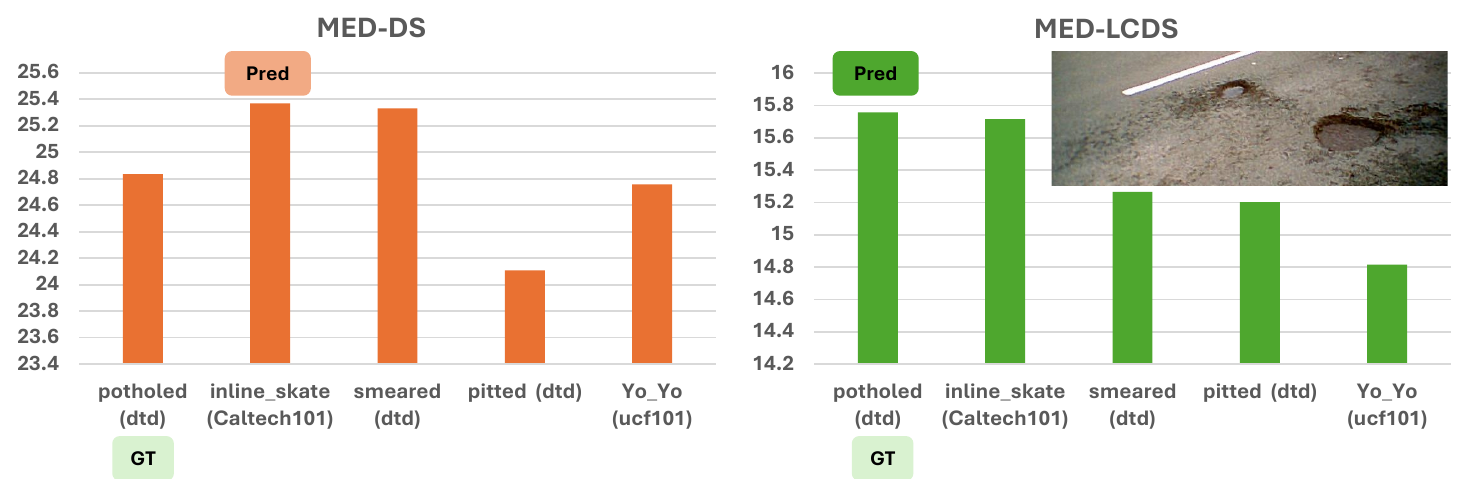}
\end{minipage}
\caption{\med{} logits with (green) and without (orange) logit scaling. Class (domain) shown in the horizontal axis, and ground-truth (predicted) class shown at the bottom (top). In the absence of logit scaling, logits are poorly calibrated across domains, and the predicted  class can be due to a spurious logit outside the image domain.}
\label{fig:qual_results}
\end{figure*}

\begin{table*}[t]
\centering
\small
\setlength{\tabcolsep}{4pt}
\resizebox{\textwidth}{!}{
\begin{tabular}{lcccccccccccc}
\toprule
Method & Mean (\%) & $\Delta$ (\%) & Caltech & EuroSAT & Cars & Food & Pets & Flowers & DTD & UCF & FGVC \\
\midrule
\midrule
\multicolumn{12}{c}{\textbf{All Split}} \\
\midrule
\ours{} & \textbf{71.71} & \textbf{+6.75} & \textbf{92.07} & \textbf{69.42} & \textbf{70.15} & 83.46 & \textbf{91.06} & \textbf{78.20} & \textbf{56.09} & \textbf{72.85} & \textbf{32.13} \\
MoLE & \underline{65.26} & \underline{+0.29} & \underline{90.00} & \underline{50.00} & 66.32 & \underline{85.22} & 89.32 & 71.05 & 42.02 & 66.16 & \underline{27.27} \\
LoRA-Mean & 64.97 & +0.00 & 87.68 & 45.81 & \underline{66.98} & \textbf{85.82} & \underline{89.78} & \underline{73.12} & \underline{42.67} & \underline{66.72} & 26.13 \\
\bottomrule
\end{tabular}
}
\caption{Cross-domain results for all split with one-shot training. Best per column in bold, second-best underlined. $\Delta$ is improvement over LoRA-Mean within each split.}
\label{tab:cross_domain_results_one_shot}
\end{table*}

Overall, these ablations confirm that domain-aware routing with logit scaling enables efficient, scalable expert merging that stays accurate and interference-free as class count grows.

\section{Qualitative Analysis}

Figure~\ref{fig:qual_results} compares the predictions of \ours{} without (orange) and with (green) logit scaling.
When combining domain-specialized experts without logit scaling, the model exhibits more cross-domain interference: logits from unrelated domains attain larger magnitudes and dominate the joint softmax, leading to incorrect predictions.
The use of domain-wise logit scaling (right) aligns logit magnitude across experts while preserving within-domain discrimination. This restores global comparability in the joint label space and reduces artificial dominance by out-of-domain classes. As a result, the correct class receives the highest calibrated logit, demonstrating that logit scaling effectively mitigates cross-domain interference and stabilizes multi-domain predictions. For additional qualitative results, refer to supplementary material.

\section{Conclusion}

In this work, we studied the \med{} classification problem, which seeks to merge domain-specific LoRA adapters for CLIP under realistic open-set settings, where base (seen) and new (unseen) classes from multiple datasets must be evaluated jointly. Our analysis revealed \emph{cross-domain interference} as the key limitation of existing approaches: independently trained experts produce miscalibrated logits that compete improperly under a shared softmax, leading to degradation in unified label spaces, especially in the challenging \emph{all} class split.
We proposed \ours{}, a domain-aware data-efficient mixture-of-experts framework that combines supervised domain routing with domain-wise logit scaling. The routing enforces structured expert selection, while learned temperature parameters restore cross-domain logit comparability without affecting within-domain discrimination.
Across extensive experiments, \ours{} consistently outperformed previous approaches to the \med{} classification problem. It substantially outperforms naive merging and matches or surpasses oracle expert selection, while preserving in-domain accuracy. Ablations with varying label-space sizes further confirm that MED scales robustly as cross-domain competition increases.
In summary, we show that restoring global logit alignment is essential for building unified, interference-resistant multi-domain vision–language systems.

\section*{Acknowledgements}

This work was partially funded by NSF awards IIS-2303153 and NAIRR-240300, a NVIDIA Academic grant, and a gift from Qualcomm. We also acknowledge and thank the use of the Nautilus platform for some of the experiments.

\bibliographystyle{splncs04}
\bibliography{main}

\clearpage
\appendix

\section{Implementation details}
We use 16-shot data from each class to train the baseline LoRAs on each domain and use 16-shot/1-shot data for training the gating network and temperature parameters of \ours{}. The hyperparameter settings for training the LoRAs are shown in Table~\ref{tab:method_hparams}. The total number of classes considered in each domain after filtering with the base and new splits are shown in Table~\ref{tab:cross_domain_classes}.

\begin{table*}[t]
\centering
\small
\setlength{\tabcolsep}{6pt}
\resizebox{0.7\textwidth}{!}{
\begin{tabular}{l l c l l}
\toprule
\textbf{Method} & \textbf{Batch Size} & \textbf{Optimizer} & \textbf{Learning rate} \\
\midrule
MED  & 64 & Adam  & 2e-3 \\
MoLE & 64 & Adam  & 2e-3 \\
Single-LoRA & 64 & AdamW & 1e-3\\
\bottomrule
\end{tabular}
}
\caption{Training hyperparameters used for each method.}
\label{tab:method_hparams}
\end{table*}

\begin{table*}[t]
\centering
\small
\setlength{\tabcolsep}{6pt}
\resizebox{0.5\textwidth}{!}{
\begin{tabular}{lccc}
\toprule
\textbf{Domain} & \textbf{All} & \textbf{Base} & \textbf{New} \\
\midrule
Caltech101     & 89  & 46 & 43 \\
EuroSAT        & 10  & 5  & 5  \\
Stanford Cars  & 196 & 98 & 98 \\
Food101        & 101 & 51 & 50 \\
Oxford Pets    & 37  & 19 & 18 \\
Oxford Flowers & 102 & 51 & 51 \\
DTD            & 47  & 24 & 23 \\
UCF101         & 101 & 51 & 50 \\
FGVC Aircraft  & 100 & 50 & 50 \\
\midrule
\textbf{Total} & \textbf{783} & \textbf{395} & \textbf{388} \\
\bottomrule
\end{tabular}
}
\caption{Number of classes per dataset after filtering for cross-domain experiments.}
\label{tab:cross_domain_classes}
\end{table*}

\begin{table*}[t]
\centering
\small
\setlength{\tabcolsep}{4pt}
\resizebox{\textwidth}{!}{
\begin{tabular}{cccc|cccccccccccc}
\toprule
Log & Dom & Soft & Sig & Mean (\%) & $\Delta$ (\%) & Caltech & EuroSAT & Cars & Food & Pets & Flowers & DTD & UCF & FGVC \\
Scal & Sup &  &  & &  &  &  &  &  &  &  &  &  &  \\
\midrule
\midrule
\multicolumn{15}{c}{\textbf{Base Split}} \\
\midrule
\checkmark & \checkmark & \checkmark & & \underline{85.89} & +14.04 & \underline{98.05} & \textbf{95.05} & \textbf{81.73} & 88.78 & \underline{96.17} & \textbf{97.91} & 82.99 & \underline{83.40} & 48.92 \\
& \checkmark & \checkmark & & {\bf 85.94} & +14.08 & \textbf{98.13} & \textbf{95.05} & 81.41 & 88.62 & \textbf{96.33} & \textbf{97.91} & \textbf{83.56} & 83.30 & \textbf{49.10} \\
\checkmark & \checkmark & & \checkmark & 79.92 & +8.06 & 97.65 & 90.71 & 73.46 & \underline{89.99} & 94.90 & 87.75 & 72.22 & 77.04 & 35.53 \\
& \checkmark & & \checkmark & 79.35 & +7.50 & 97.49 & 90.33 & 73.69 & \textbf{90.05} & 94.74 & 84.14 & 69.79 & 76.94 & 36.97 \\
& &\checkmark & & 72.90 & +1.05 & 96.92 & 73.95 & 63.47 & 89.27 & 92.72 & 72.74 & 64.47 & 72.29 & 30.31 \\
\midrule
\multicolumn{15}{c}{\textbf{All Split}} \\
\midrule
\checkmark & \checkmark & \checkmark & &  \underline{72.33} & +5.82 & \underline{94.38} & 68.63 & \textbf{70.35} & 83.90 & \textbf{91.63} & 78.32 & 58.69 & \textbf{72.80} & 32.22 \\
& \checkmark & \checkmark & &  \textbf{72.56} & +6.06 & 94.28 & 70.54 & \underline{70.22} & 83.80 & \textbf{91.63} & \underline{78.40} & \underline{58.98} & 72.35 & \textbf{32.82} \\
\checkmark & \checkmark & &\checkmark &  70.88 & +4.38 & 93.84 & \underline{70.62} & 69.39 & \underline{86.12} & 91.41 & 76.13 & 53.84 & 69.73 & 26.85 \\
& \checkmark & &\checkmark &  70.68 & +4.17 & 94.04 & \textbf{71.02} & 69.10 & \textbf{86.13} & 91.17 & 74.75 & 53.07 & 69.92 & 26.88 \\
&  & \checkmark & &  66.97 & +0.46 & 93.54 & 57.79 & 65.03 & 85.33 & 88.91 & 70.85 & 49.11 & 66.93 & 25.20 \\
\bottomrule
\end{tabular}
}
\caption{In-domain results. Rows grouped by base/all splits. Best per column in bold, second-best underlined. $\Delta$ is improvement over LoRA-Mean within each split.}
\label{tab:ablation_in_domain_results}
\end{table*}

\section{Comparison to Prior Methods}

\paragraph{Base Split.}
On the base split, \ours{} achieves the best mean accuracy of \textbf{85.51\%}, improving over LoRA-Mean (70.12\%) by \textbf{+15.39} points and slightly surpassing the oracle Expert-LoRA (85.48\%). This improvement can be attributed to the LoRA combination with gating, which mitigates mild overfitting in individual expert adapters while preserving domain-specific knowledge.

Compared to existing parameter merging and model-editing approaches with  LoRAs, \ours{} shows a substantial advantage. In particular, \ours{} outperforms \textsc{KnOTS-DARE-TIES} by more than \textbf{7} points and \textsc{MoLE} by nearly \textbf{14} points in mean accuracy. The gap is even larger relative to \textsc{PHATGOOSE}, which trails \ours{} by almost \textbf{12} points. These differences highlight the limitations of weight-space merging strategies when combining heterogeneous domain experts. While methods such as \textsc{KnOTS} and \textsc{PHATGOOSE} attempt to reconcile conflicting parameter updates across experts, they often introduce interference that degrades downstream classification performance.

Across individual datasets, \ours{} achieves the best performance on most domains, including Caltech, EuroSAT, Cars, and Pets, while remaining competitive on the remaining datasets. In contrast, the prior methods exhibit larger performance fluctuations across domains, suggesting weaker robustness when merging experts trained on diverse data distributions.

\paragraph{All Split.}
The \textbf{all split} presents a more challenging setting where models must generalize to unseen classes. As expected, overall accuracy decreases relative to the base split, but the relative ranking of methods remains largely consistent. \ours{} again achieves the best mean accuracy at \textbf{71.80\%}, outperforming LoRA-Mean (64.97\%) by \textbf{+6.83} points and slightly improving over the oracle Expert-LoRA (71.60\%).

Compared to other approaches, \ours{} maintains a clear advantage. Specifically, \ours{} exceeds \textsc{KnOTS-DARE-TIES} by more than \textbf{3} points and improves over \textsc{MoLE} by nearly \textbf{6} points in mean accuracy. The gap with \textsc{PHATGOOSE} becomes even more pronounced in this split, where its performance drops significantly, indicating sensitivity to distribution shift and class generalization.

These improvements are particularly meaningful in the all-split setting, where cross-domain interference is strongest. Logits from independently trained experts must remain comparable across domains while also generalizing to unseen classes. Weight-merging methods such as \textsc{KnOTS}, \textsc{MoLE}, and \textsc{PHATGOOSE} operate purely in parameter space and do not explicitly enforce such calibration, which can lead to degraded performance when experts disagree. In contrast, \ours{} maintains stable cross-domain behavior while preserving expert specialization.

\begin{table*}[t]
\centering
\small
\setlength{\tabcolsep}{4pt}
\resizebox{\textwidth}{!}{
\begin{tabular}{lcccccccccccc}
\toprule
Method & Mean (\%) & $\Delta$ (\%) & Caltech & EuroSAT & Cars & Food & Pets & Flowers & DTD & UCF & FGVC \\
\midrule
\midrule
\multicolumn{12}{c}{\textbf{Base Split}} \\
\midrule Expert-LoRA  & \underline{85.48} & \underline{+15.36} & \underline{96.67} & \underline{94.81} & \underline{81.71} & 88.56 & \underline{95.96} & \textbf{97.91} & \textbf{81.02} & \textbf{83.61} & \textbf{49.10} \\
\midrule
\textbf{\ours{}} & \textbf{85.51} & \textbf{+15.39} & \textbf{97.40} & \textbf{94.90} & \textbf{81.73} & 88.70 & \textbf{96.17} & \textbf{97.91} & \underline{80.67} & \underline{83.20} & \underline{48.92} \\
KnOTS-DARE-TIES~\cite{stoica2025knots} & 78.17 & +8.05 & 93.76 & 78.81 & 74.74 & 89.62 & 94.21 & 88.51 & 67.94 & 77.51 & 38.42 \\
KnOTS-TIES~\cite{stoica2025knots} & 74.91 & +4.79 & 93.51 & 71.76 & 70.84 & 89.86 & 94.15 & 81.67 & 61.81 & 74.82 & 35.77 \\
PHATGOOSE~\cite{muqeeth2024learning} & 72.97 & +2.85 & 92.62 & 66.12 & 70.01 & 89.68 & 93.67 & 79.20 & 57.18 & 74.04 & 34.21 \\
MoLE~\cite{wu2024mixture} & 71.80 & +1.69 & 93.11 & 72.90 & 63.47 & \underline{89.16} & 92.40 & 72.74 & 60.07 & 72.08 & 30.31 \\
LoRA-Mean & 70.12 & +0.00 & 89.94 & 60.29 & 66.42 & \textbf{89.70} & 93.04 & \underline{75.97} & 53.59 & 71.15 & 30.97 \\
\midrule
Single-LoRA & 42.35 & -27.77 & 83.54 & 44.17 & 26.39 & 60.56 & 53.32 & 32.10 & 33.68 & 45.40 & 1.98 \\
CLIP & 66.22 & -3.90 & 84.43 & 52.69 & 63.07 & 89.46 & 89.79 & 72.17 & 49.07 & 67.27 & 28.03 \\
\midrule
\multicolumn{12}{c}{\textbf{All Split}} \\
\midrule
Expert-LoRA & \underline{71.60} & \underline{+6.63} & \underline{92.56} & \underline{67.58} & \underline{70.10} & 83.29 & \underline{91.03} & \textbf{78.44} & \textbf{56.50} & \textbf{72.56} & \textbf{32.34} \\
\midrule
\textbf{\ours{}} & \textbf{71.80} & \textbf{+6.83} & \textbf{92.95} & \textbf{68.43} & \textbf{70.33} & 83.73 & \textbf{91.55} & \underline{78.32} & \underline{56.26} & \underline{72.43} & \underline{32.19} \\
KnOTS-DARE-TIES~\cite{stoica2025knots} & 68.66 & +3.69 & 90.64 & 58.28 & \textbf{70.51} & 84.79 & 90.57 & 76.00 & 49.88 & 69.73 & 27.51 \\
KnOTS-TIES~\cite{stoica2025knots} & 67.70 & +2.73 & 90.69 & 54.19 & 69.22 & 85.72 & 90.81 & 74.95 & 46.10 & 69.18 & 28.47 \\
PHATGOOSE~\cite{muqeeth2024learning} & 59.77 & -5.20 & 82.85 & 23.10 & 65.86 & 78.33 & 88.33 & 67.48 & 47.10 & 61.49 & 23.37 \\
MoLE~\cite{wu2024mixture} & 65.85 & +0.88 & 90.19 & 56.25 & 65.03 & 85.02 & 88.66 & 70.85 & 44.86 & 66.56 & 25.20 \\
LoRA-Mean & 64.97 & +0.00 & 87.68 & 45.81 & 66.98 & \textbf{85.82} & 89.78 & 73.12 & 42.67 & 66.72 & 26.13 \\
\midrule
Single-LoRA & 36.28 & -28.69 & 73.63 & 28.10 & 27.27 & 57.44 & 54.18 & 22.25 & 25.65 & 36.98 & 0.99 \\
CLIP & 62.52 & -2.45 & 85.26 & 40.21 & 65.13 & \underline{85.04} & 87.38 & 70.60 & 40.54 & 63.94 & 24.57 \\
\bottomrule
\end{tabular}
}
\caption{Cross-domain results. Rows grouped by base/all splits. Best per column in bold, second-best underlined. $\Delta$ is improvement over LoRA-Mean within each split.}
\label{tab:cross_domain_results_comparison}
\end{table*}

\begin{table*}[t]
\centering
\small
\setlength{\tabcolsep}{4pt}
\resizebox{\textwidth}{!}{
\begin{tabular}{lcccccccccccc}
\toprule
Method & Mean (\%) & $\Delta$ (\%) & Caltech & EuroSAT & Cars & Food & Pets & Flowers & DTD & UCF & FGVC \\
\midrule
\midrule
\multicolumn{12}{c}{\textbf{Base Split}} \\
\midrule Expert-LoRA  & \underline{85.48} & \underline{+15.36} & \underline{96.67} & \underline{94.81} & \underline{81.71} & 88.56 & \underline{95.96} & \textbf{97.91} & \textbf{81.02} & \textbf{83.61} & \textbf{49.10} \\
\midrule
\textbf{\ours{}} & \textbf{85.51} & \textbf{+15.39} & \textbf{97.40} & \textbf{94.90} & \textbf{81.73} & 88.70 & \textbf{96.17} & \textbf{97.91} & \underline{80.67} & \underline{83.20} & \underline{48.92} \\
KnOTS-DARE-TIES~\cite{stoica2025knots} & 79.29 & +7.44 & 97.65 & 81.62 & 74.74 & 89.75 & 94.84 & 88.51 & 70.37 & 77.77 & 38.42 \\
KnOTS-TIES~\cite{stoica2025knots} & 76.26 & +4.41 & 97.40 & 74.60 & 70.84 & \underline{89.99} & 94.63 & 81.67 & 66.32 & 75.13 & 35.77 \\
PHATGOOSE~\cite{muqeeth2024learning} & 74.78 & +2.93 & 92.05 & 55.95 & 76.44 & 84.65 & 94.42 & 89.08 & 70.72 & 74.20 & 35.53 \\
MoLE~\cite{wu2024mixture} & 71.80 & +1.69 & 93.11 & 72.90 & 63.47 & \underline{89.16} & 92.40 & 72.74 & 60.07 & 72.08 & 30.31 \\
LoRA-Mean & 70.12 & +0.00 & 89.94 & 60.29 & 66.42 & \textbf{89.70} & 93.04 & \underline{75.97} & 53.59 & 71.15 & 30.97 \\
\midrule
Single-LoRA & 42.35 & -27.77 & 83.54 & 44.17 & 26.39 & 60.56 & 53.32 & 32.10 & 33.68 & 45.40 & 1.98 \\
CLIP & 66.22 & -3.90 & 84.43 & 52.69 & 63.07 & 89.46 & 89.79 & 72.17 & 49.07 & 67.27 & 28.03 \\
\midrule
\multicolumn{12}{c}{\textbf{All Split}} \\
\midrule
Expert-LoRA & \underline{72.27} & \underline{+5.77} & \textbf{94.43} & \underline{67.89} & \underline{70.19} & 83.79 & \underline{91.31} & \textbf{78.48} & \textbf{59.16} & \textbf{72.80} & \textbf{32.37} \\
\midrule
\textbf{\ours{}} & \textbf{72.33} & \textbf{+5.82} & \underline{94.38} & \textbf{68.63} & \textbf{70.35} & 83.90 & \textbf{91.63} & \underline{78.32} & \underline{58.69} & \textbf{72.80} & \underline{32.22} \\
KnOTS-DARE-TIES~\cite{stoica2025knots} & 69.78 & +3.27 & 93.69 & 60.78 & \textbf{70.53} & 85.18 & 91.06 & 76.00 & 53.31 & 69.92 & 27.51 \\
KnOTS-TIES~\cite{stoica2025knots} & 68.92 & +2.42 & 93.79 & 56.98 & 69.22 & 86.02 & 91.25 & 74.95 & 50.18 & 69.44 & 28.47 \\
PHATGOOSE~\cite{muqeeth2024learning} & 62.66 & -3.84 & 87.48 & 38.44 & 65.86 & 78.72 & 88.85 & 67.56 & 50.30 & 63.34 & 23.40 \\
MoLE~\cite{wu2024mixture} & 66.97 & +0.46 & 93.54 & 57.79 & 65.03 & \underline{85.33} & 88.91 & 70.85 & 49.11 & 66.93 & 25.20 \\
LoRA-Mean & 66.50 & +0.00 & 92.46 & 48.57 & 66.98 & \textbf{86.05} & 90.16 & 73.12 & 47.87 & \underline{67.17} & 26.13 \\
\midrule
Single-LoRA & 37.47 & -29.03 & 76.34 & 28.58 & 27.50 & 58.07 & 55.14 & 22.53 & 27.60 & 40.52 & 0.99 \\
CLIP (Base) & 63.87 & -2.63 & 87.88 & 43.00 & 65.13 & 85.29 & 87.65 & 70.60 & 46.22 & 64.47 & 24.57 \\
\bottomrule
\end{tabular}
}
\caption{In-domain results. Rows grouped by base/all splits. Best per column in bold, second-best underlined. $\Delta$ is improvement over LoRA-Mean within each split.}
\label{tab:in_domain_results_comparison}
\end{table*}

\section{Logits Visualization}
\begin{figure}
    \centering
    \includegraphics[width=0.8\linewidth]{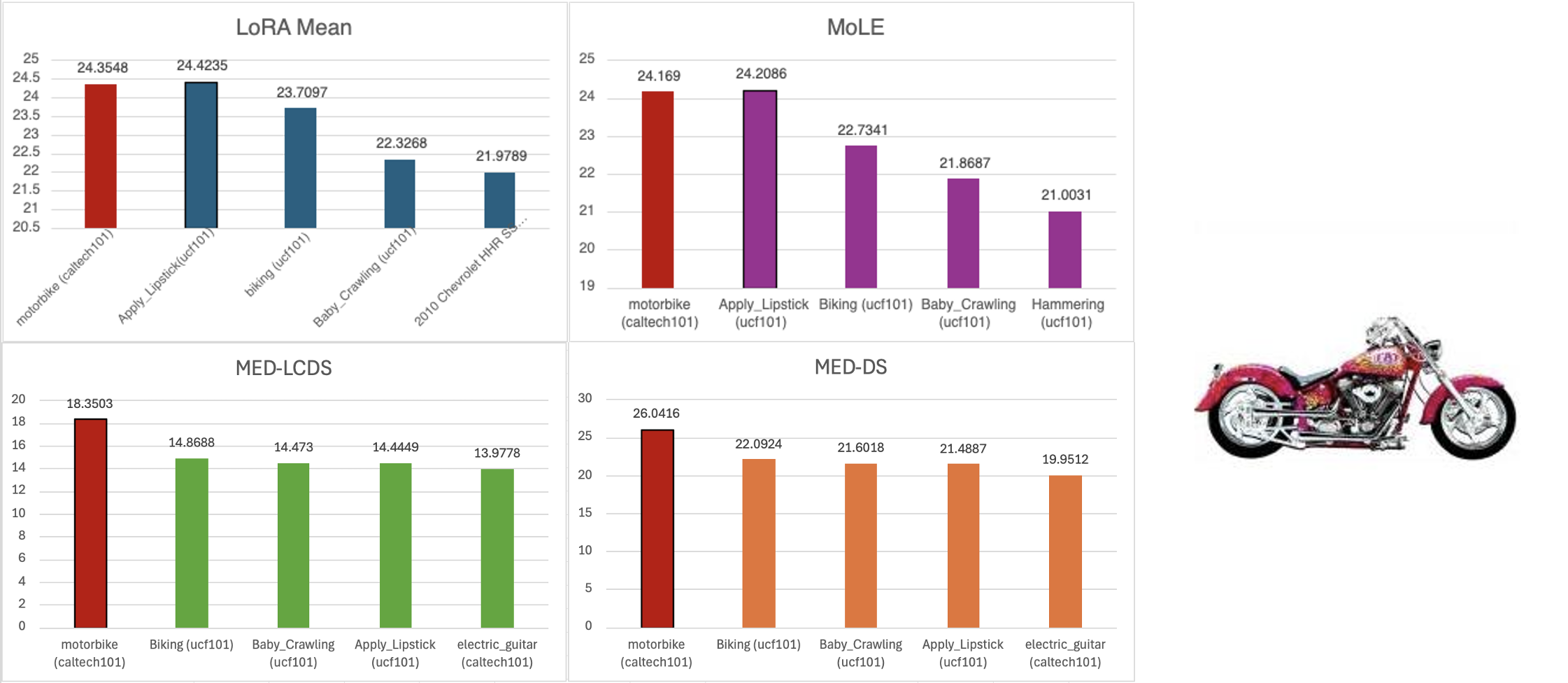}
    \caption{Comparison with other methods for an image from Motorbike (Caltech101) class. Class (domain) shown in the horizontal axis, and ground-truth (predicted) shown in red color. \ours{} outperforms other methods and in the absence of logit scaling, logits are poorly calibrated across domains, and the predicted  class can be due to a spurious logit outside the image domain.}
    \label{fig:motorbike}
\end{figure}

\begin{figure}
    \centering
    \includegraphics[width=0.8\linewidth]{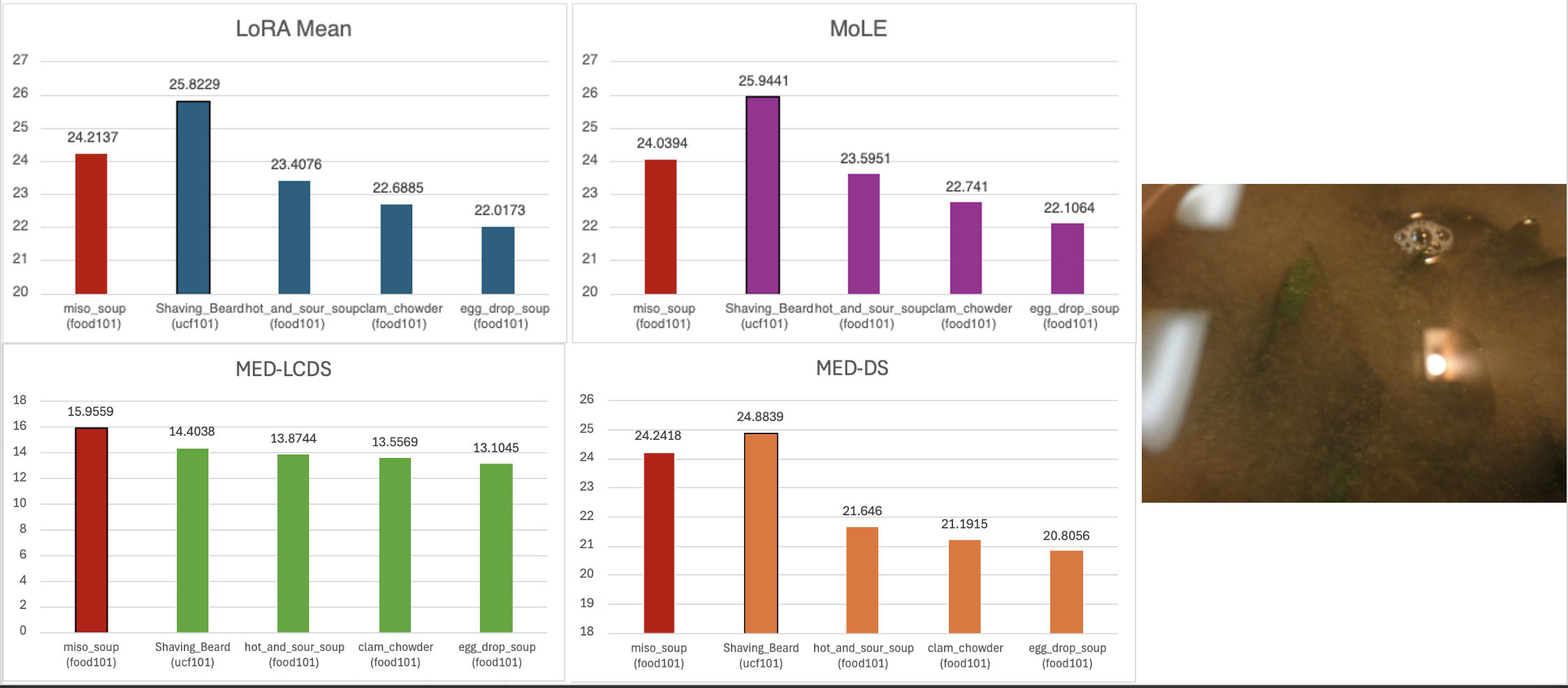}
    \caption{Comparison with other methods for an image from Miso\_soup (Food101) class. Class (domain) shown in the horizontal axis, and ground-truth (predicted) shown in red color. \ours{} outperforms other methods and in the absence of logit scaling, logits are poorly calibrated across domains, and the predicted  class can be due to a spurious logit outside the image domain.}
    \label{fig:miso_soup}
\end{figure}

\begin{figure}
    \centering
    \includegraphics[width=0.8\linewidth]{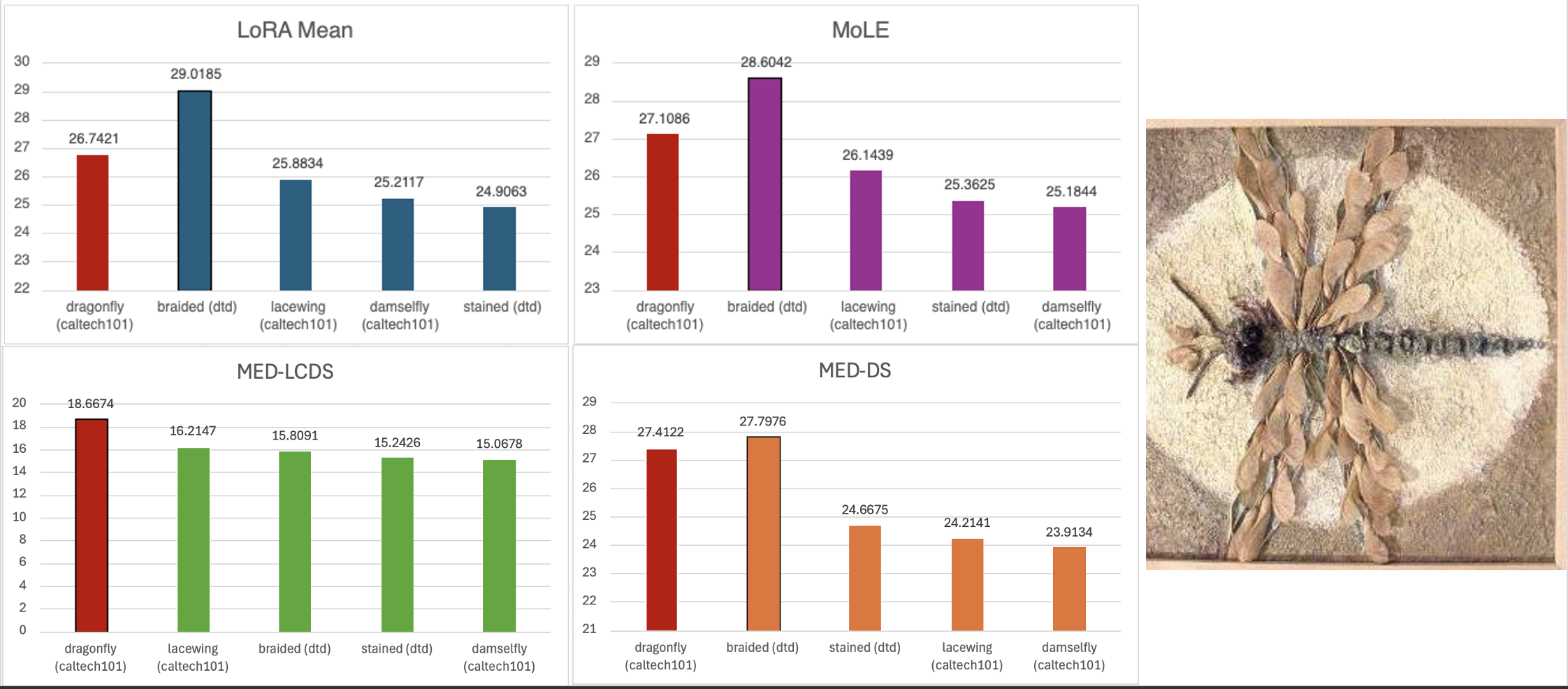}
    \caption{Comparison with other methods for an image from Dragonfly (Caltech101) class. Class (domain) shown in the horizontal axis, and ground-truth (predicted) shown in red color. \ours{} outperforms other methods and in the absence of logit scaling, logits are poorly calibrated across domains, and the predicted  class can be due to a spurious logit outside the image domain.}
    \label{fig:dragonfly}
\end{figure}

\begin{figure}
    \centering
    \includegraphics[width=0.8\linewidth]{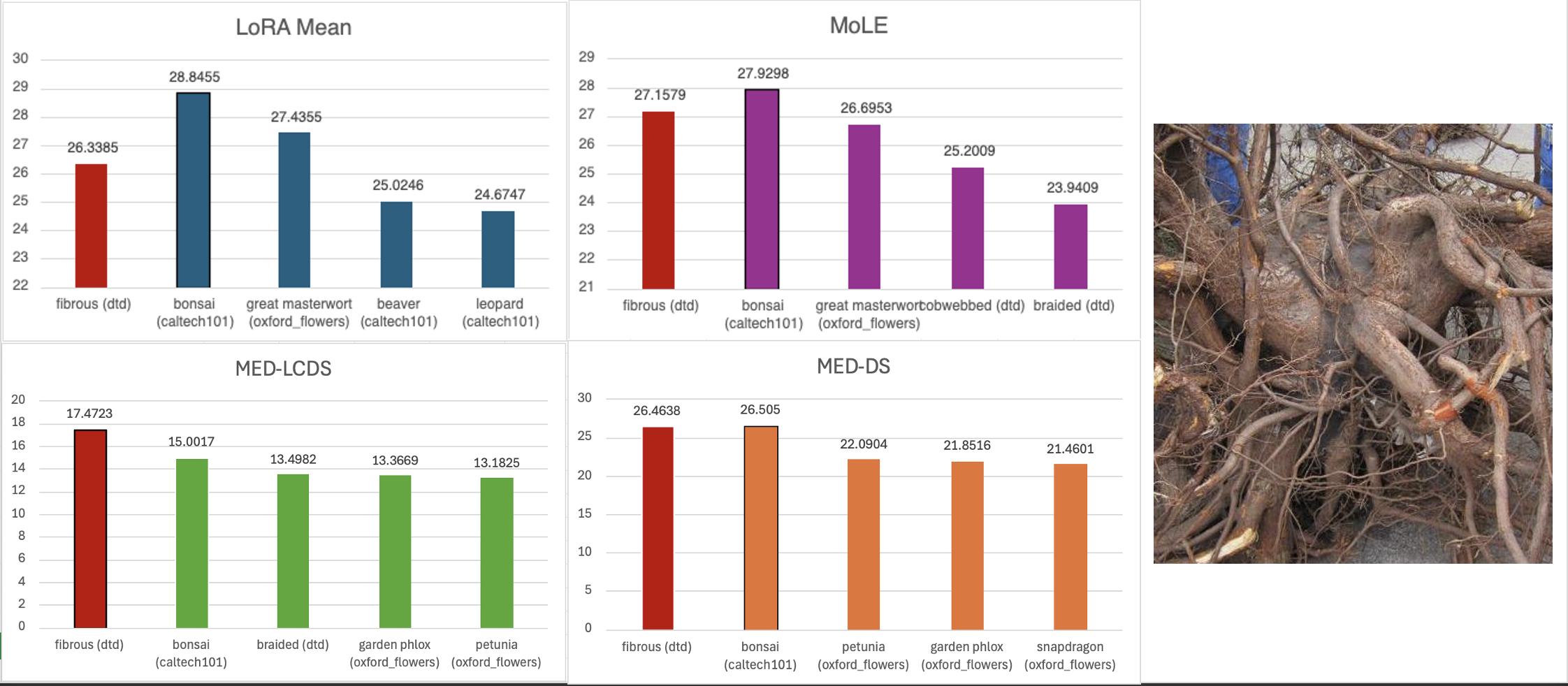}
    \caption{Comparison with other methods for an image from Fibrous (DTD) class. Class (domain) shown in the horizontal axis, and ground-truth (predicted) shown in red color. \ours{} outperforms other methods and in the absence of logit scaling, logits are poorly calibrated across domains, and the predicted  class can be due to a spurious logit outside the image domain.}
    \label{fig:fibrous}
\end{figure}

From Figure~\ref{fig:motorbike} to Figure~\ref{fig:fibrous}, we show 4 examples of the cross-domain interference phenomenon. The plots visualize the top-5 class logits predicted by LoRA-Mean, MoLE and our methods for several example images. For each method, the bar chart shows the logit scores assigned to different candidate classes.
\begin{itemize}
    \item \textbf{Red bar}: logit of the ground-truth class
    \item \textbf{Other colored bars}: logits of competing classes predicted by the model
\end{itemize}

These visualizations highlight how effectively each method separates the correct class from competing alternatives. In several cases, the highest competing classes originate from a \emph{different domain} than the input image, revealing the presence of cross-domain interference.
For example, in Figure~\ref{fig:miso_soup}, when the input image corresponds to \textit{miso\_soup}, the unrelated class \textit{shaving beard} receives a high logit score. Both \textit{LoRA-Mean} and \textit{MoLE} assign relatively large logits to this incorrect class, indicating poor cross-domain calibration. Incorporating domain-aware training improves the situation: \textit{MED w/o scale} reduces the gap between the ground-truth class and the competing class, although \textit{shaving beard} still remains dominant. In contrast, when \ours{} is trained with the proposed logit calibration, the model successfully suppresses the cross-domain class and assigns the highest logit to \textit{miso\_soup}, the correct label.

Overall, these examples demonstrate how cross-domain interference can lead to incorrect high-confidence predictions and how the proposed domain-wise logit scaling in MED effectively mitigates this issue.

\section{Gating Visualization}
\begin{figure*}[htbp]
    \centering

    \begin{subfigure}[b]{0.3\linewidth}
        \centering
        \includegraphics[width=\linewidth]{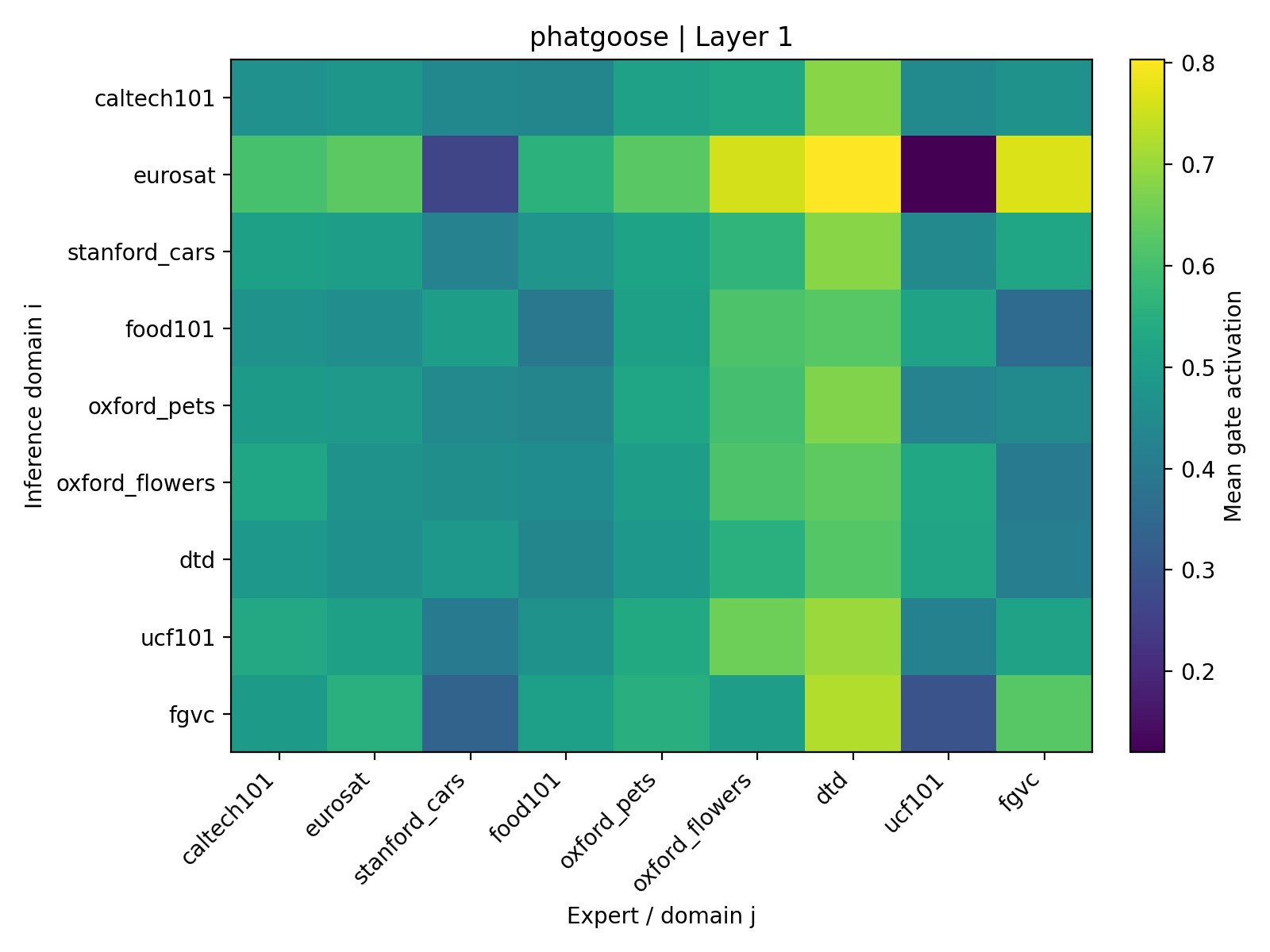}
        \caption{Layer 1 | Phatgoose}
    \end{subfigure}
    \hfill
    \begin{subfigure}[b]{0.3\linewidth}
        \centering
        \includegraphics[width=\linewidth]{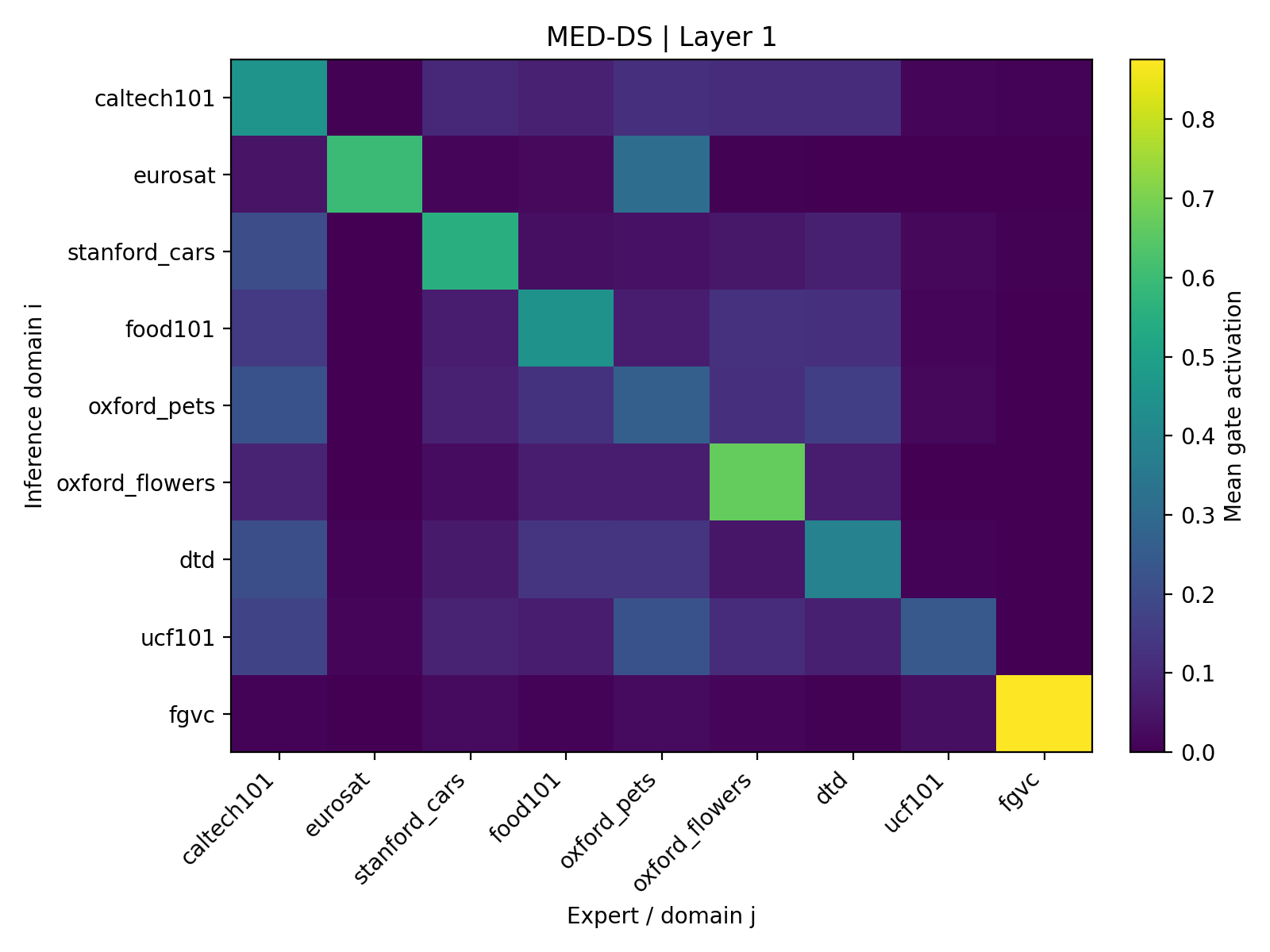}
        \caption{Layer 1 | \dsours{}}
    \end{subfigure}
    \hfill
    \begin{subfigure}[b]{0.3\linewidth}
        \centering
        \includegraphics[width=\linewidth]{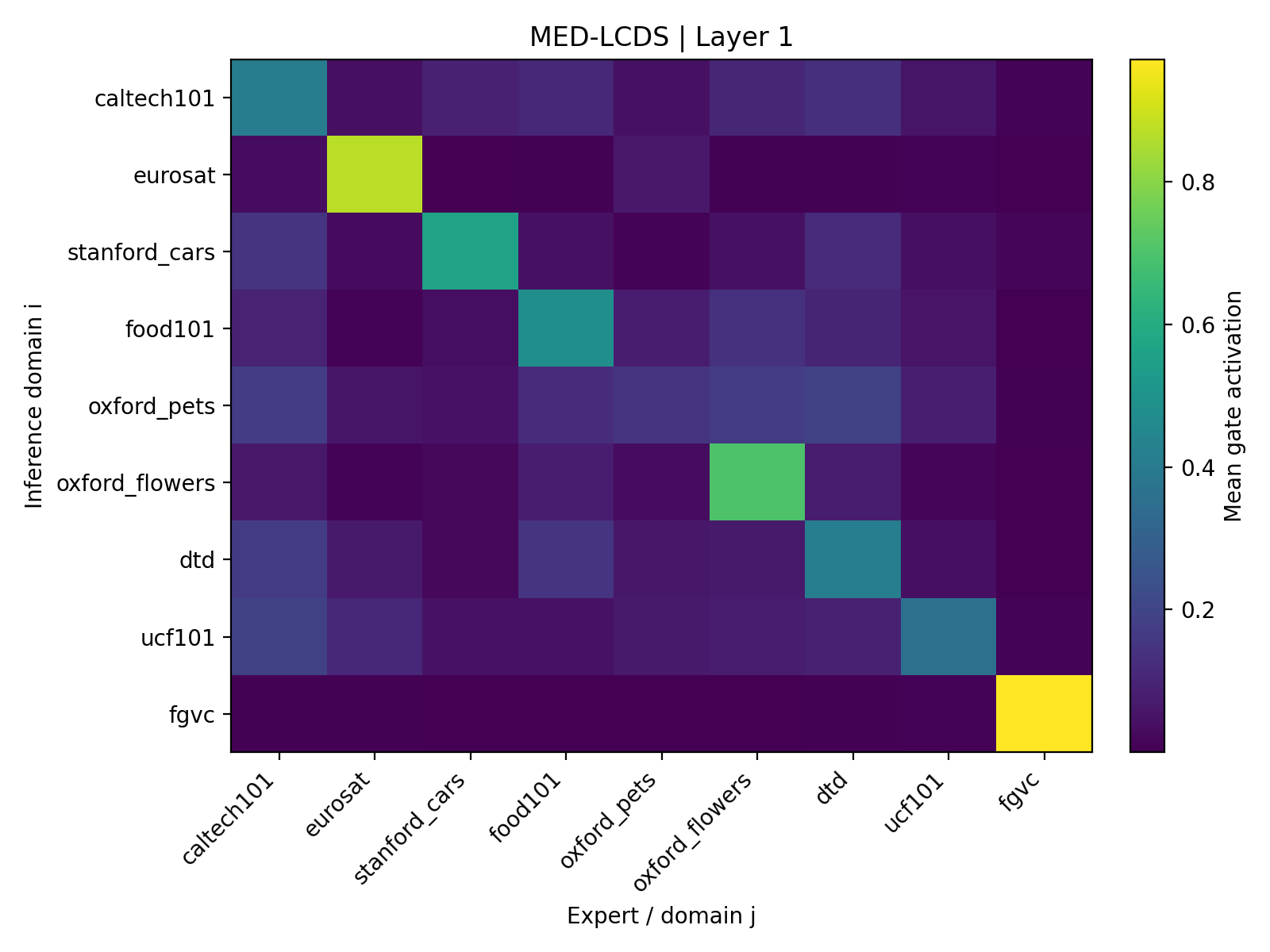}
        \caption{Layer 1 | \ours{}}
    \end{subfigure}

    \vspace{0.5em}

    \begin{subfigure}[b]{0.3\linewidth}
        \centering
        \includegraphics[width=\linewidth]{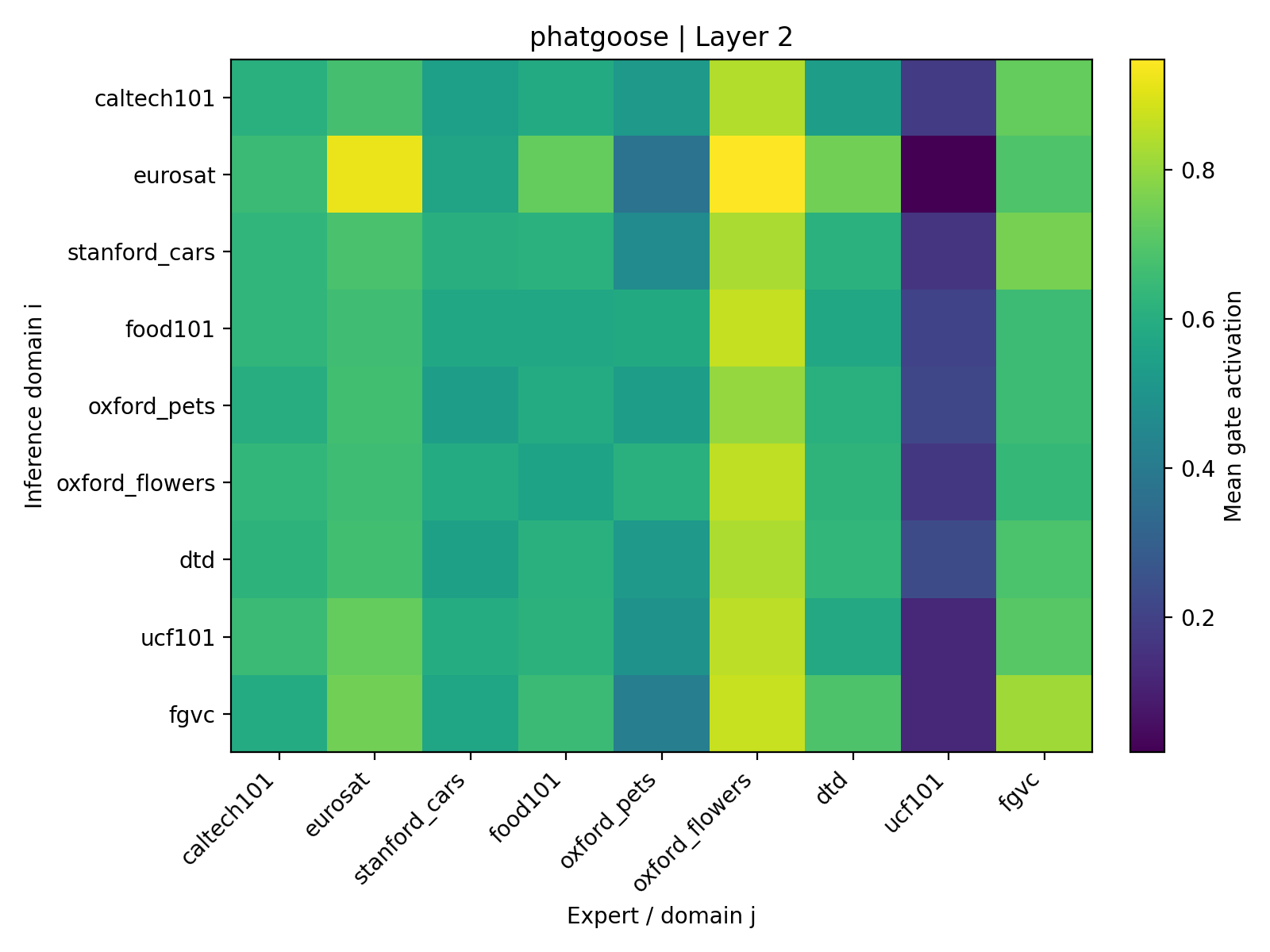}
        \caption{Layer 2 | Phatgoose}
    \end{subfigure}
    \hfill
    \begin{subfigure}[b]{0.3\linewidth}
        \centering
        \includegraphics[width=\linewidth]{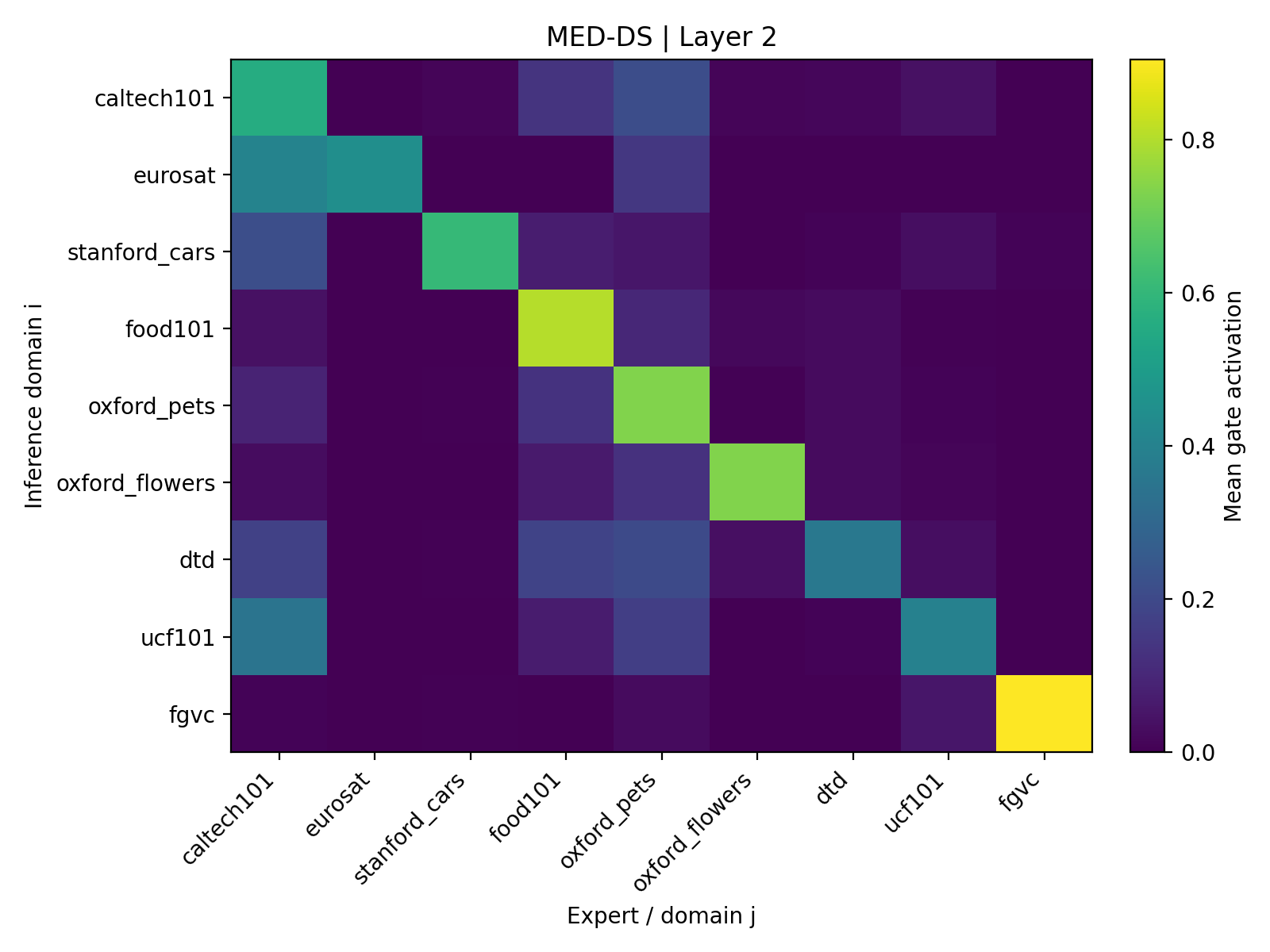}
        \caption{Layer 2 | \dsours{}}
    \end{subfigure}
    \hfill
    \begin{subfigure}[b]{0.3\linewidth}
        \centering
        \includegraphics[width=\linewidth]{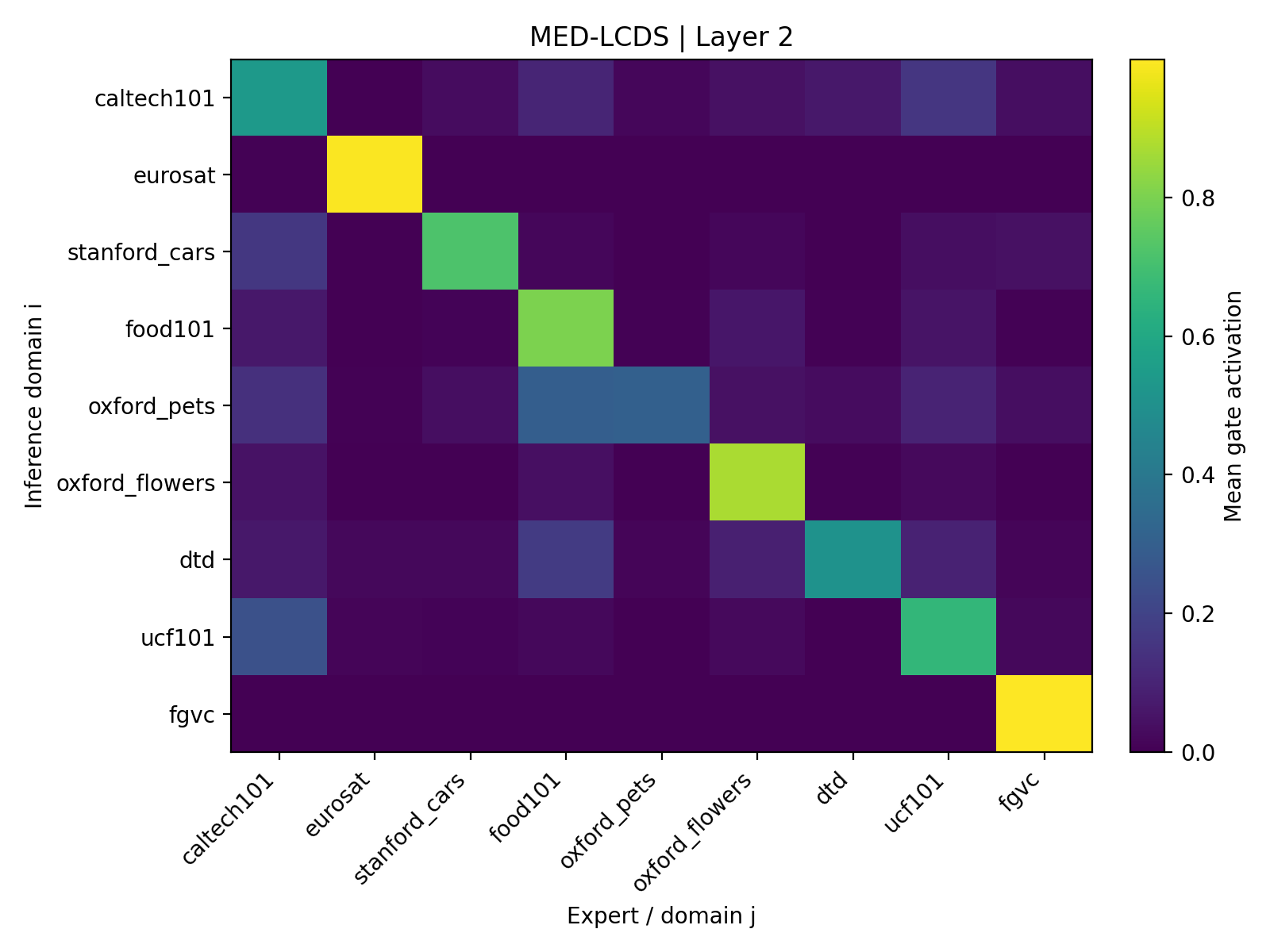}
        \caption{Layer 2 | \ours{}}
    \end{subfigure}

    \vspace{0.5em}

    \begin{subfigure}[b]{0.3\linewidth}
        \centering
        \includegraphics[width=\linewidth]{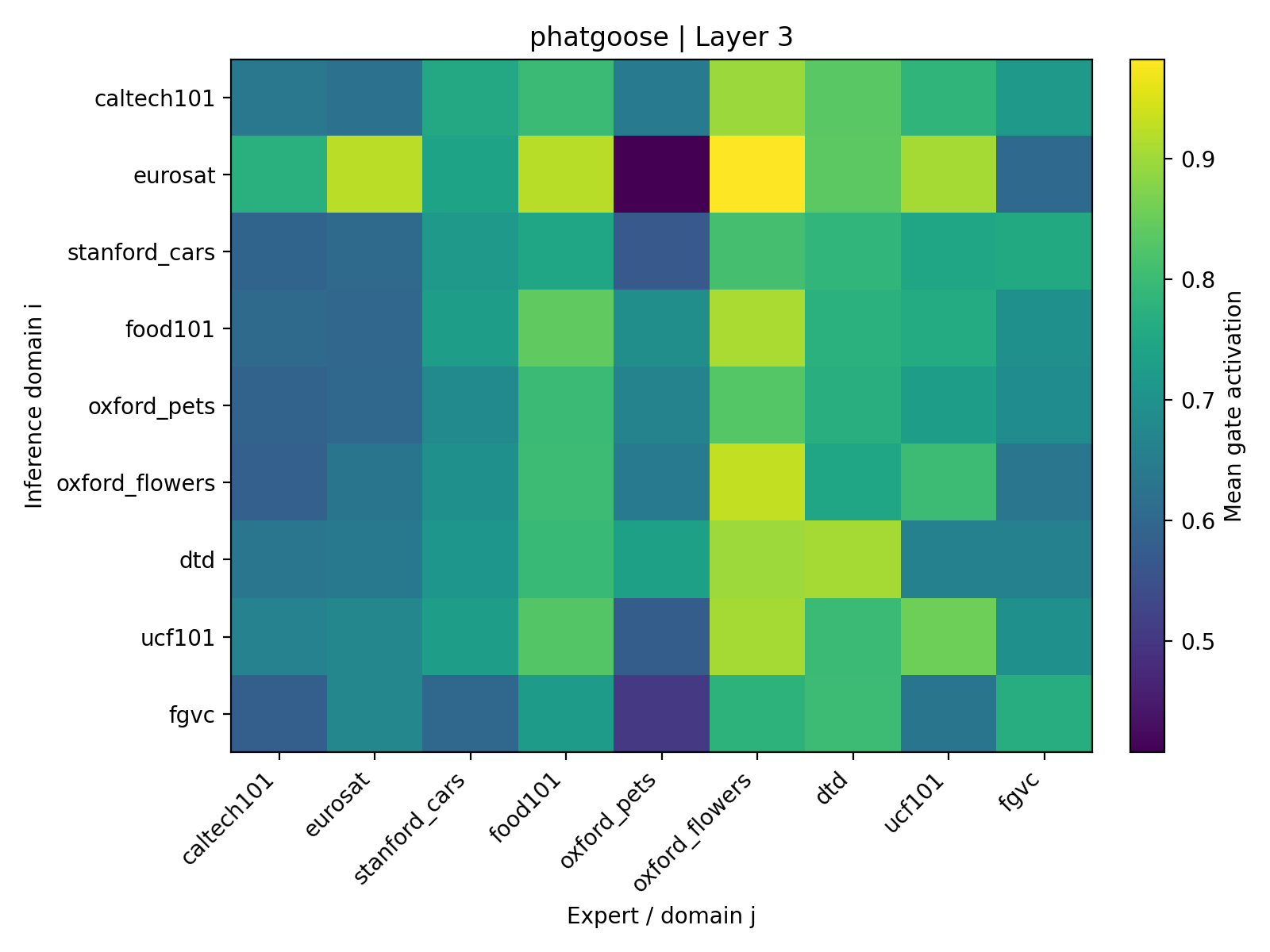}
        \caption{Layer 3 | Phatgoose}
    \end{subfigure}
    \hfill
    \begin{subfigure}[b]{0.3\linewidth}
        \centering
        \includegraphics[width=\linewidth]{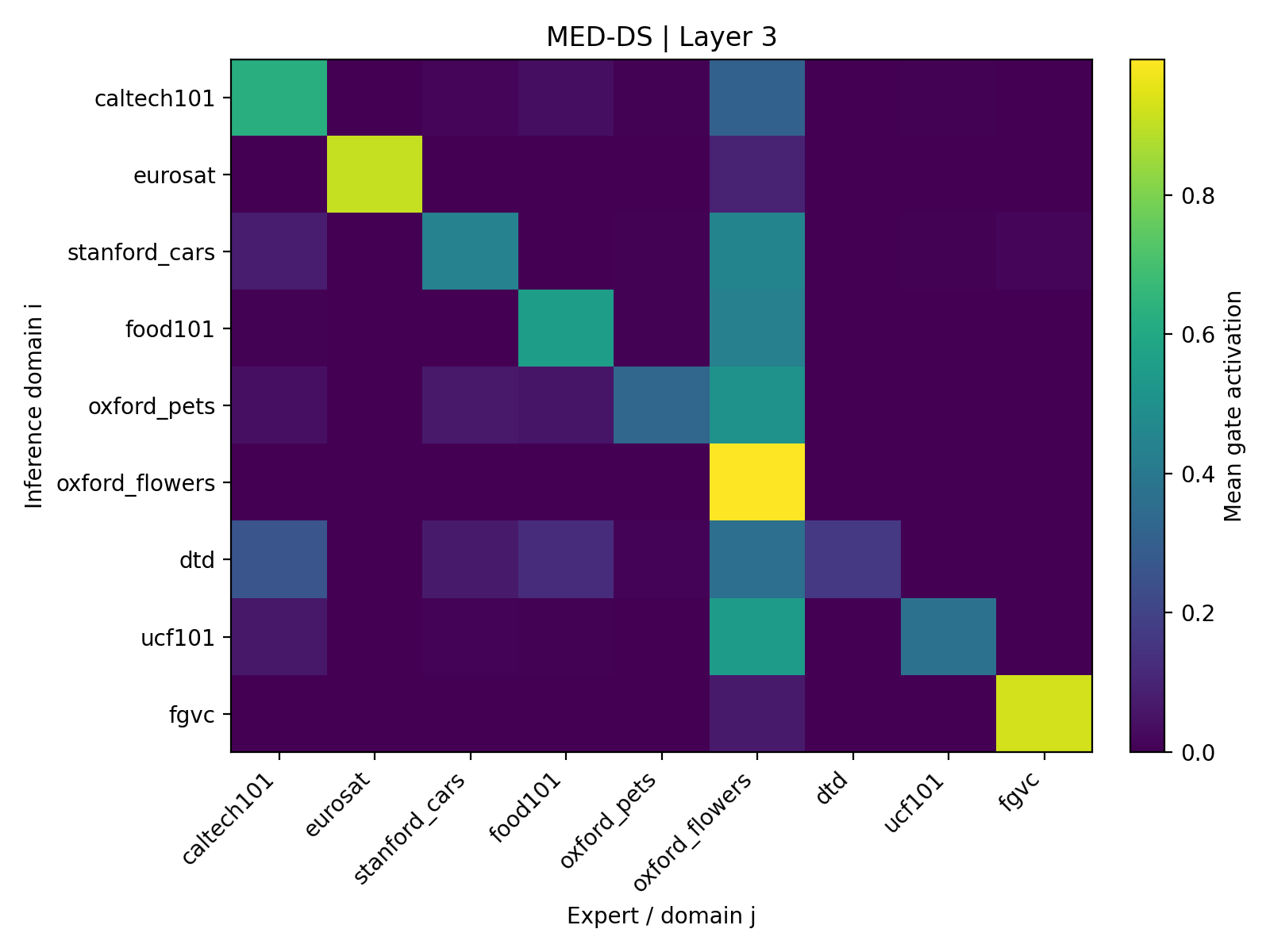}
        \caption{Layer 3 | \dsours{}}
    \end{subfigure}
    \hfill
    \begin{subfigure}[b]{0.3\linewidth}
        \centering
        \includegraphics[width=\linewidth]{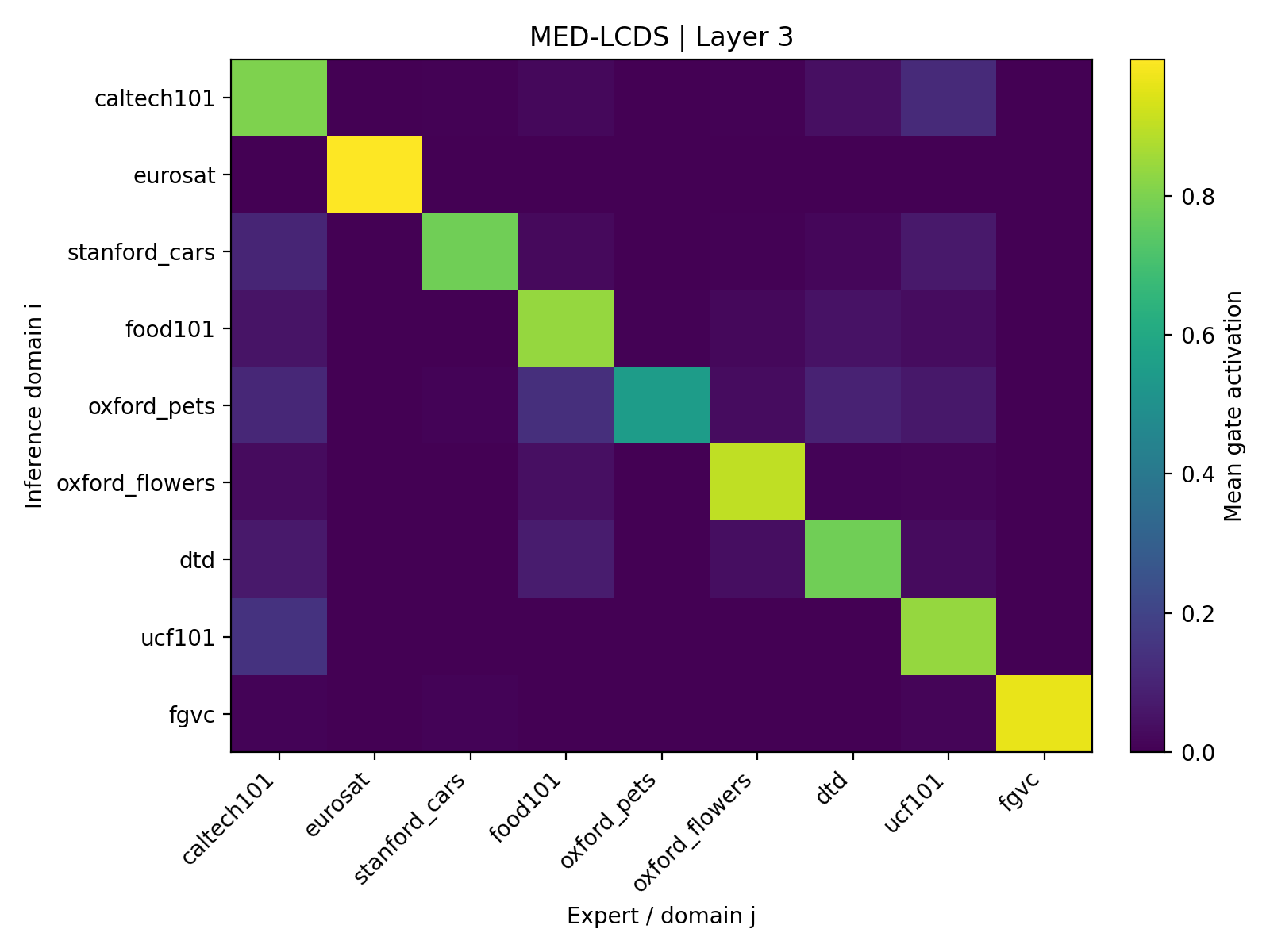}
        \caption{Layer 3 | \ours{}}
    \end{subfigure}

    \vspace{0.5em}

    \begin{subfigure}[b]{0.3\linewidth}
        \centering
        \includegraphics[width=\linewidth]{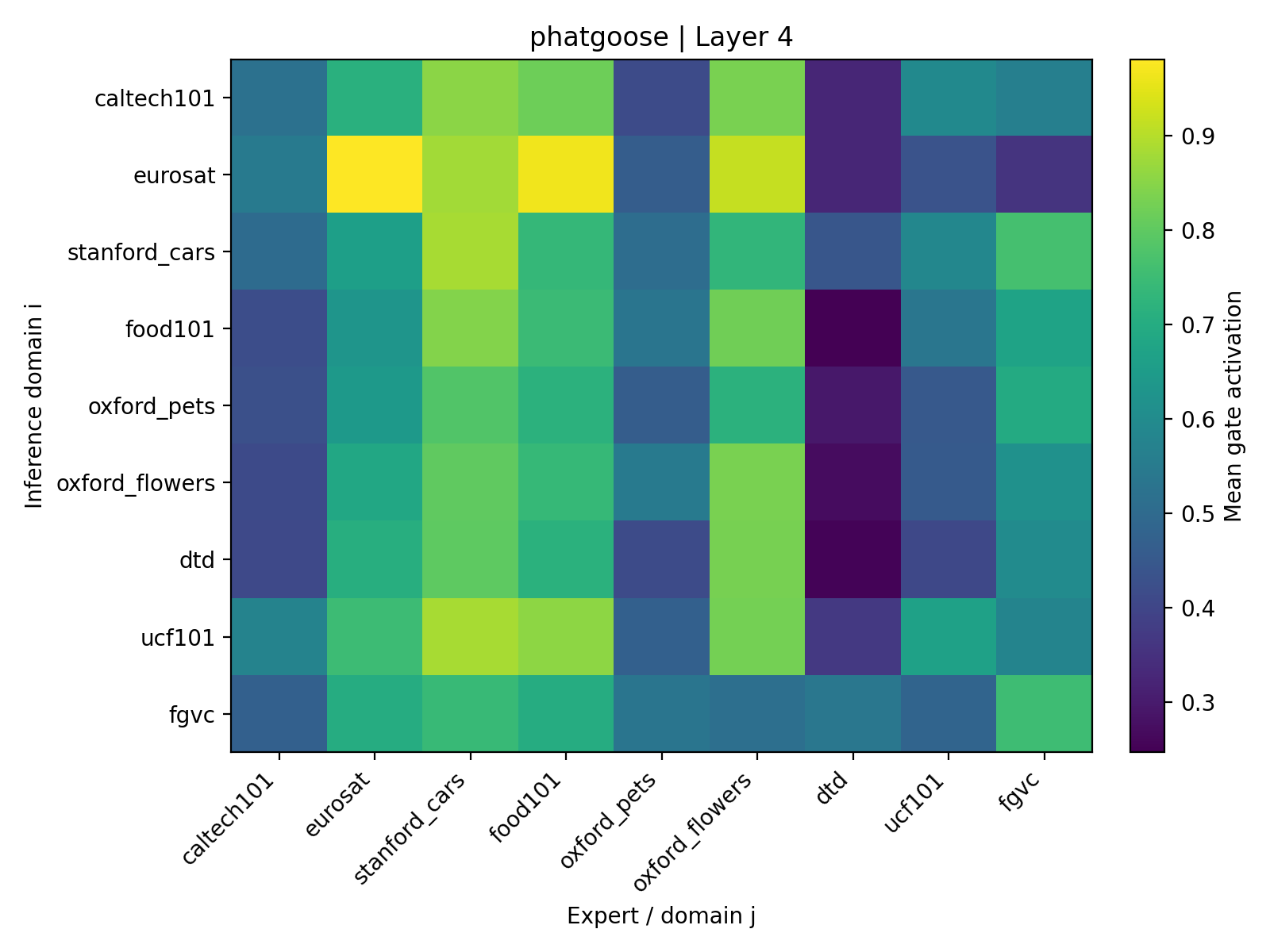}
        \caption{Layer 4 | Phatgoose}
    \end{subfigure}
    \hfill
    \begin{subfigure}[b]{0.3\linewidth}
        \centering
        \includegraphics[width=\linewidth]{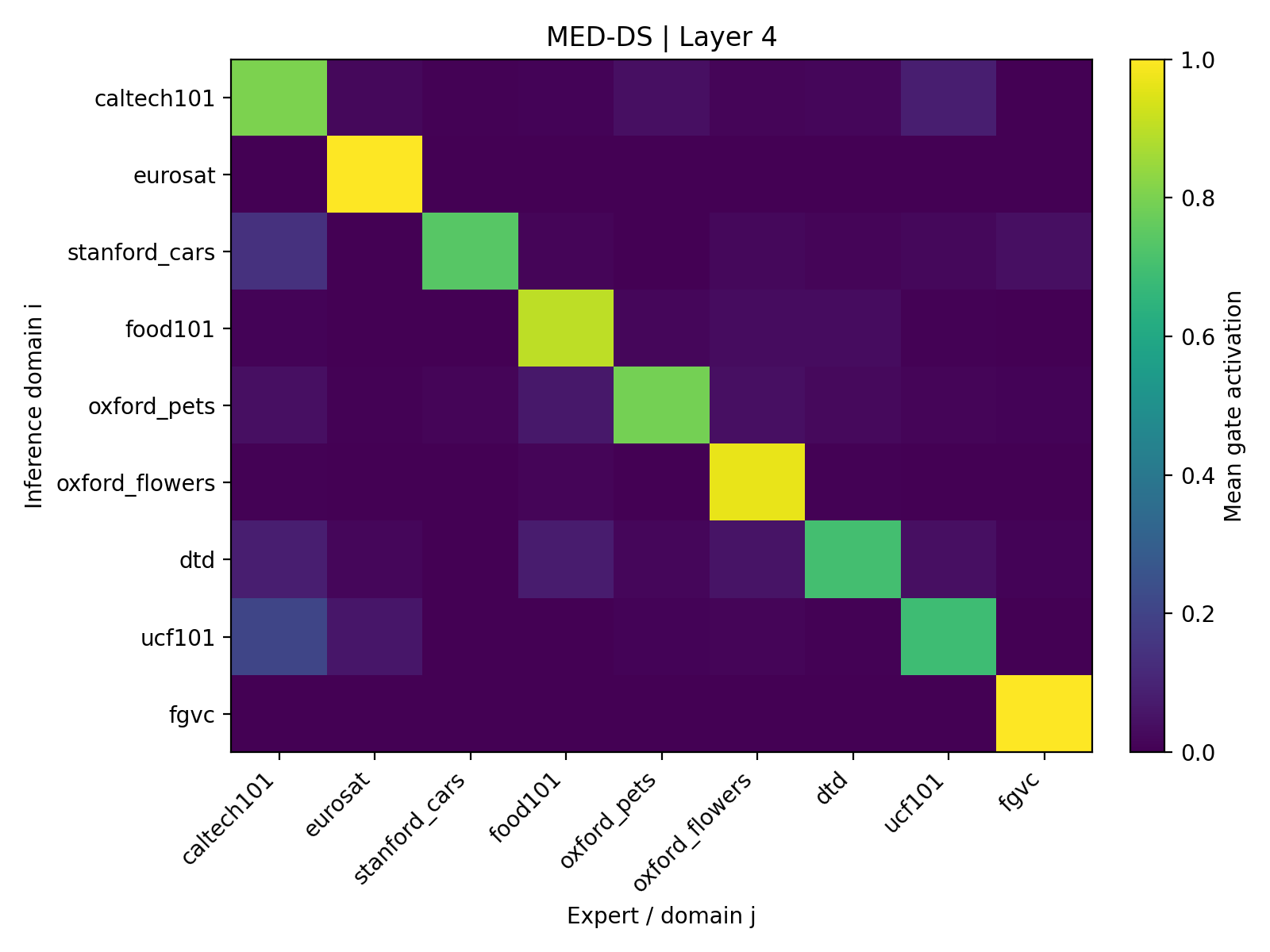}
        \caption{Layer 4 | \dsours{}}
    \end{subfigure}
    \hfill
    \begin{subfigure}[b]{0.3\linewidth}
        \centering
        \includegraphics[width=\linewidth]{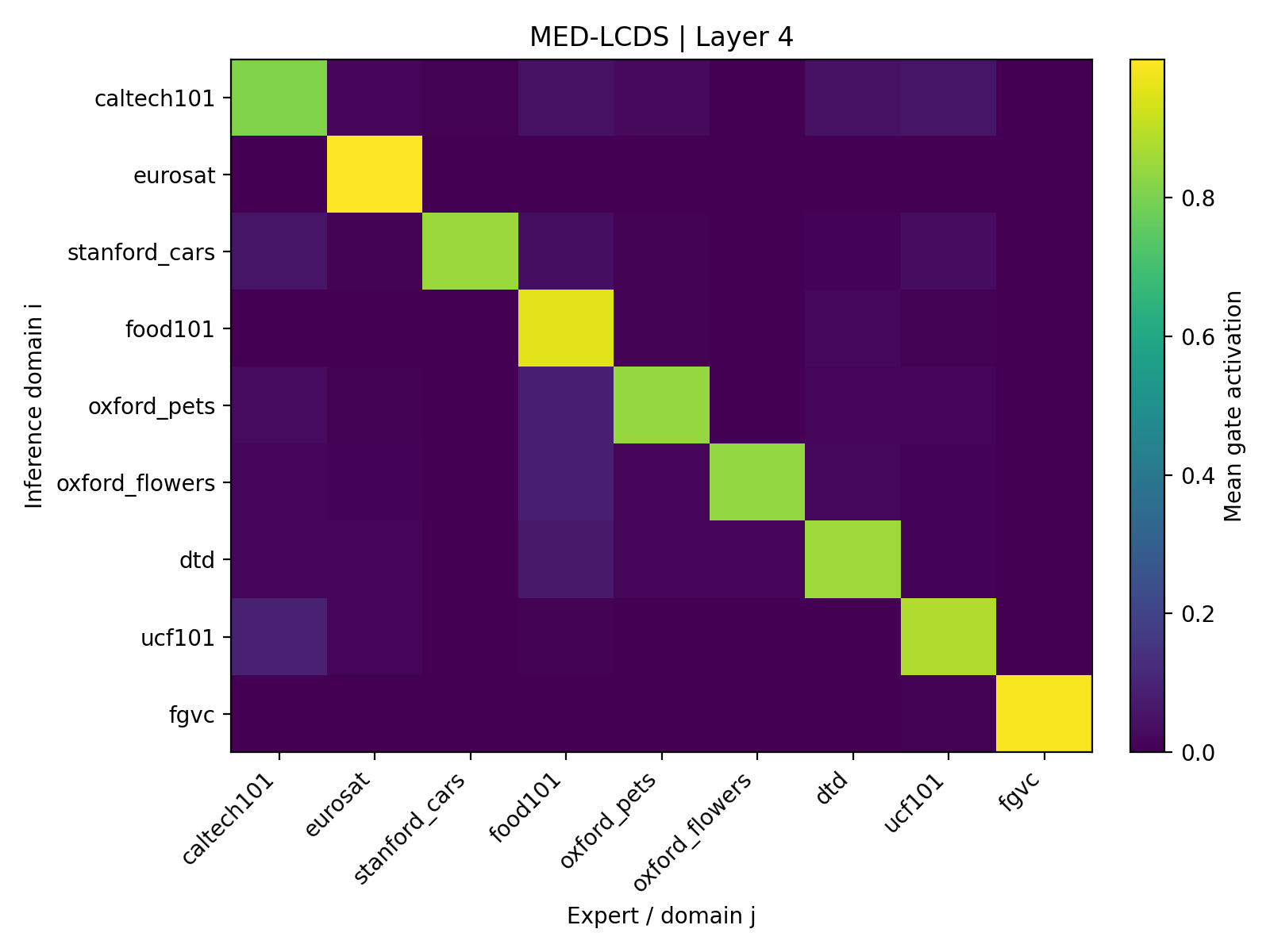}
        \caption{Layer 4 | \ours{}}
    \end{subfigure}

    \caption{Visualization of the gating at image encoder layers 1--4, where rows denote layers, columns denote methods, and color indicates the mean gating activation.}
    \label{fig:image_layers_1_4}
\end{figure*}

\begin{figure*}[htbp]
    \centering

    \begin{subfigure}[b]{0.3\linewidth}
        \centering
        \includegraphics[width=\linewidth]{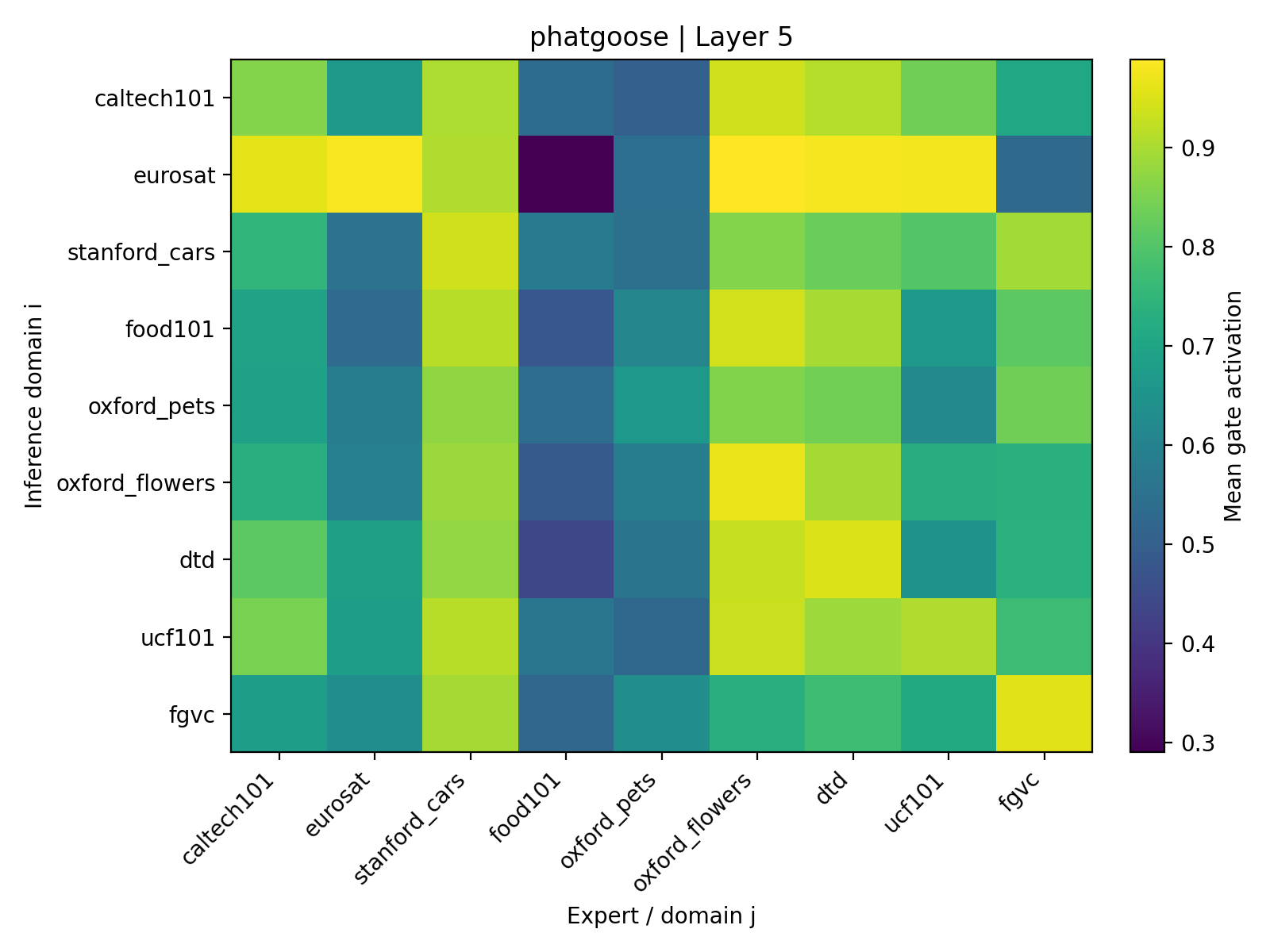}
        \caption{Layer 5 | Phatgoose}
    \end{subfigure}
    \hfill
    \begin{subfigure}[b]{0.3\linewidth}
        \centering
        \includegraphics[width=\linewidth]{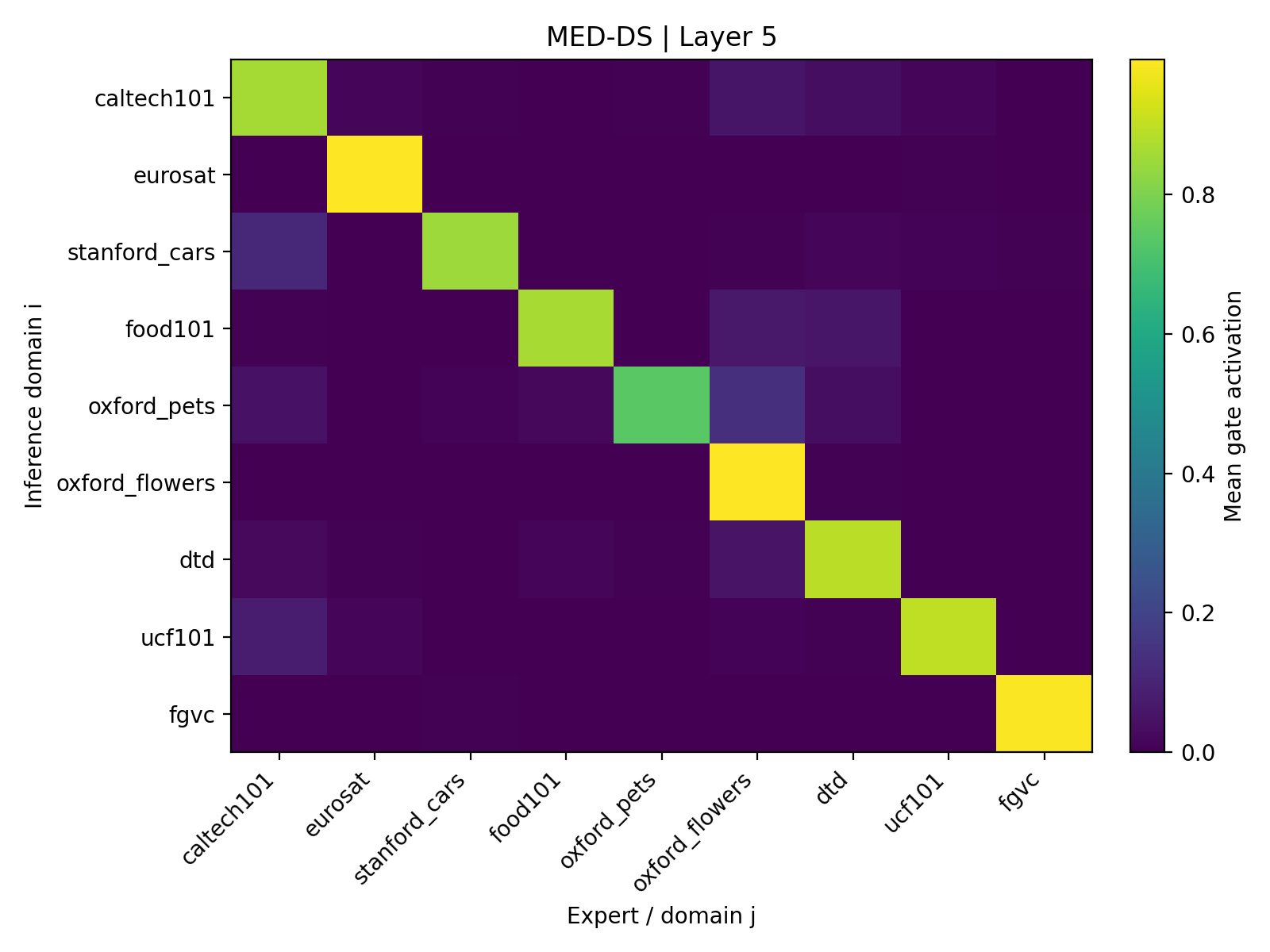}
        \caption{Layer 5 | \dsours{}}
    \end{subfigure}
    \hfill
    \begin{subfigure}[b]{0.3\linewidth}
        \centering
        \includegraphics[width=\linewidth]{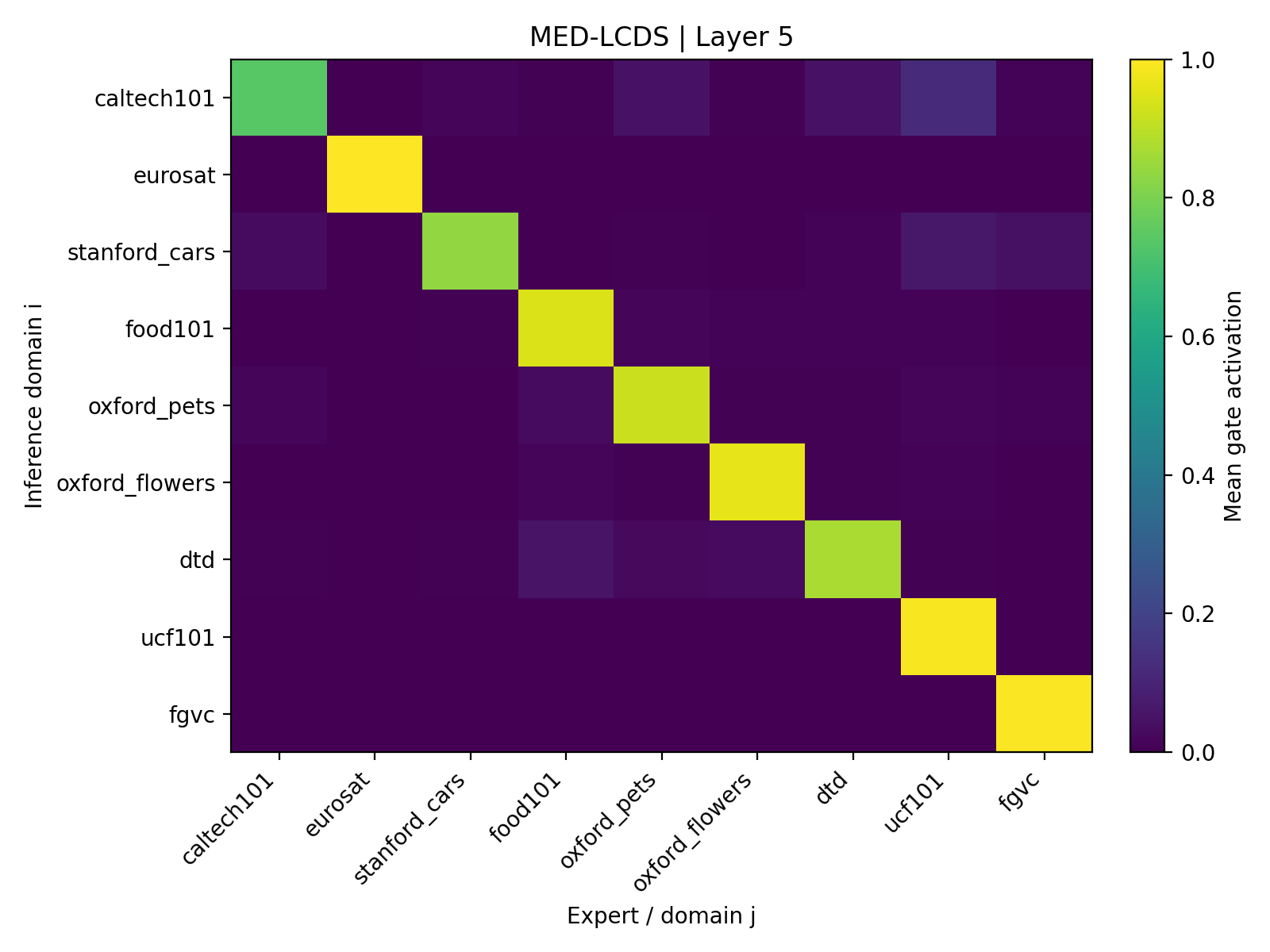}
        \caption{Layer 5 | \ours{}}
    \end{subfigure}

    \vspace{0.5em}

    \begin{subfigure}[b]{0.3\linewidth}
        \centering
        \includegraphics[width=\linewidth]{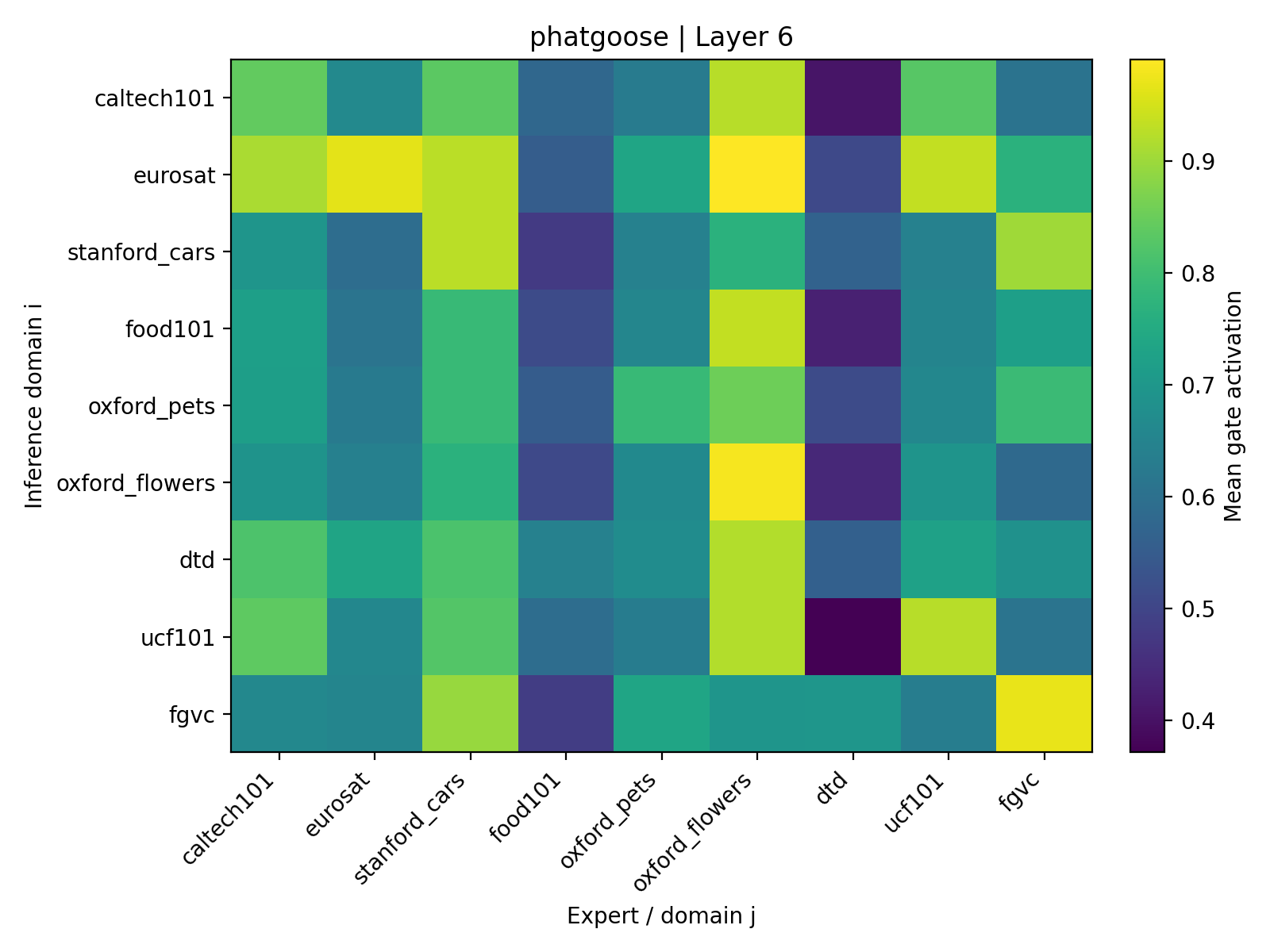}
        \caption{Layer 6 | Phatgoose}
    \end{subfigure}
    \hfill
    \begin{subfigure}[b]{0.3\linewidth}
        \centering
        \includegraphics[width=\linewidth]{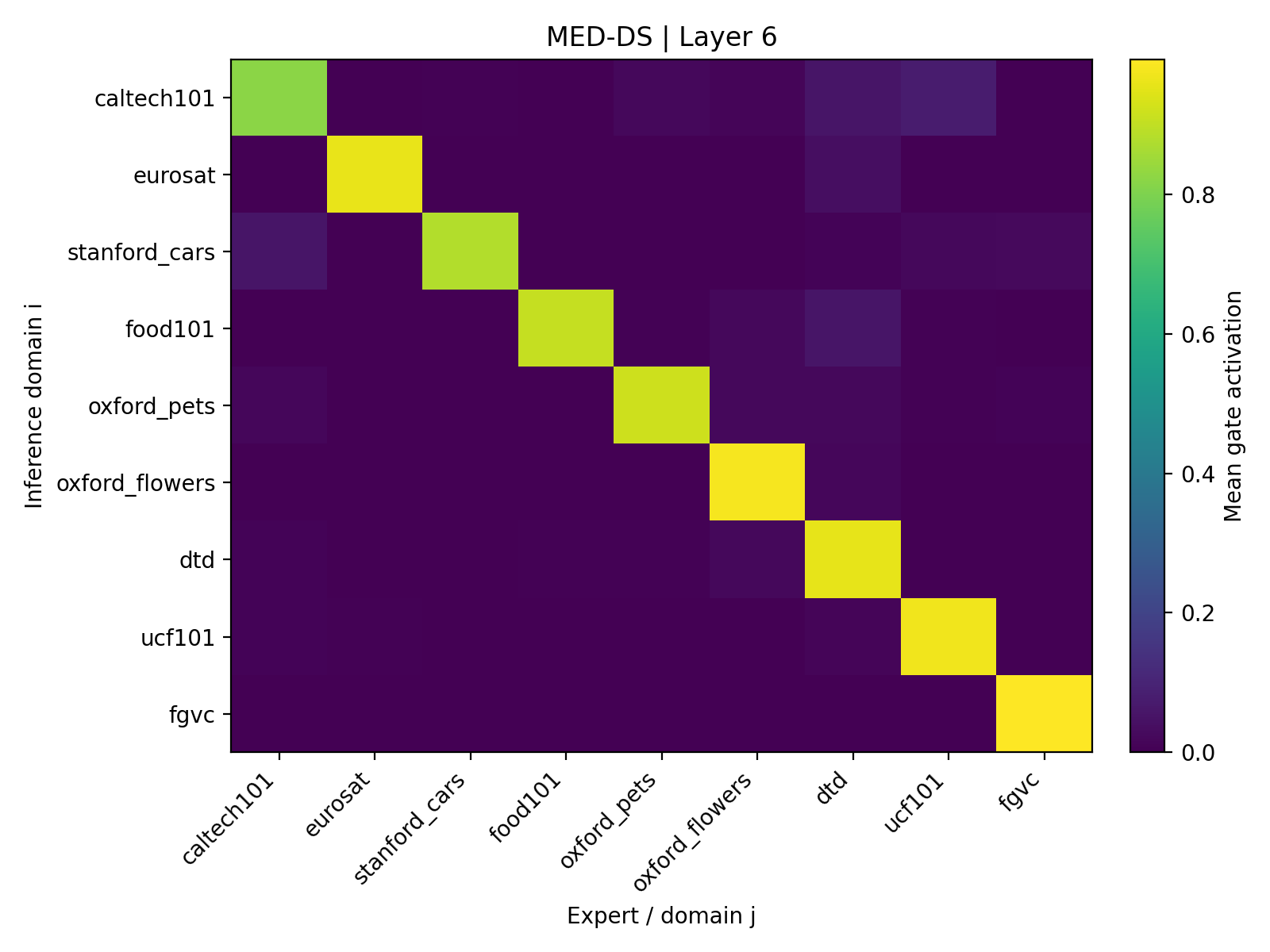}
        \caption{Layer 6 | \dsours{}}
    \end{subfigure}
    \hfill
    \begin{subfigure}[b]{0.3\linewidth}
        \centering
        \includegraphics[width=\linewidth]{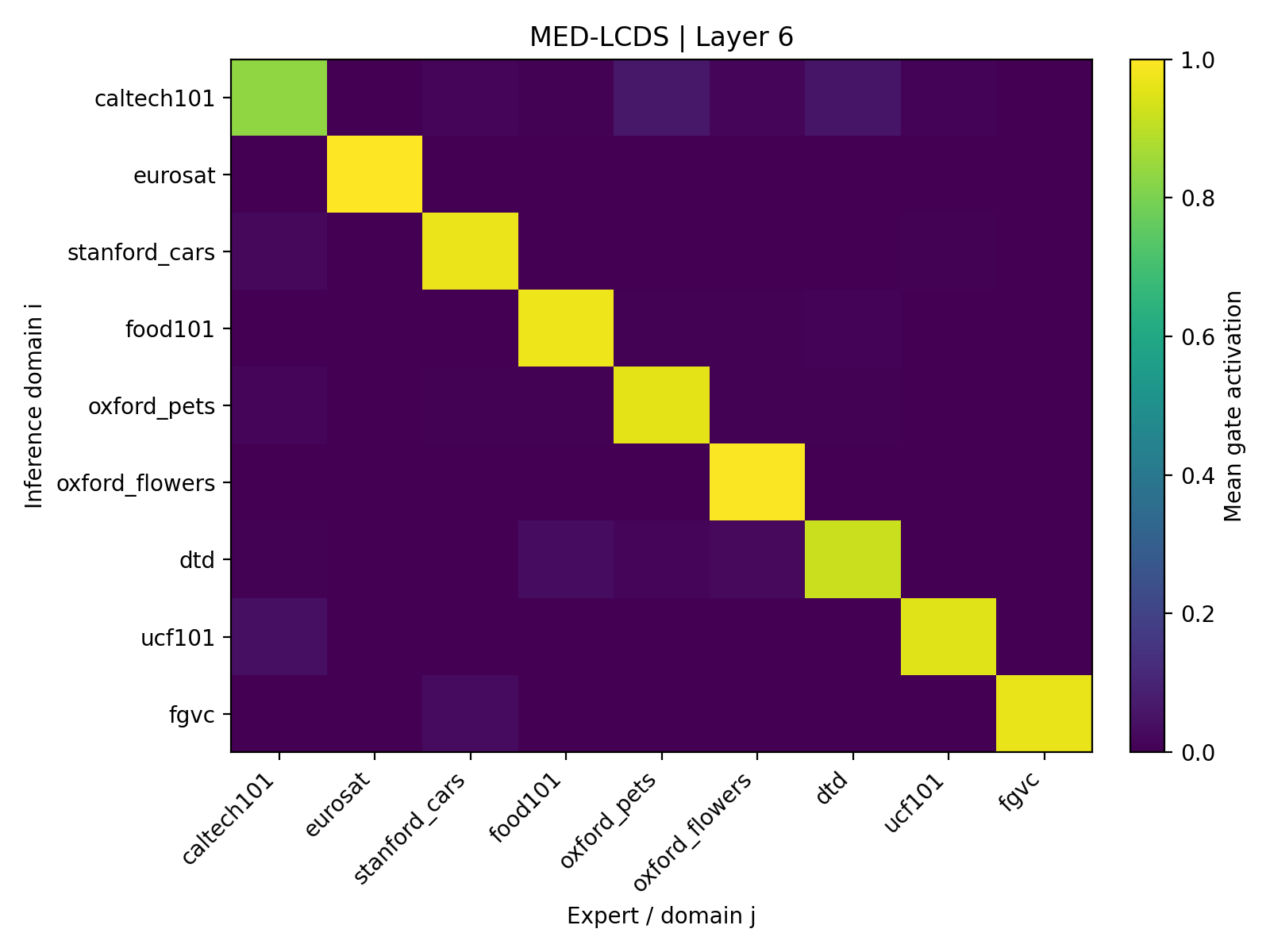}
        \caption{Layer 6 | \ours{}}
    \end{subfigure}

    \vspace{0.5em}

    \begin{subfigure}[b]{0.3\linewidth}
        \centering
        \includegraphics[width=\linewidth]{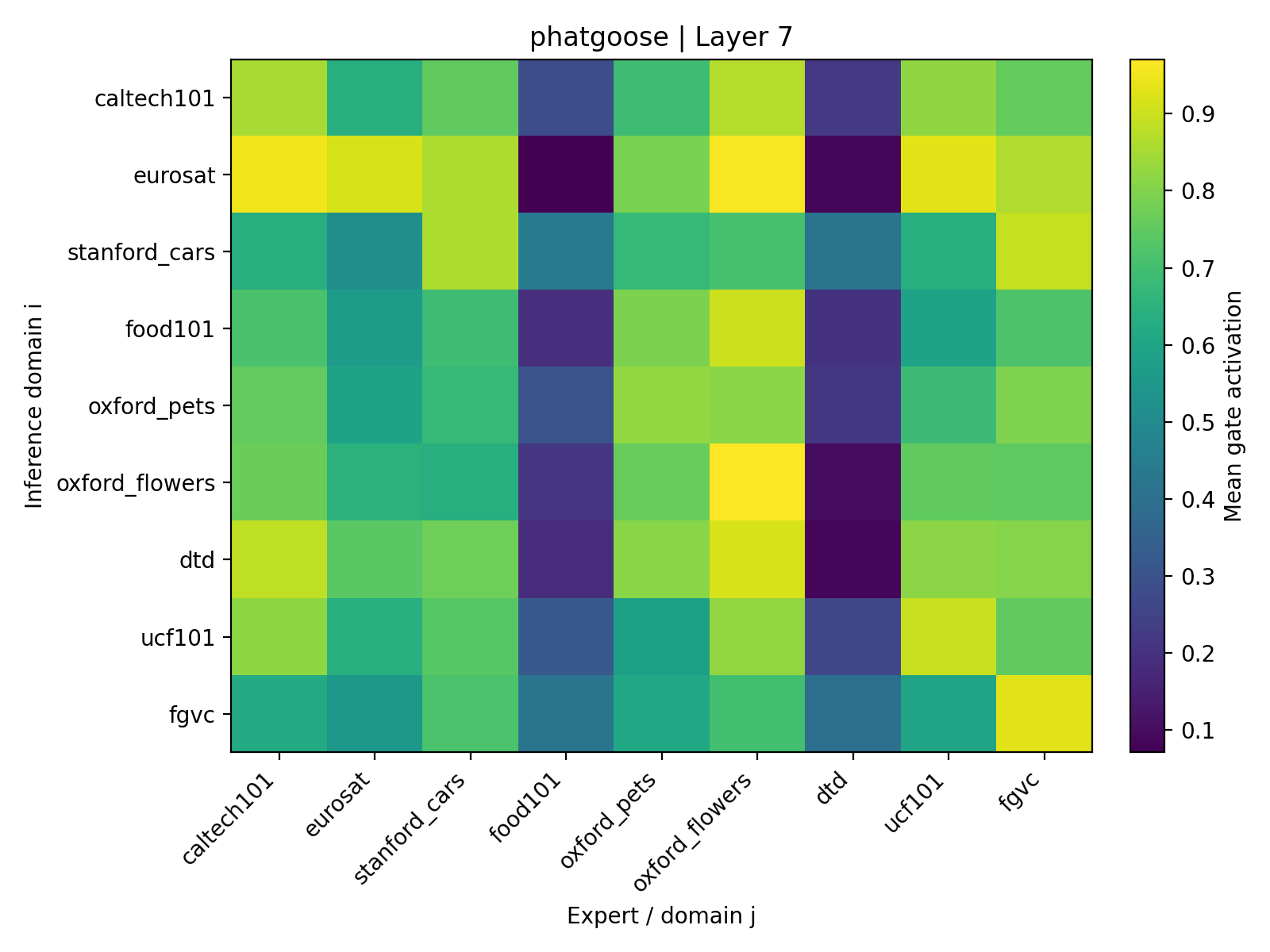}
        \caption{Layer 7 | Phatgoose}
    \end{subfigure}
    \hfill
    \begin{subfigure}[b]{0.3\linewidth}
        \centering
        \includegraphics[width=\linewidth]{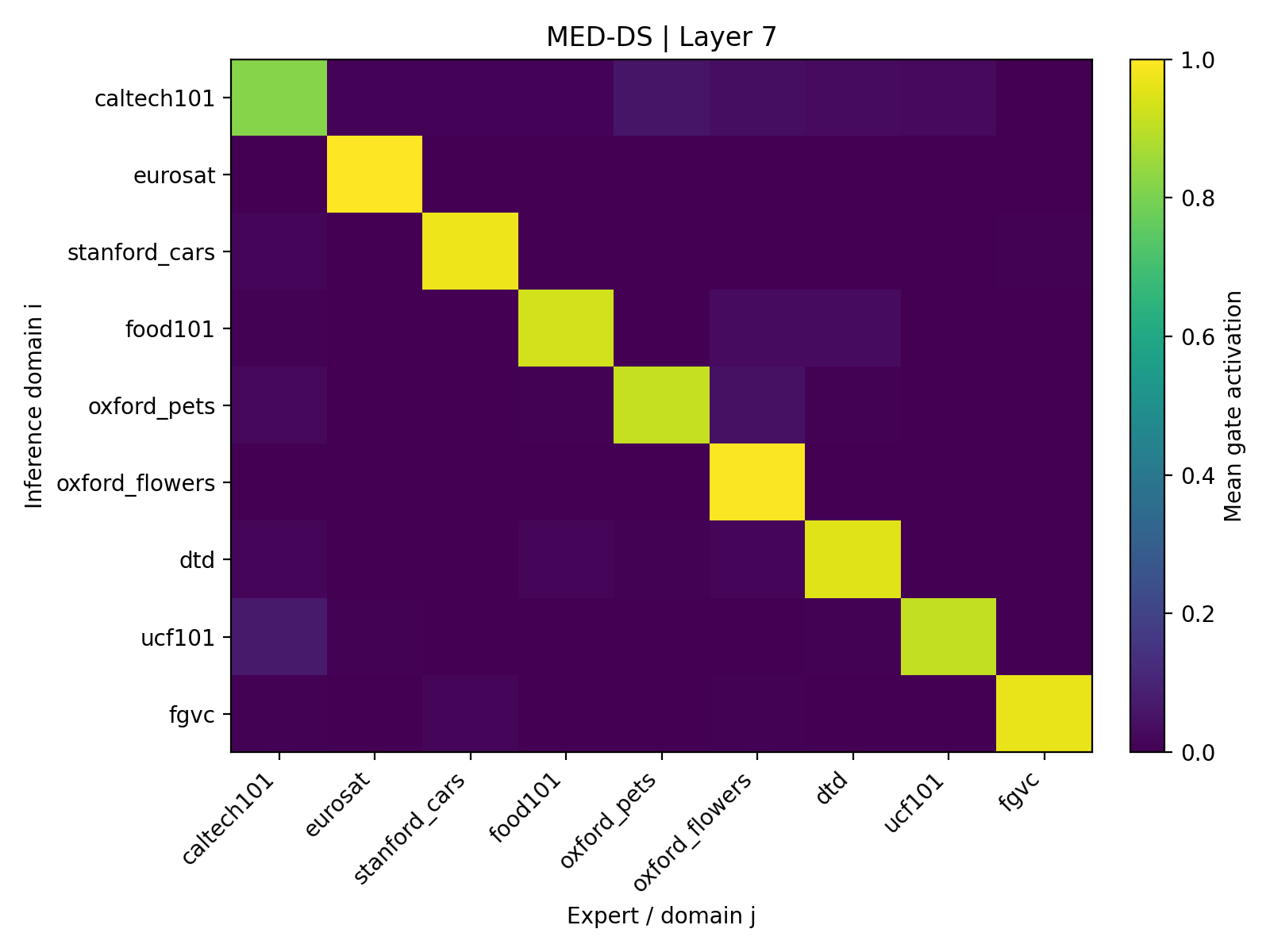}
        \caption{Layer 7 | \dsours{}}
    \end{subfigure}
    \hfill
    \begin{subfigure}[b]{0.3\linewidth}
        \centering
        \includegraphics[width=\linewidth]{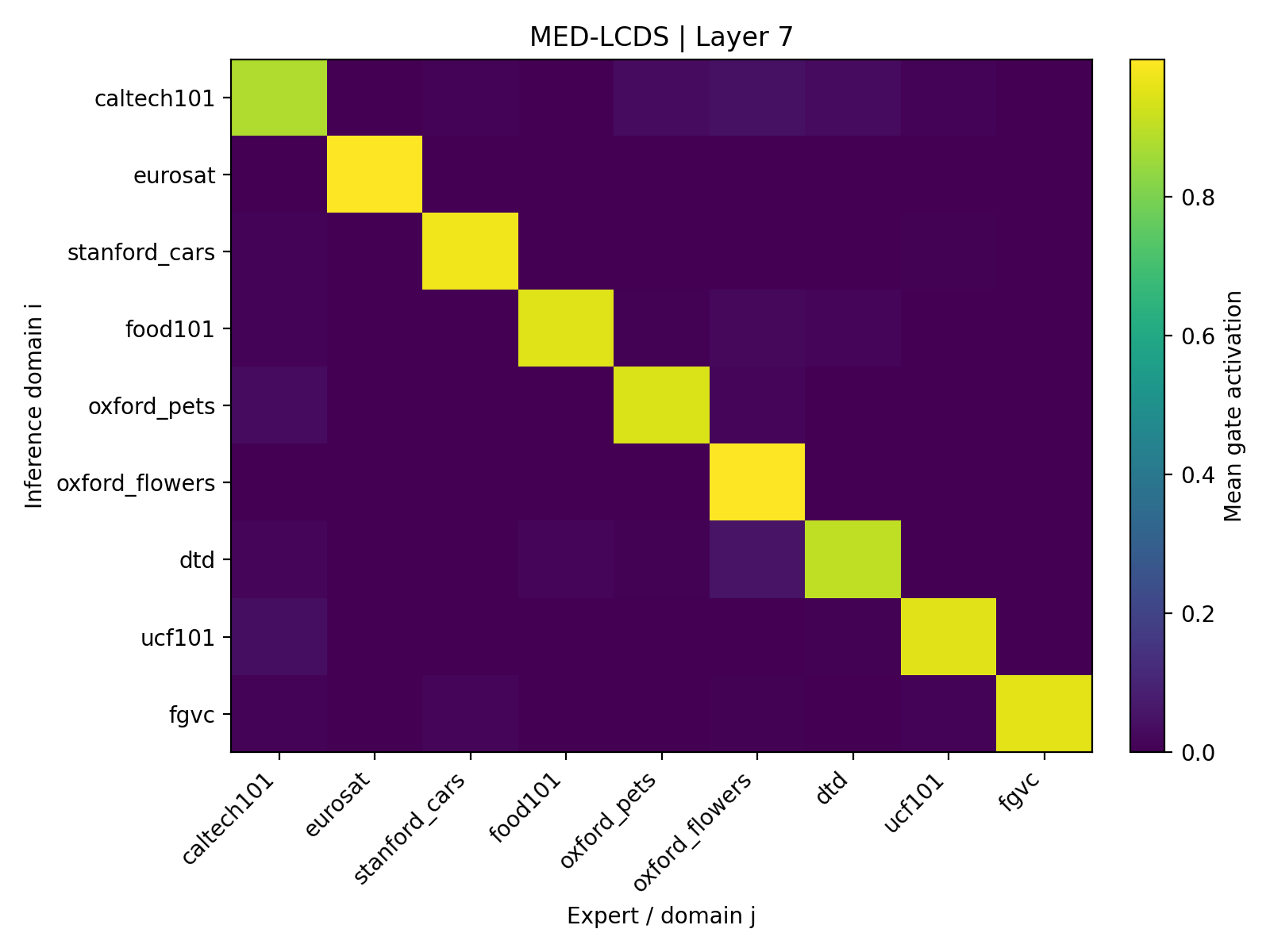}
        \caption{Layer 7 | \ours{}}
    \end{subfigure}

    \vspace{0.5em}

    \begin{subfigure}[b]{0.3\linewidth}
        \centering
        \includegraphics[width=\linewidth]{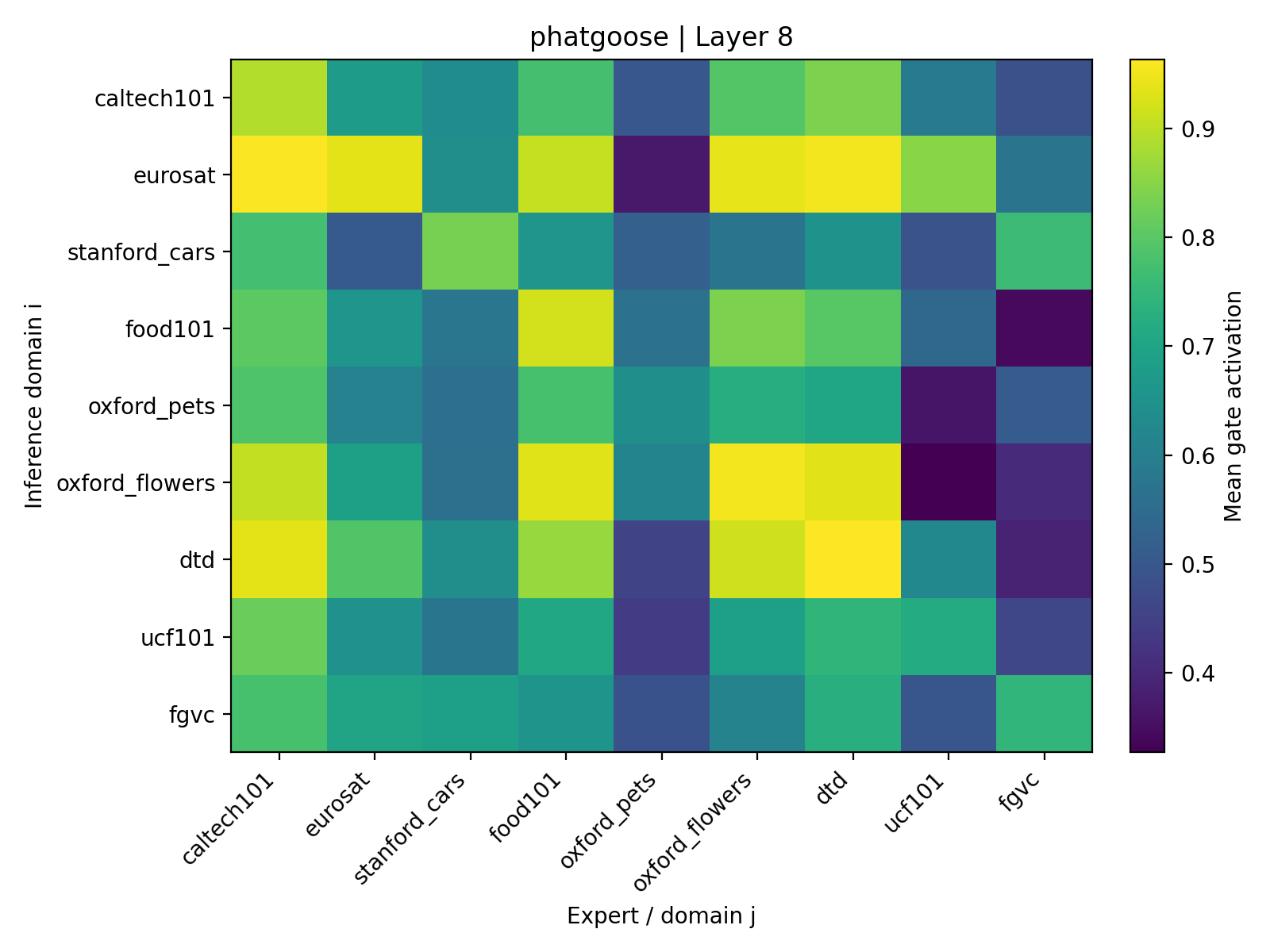}
        \caption{Layer 8 | Phatgoose}
    \end{subfigure}
    \hfill
    \begin{subfigure}[b]{0.3\linewidth}
        \centering
        \includegraphics[width=\linewidth]{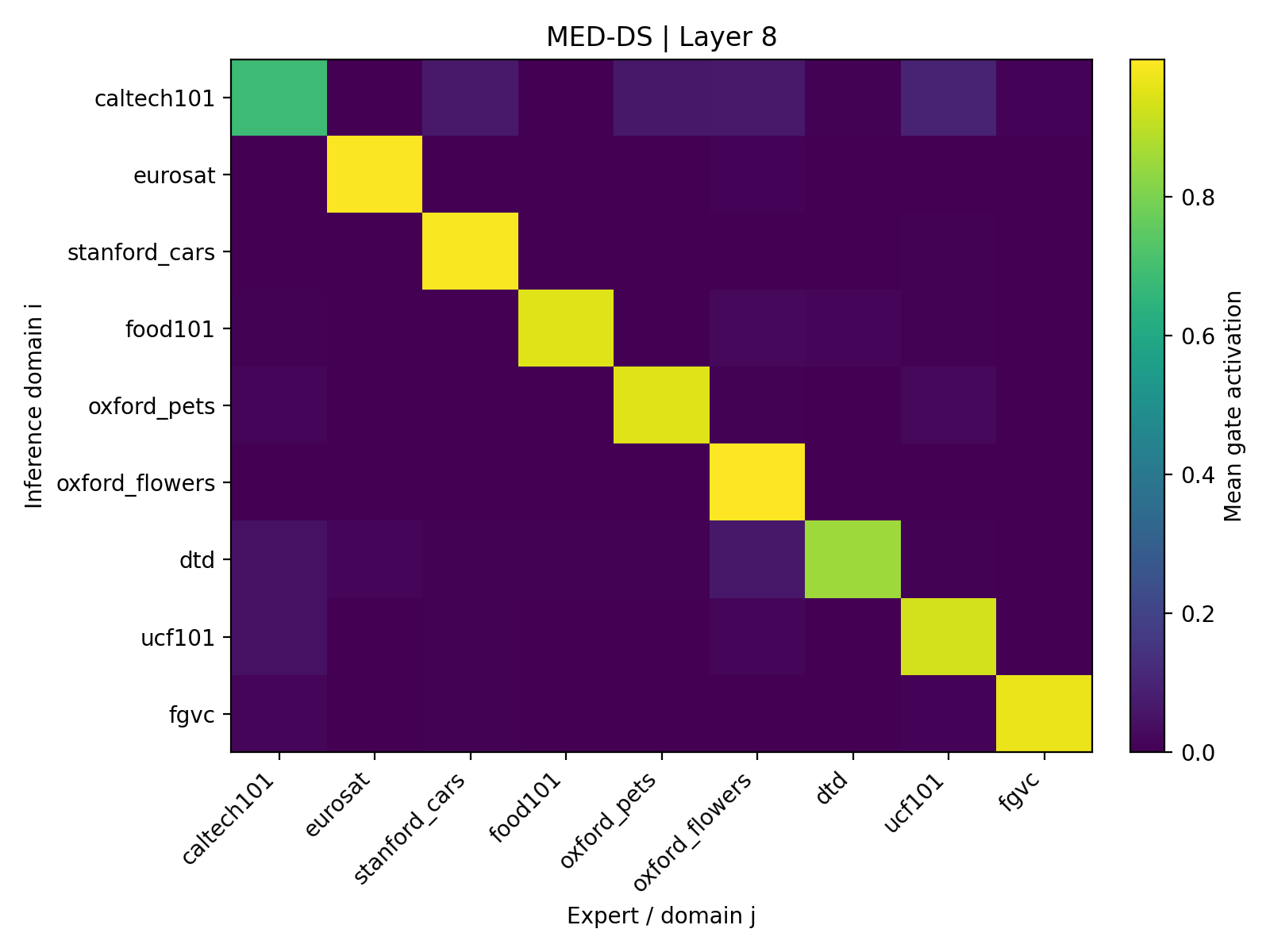}
        \caption{Layer 8 | \dsours{}}
    \end{subfigure}
    \hfill
    \begin{subfigure}[b]{0.3\linewidth}
        \centering
        \includegraphics[width=\linewidth]{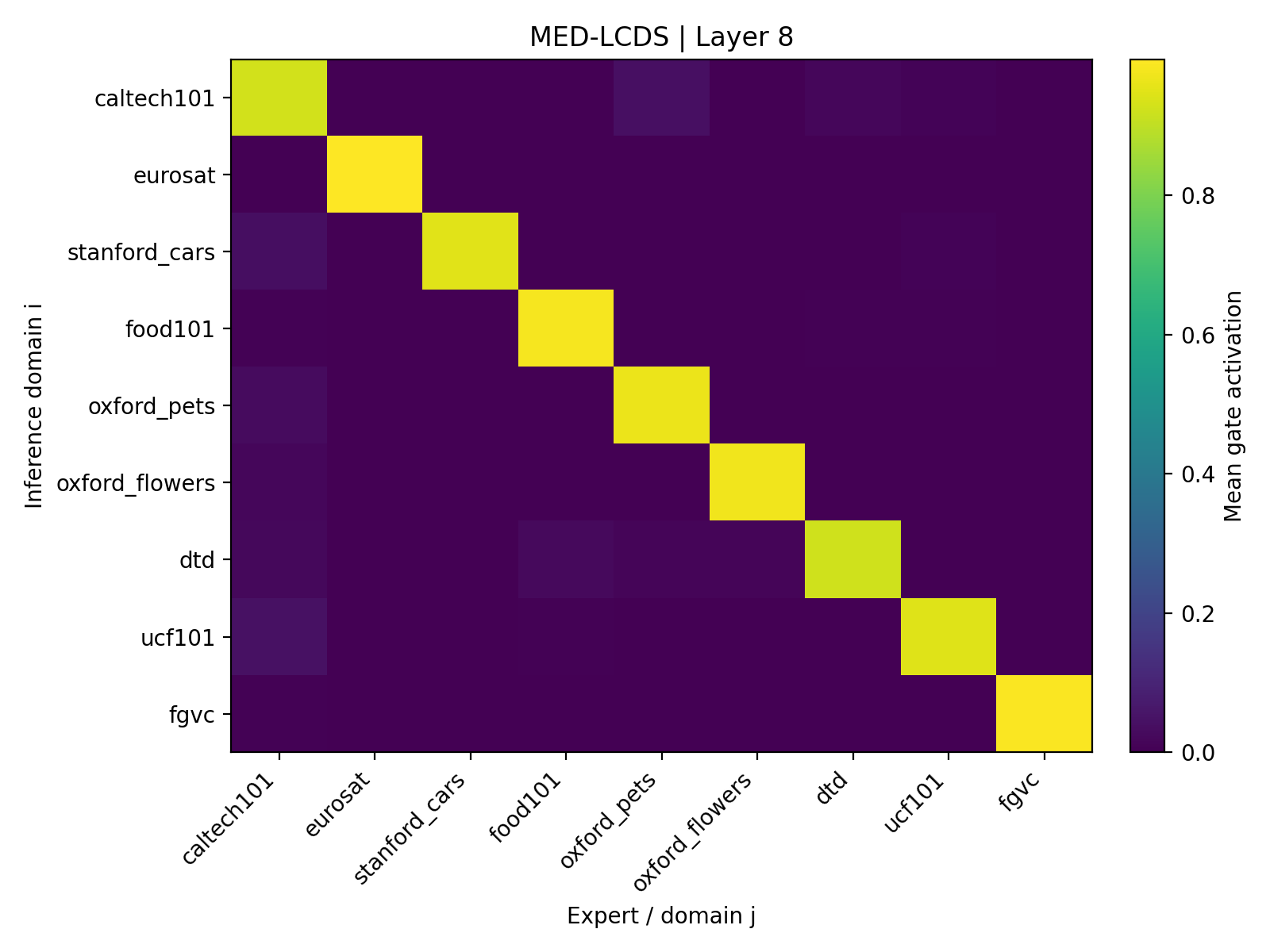}
        \caption{Layer 8 | \ours{}}
    \end{subfigure}

    \caption{Visualization of the gating at image encoder layers 5--8, where rows denote layers, columns denote methods, and color indicates the mean gating activation.}
    \label{fig:image_layers_5_8}
\end{figure*}

\begin{figure*}[htbp]
    \centering

    \begin{subfigure}[b]{0.3\linewidth}
        \centering
        \includegraphics[width=\linewidth]{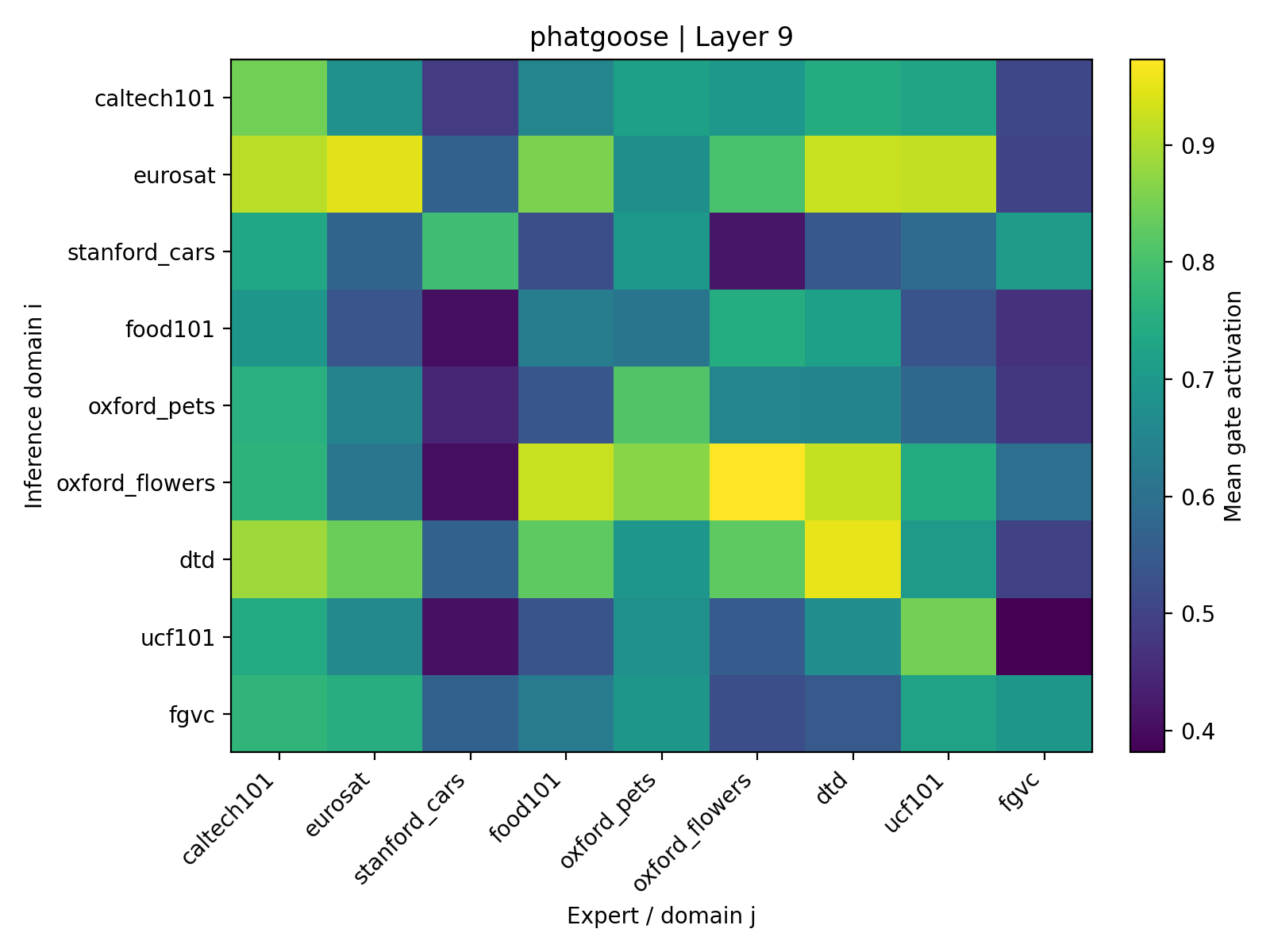}
        \caption{Layer 9 | Phatgoose}
    \end{subfigure}
    \hfill
    \begin{subfigure}[b]{0.3\linewidth}
        \centering
        \includegraphics[width=\linewidth]{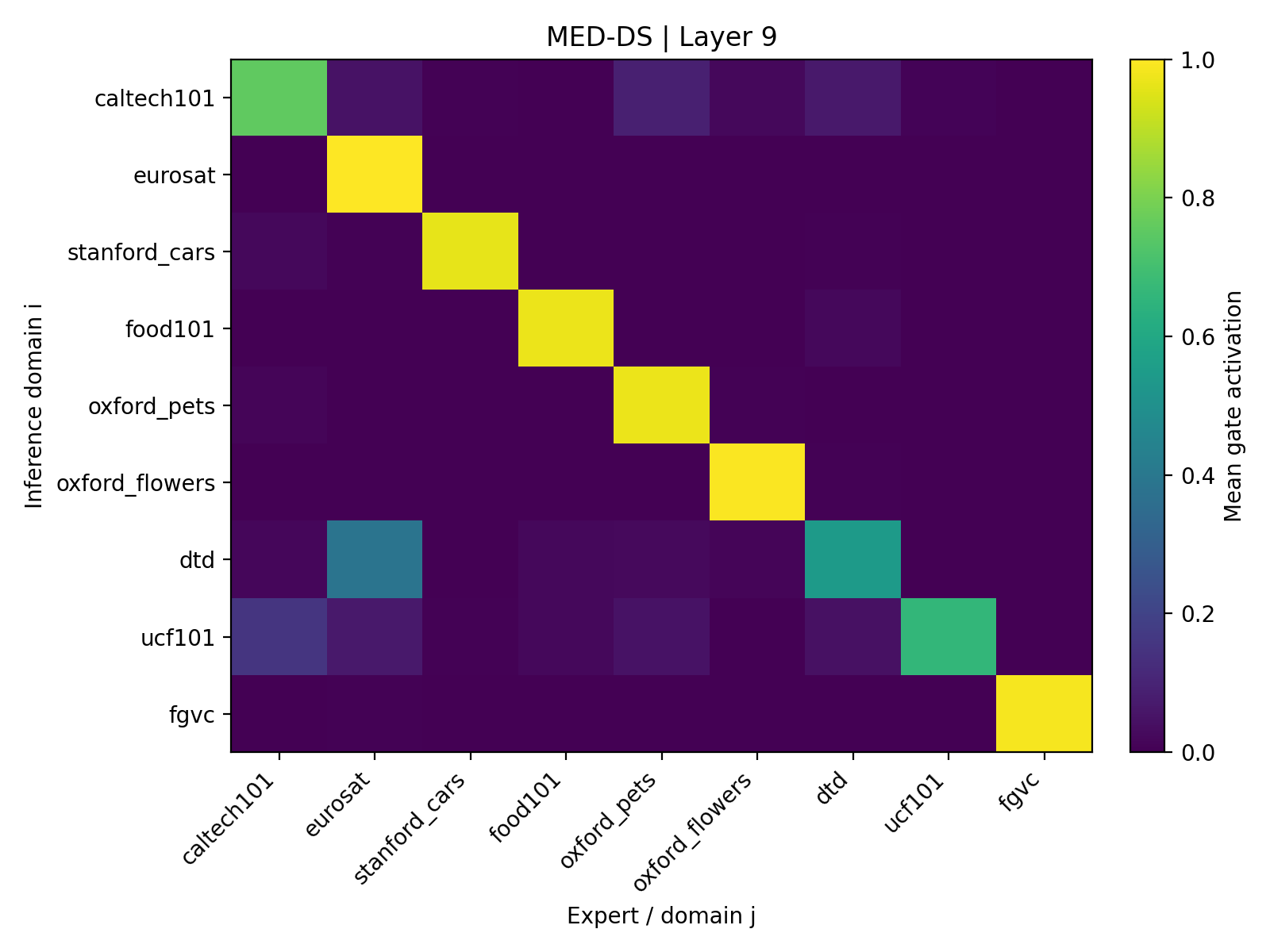}
        \caption{Layer 9 | \dsours{}}
    \end{subfigure}
    \hfill
    \begin{subfigure}[b]{0.3\linewidth}
        \centering
        \includegraphics[width=\linewidth]{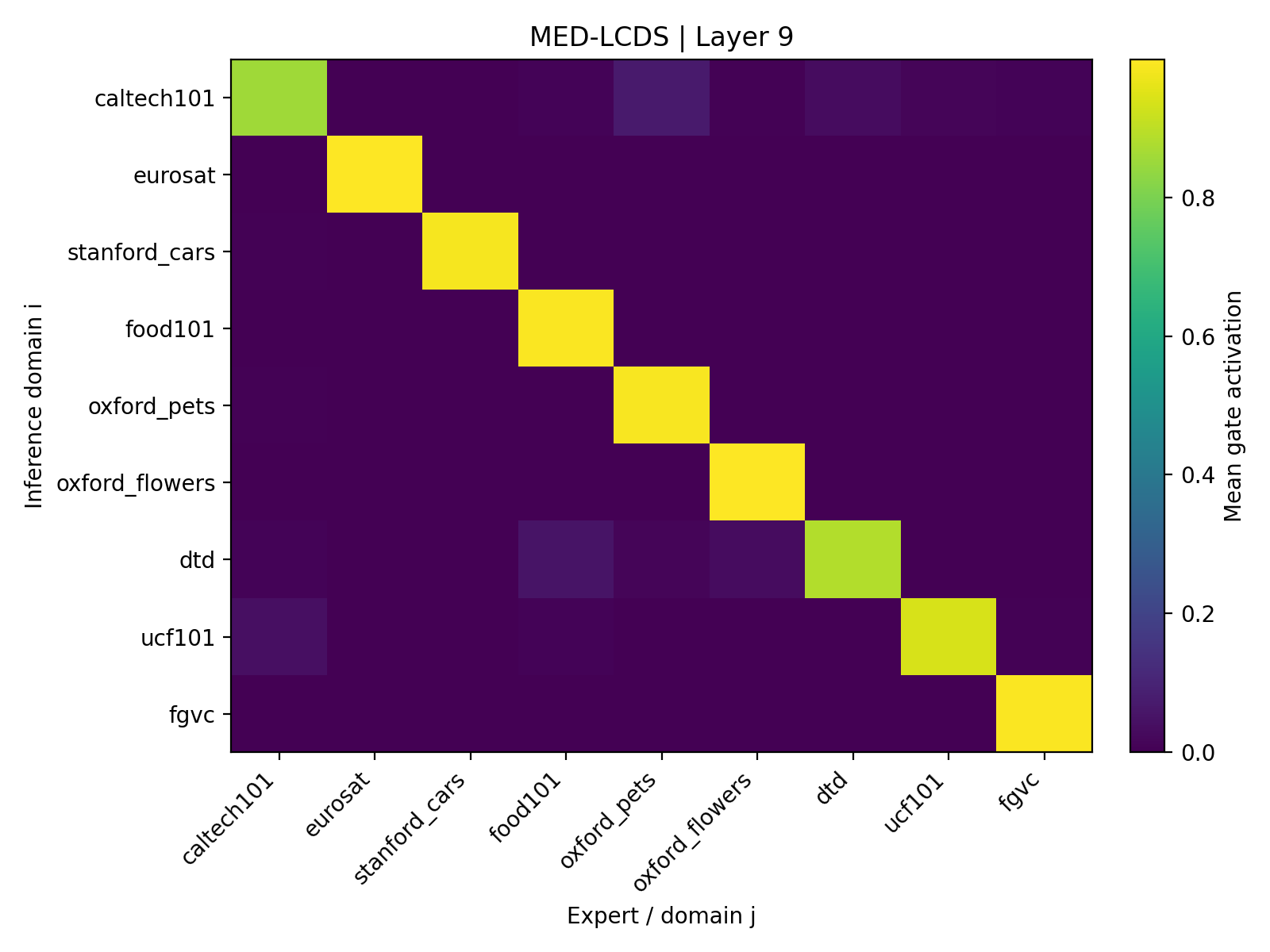}
        \caption{Layer 9 | \ours{}}
    \end{subfigure}

    \vspace{0.5em}

    \begin{subfigure}[b]{0.3\linewidth}
        \centering
        \includegraphics[width=\linewidth]{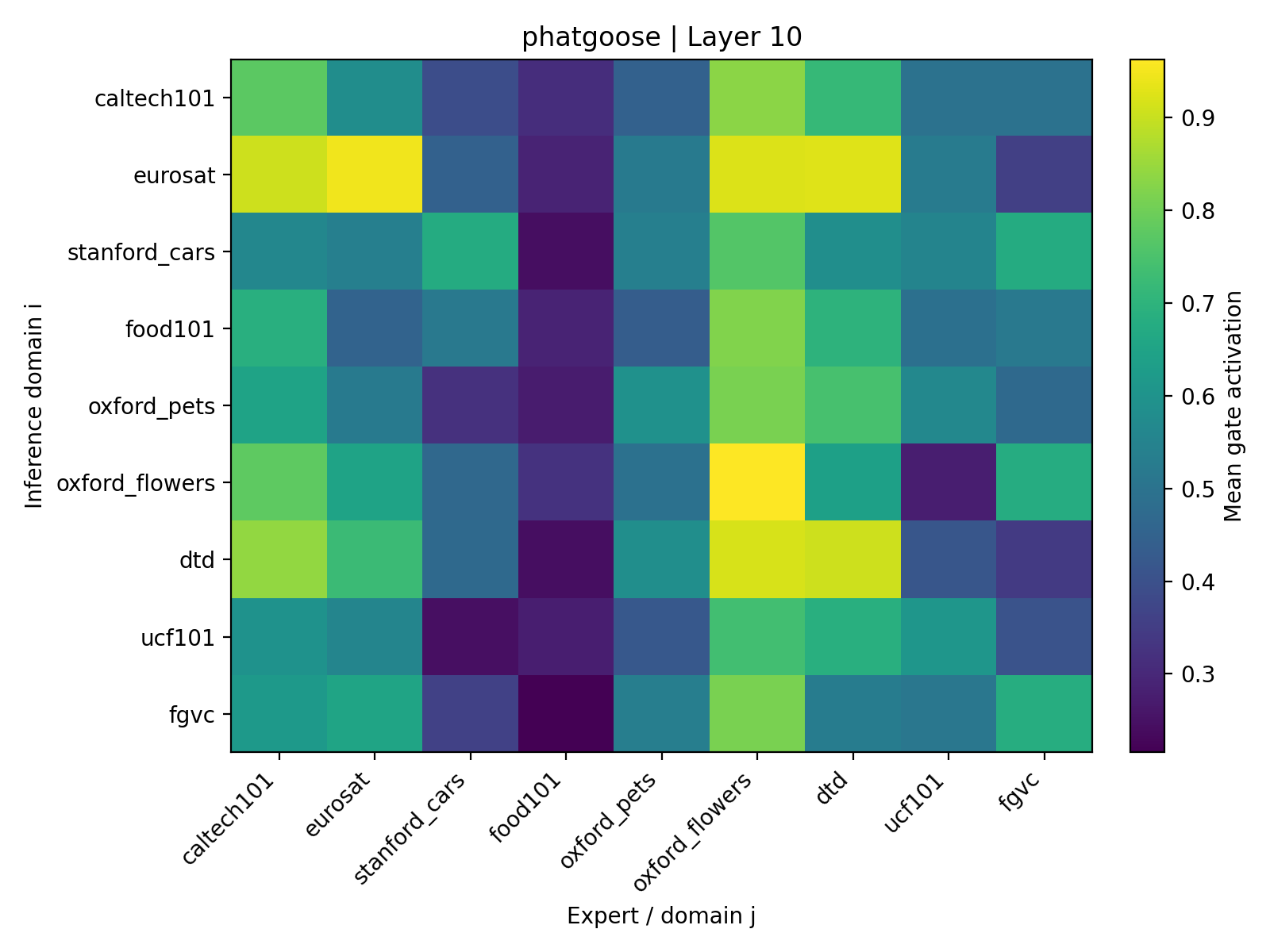}
        \caption{Layer 10 | Phatgoose}
    \end{subfigure}
    \hfill
    \begin{subfigure}[b]{0.3\linewidth}
        \centering
        \includegraphics[width=\linewidth]{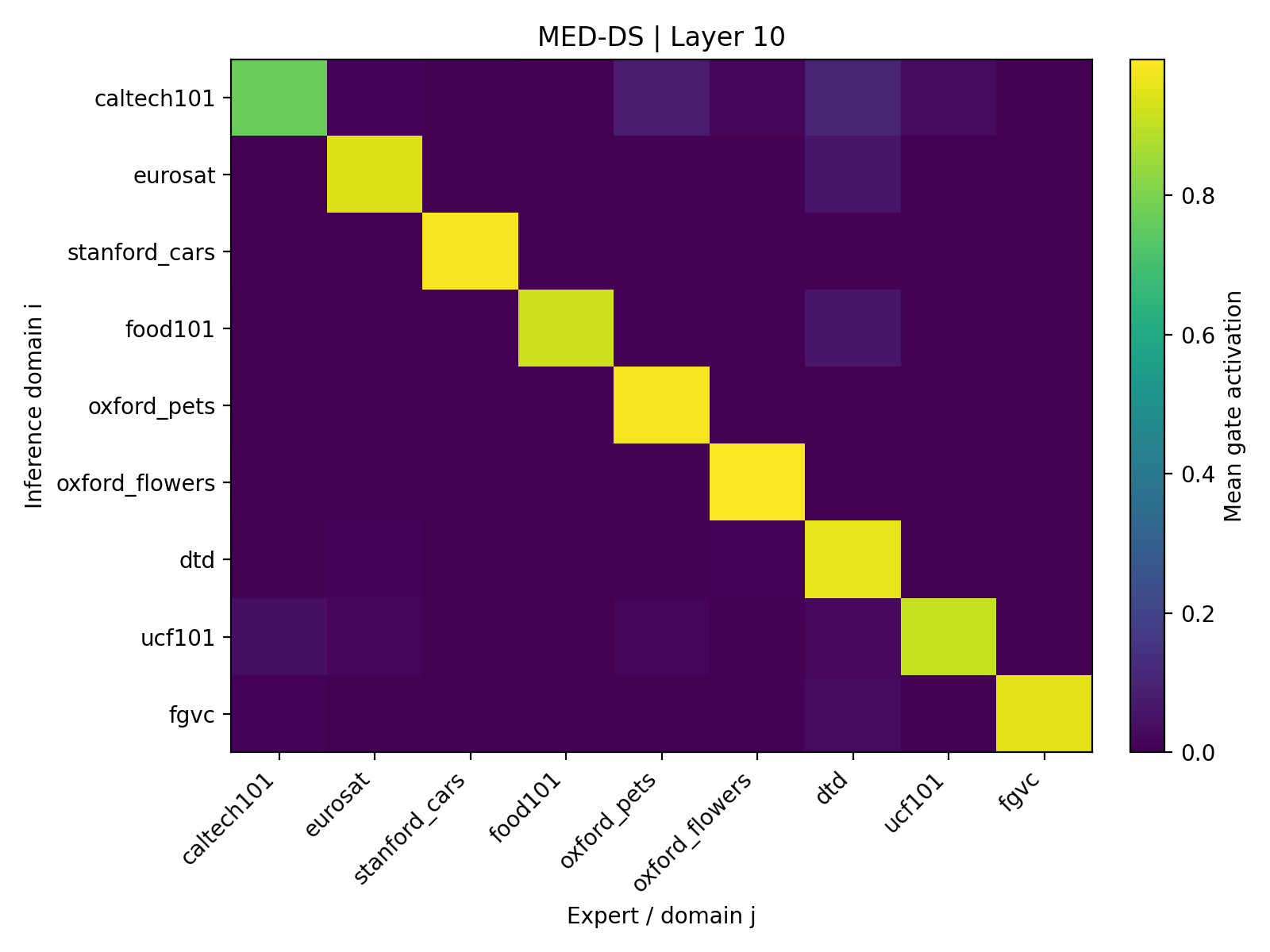}
        \caption{Layer 10 | \dsours{}}
    \end{subfigure}
    \hfill
    \begin{subfigure}[b]{0.3\linewidth}
        \centering
        \includegraphics[width=\linewidth]{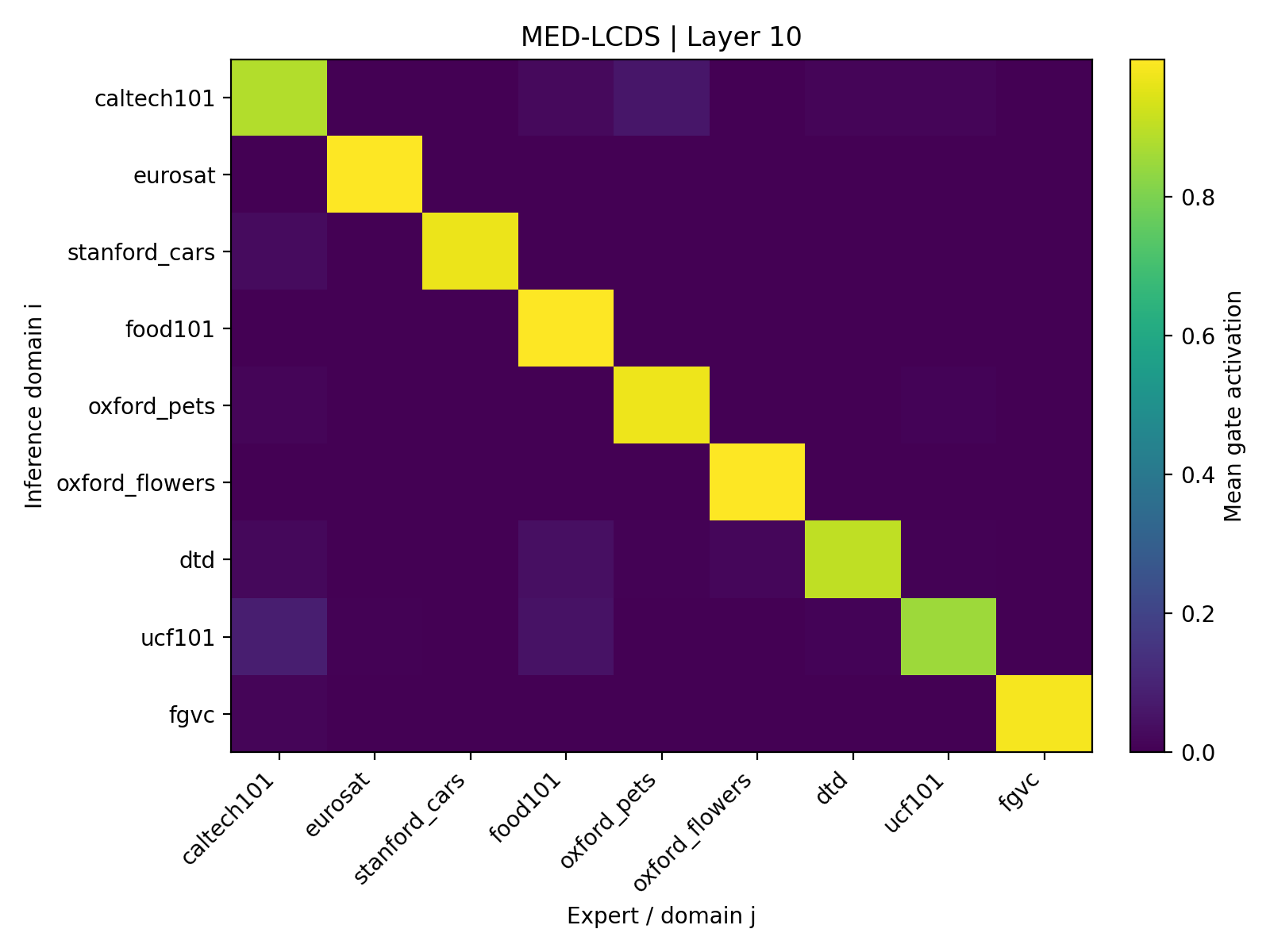}
        \caption{Layer 10 | \ours{}}
    \end{subfigure}

    \vspace{0.5em}

    \begin{subfigure}[b]{0.3\linewidth}
        \centering
        \includegraphics[width=\linewidth]{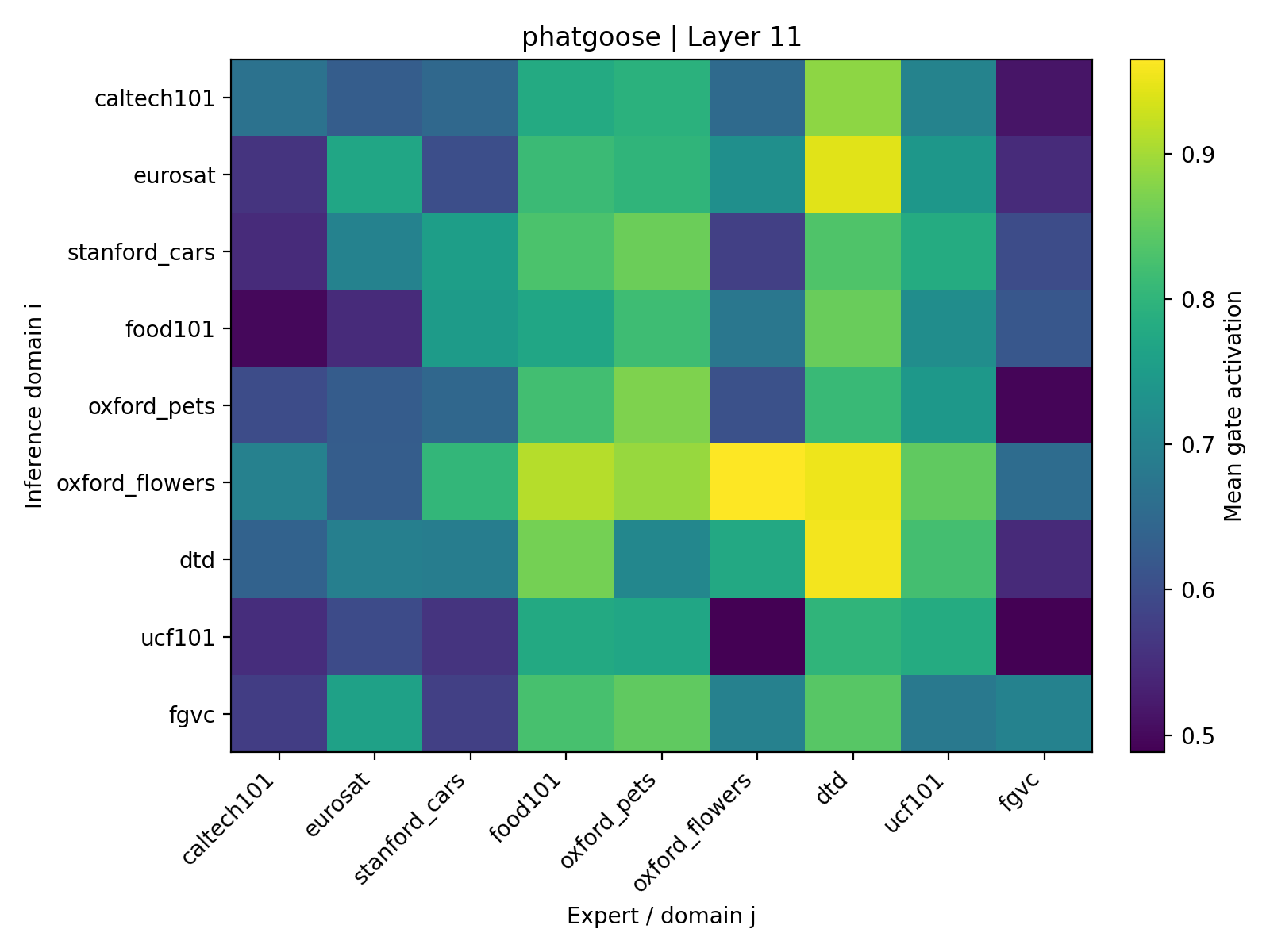}
        \caption{Layer 11 | Phatgoose}
    \end{subfigure}
    \hfill
    \begin{subfigure}[b]{0.3\linewidth}
        \centering
        \includegraphics[width=\linewidth]{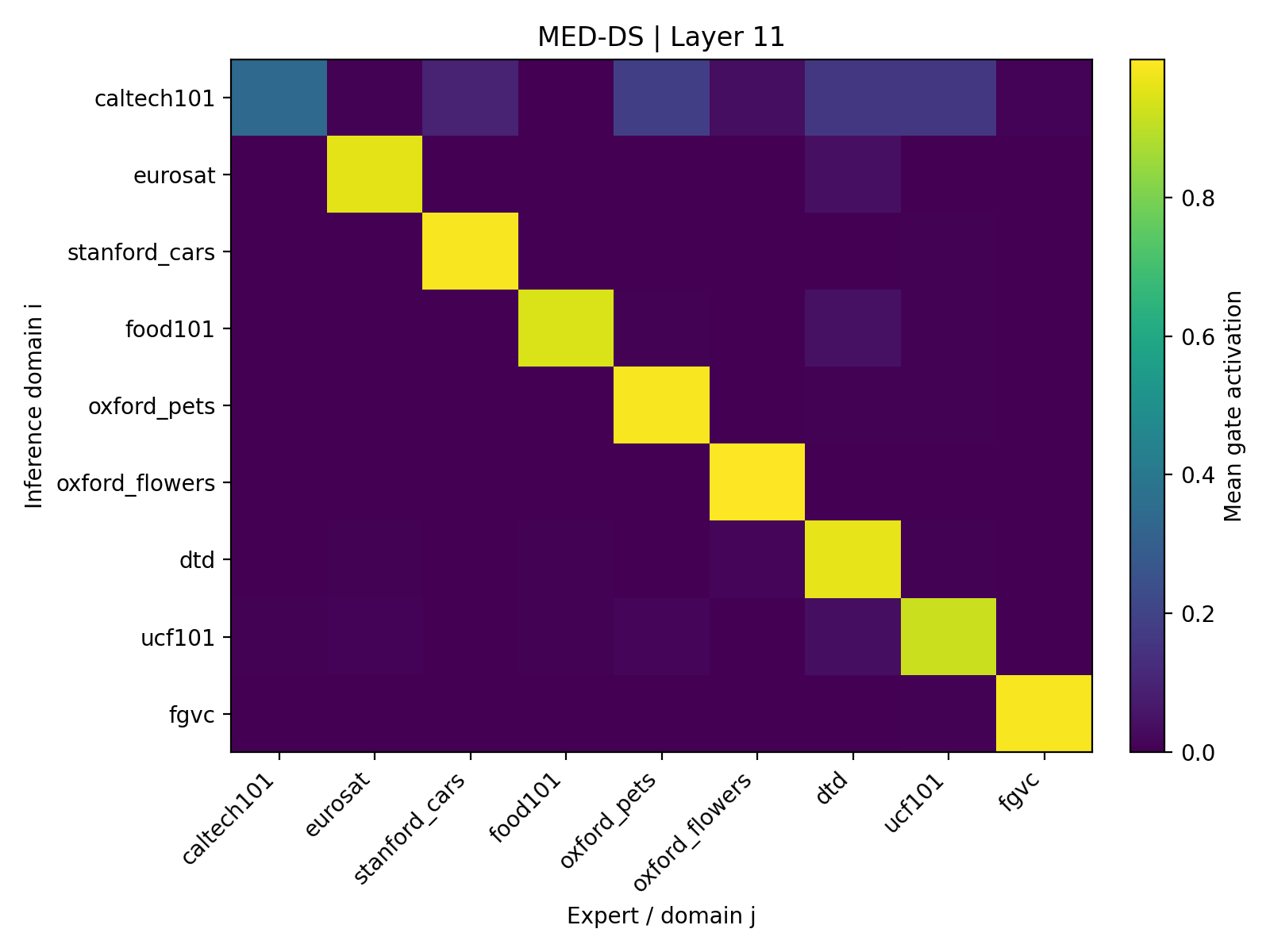}
        \caption{Layer 11 | \dsours{}}
    \end{subfigure}
    \hfill
    \begin{subfigure}[b]{0.3\linewidth}
        \centering
        \includegraphics[width=\linewidth]{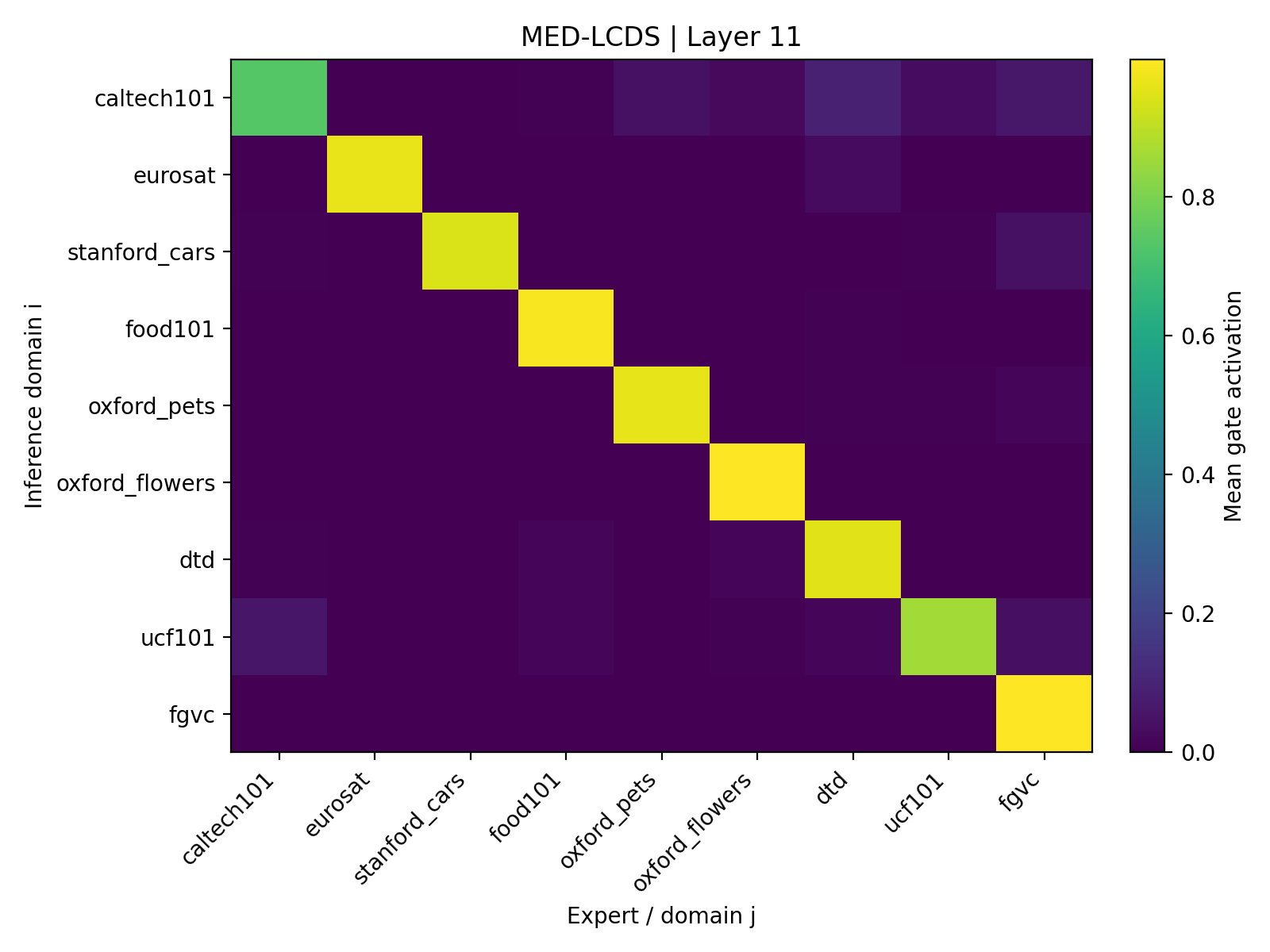}
        \caption{Layer 11 | \ours{}}
    \end{subfigure}

    \vspace{0.5em}

    \begin{subfigure}[b]{0.3\linewidth}
        \centering
        \includegraphics[width=\linewidth]{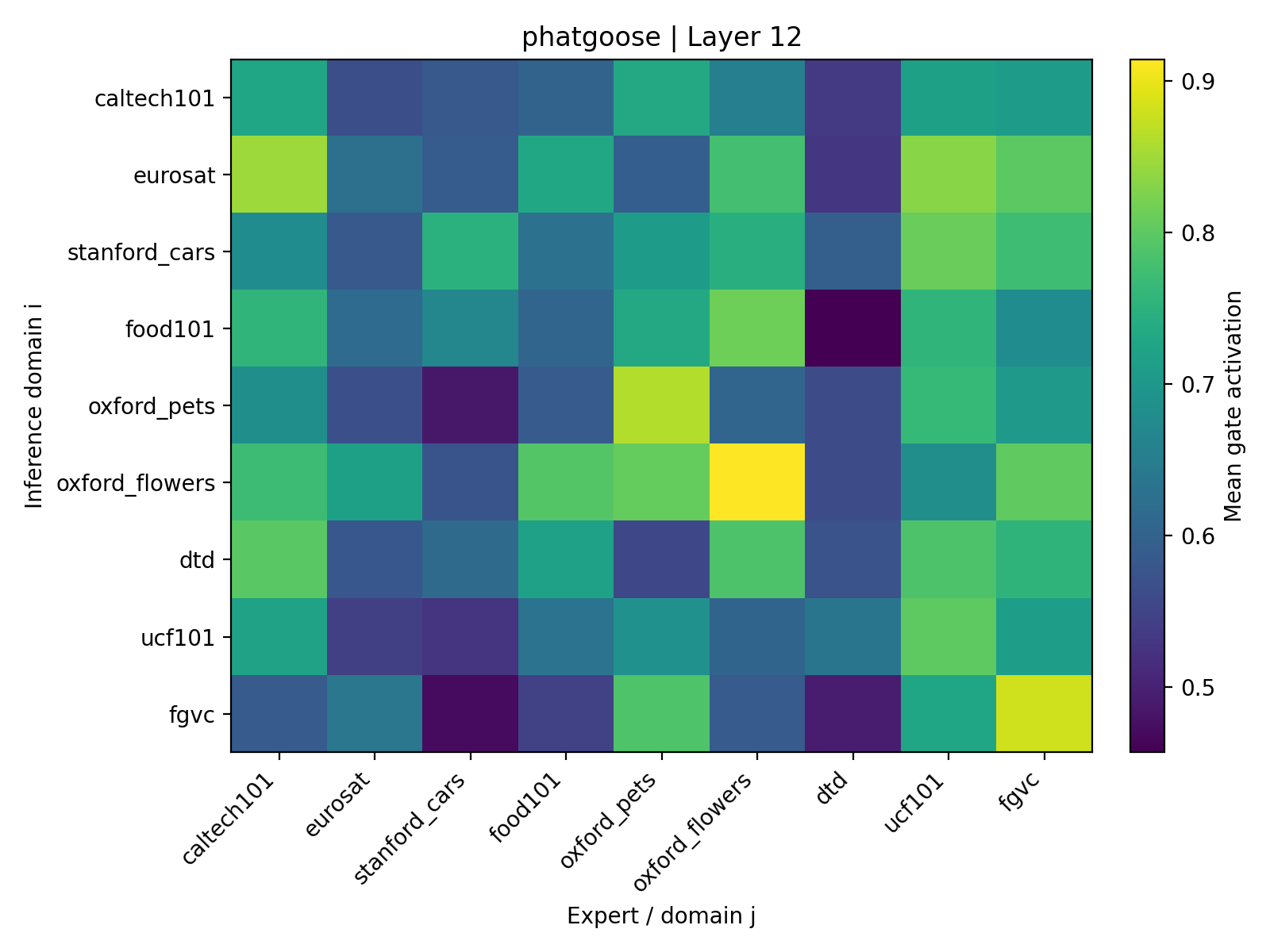}
        \caption{Layer 12 | Phatgoose}
    \end{subfigure}
    \hfill
    \begin{subfigure}[b]{0.3\linewidth}
        \centering
        \includegraphics[width=\linewidth]{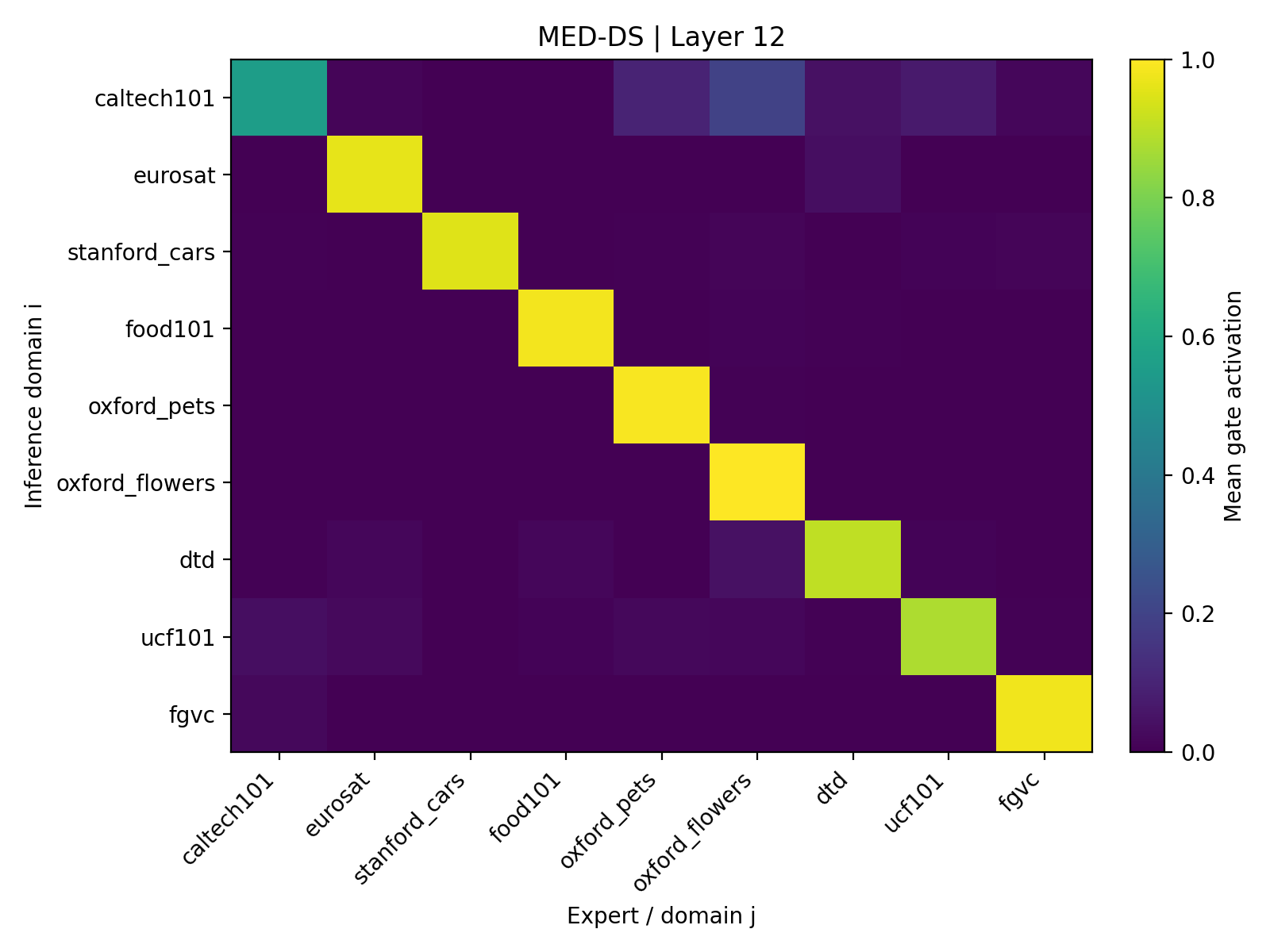}
        \caption{Layer 12 | \dsours{}}
    \end{subfigure}
    \hfill
    \begin{subfigure}[b]{0.3\linewidth}
        \centering
        \includegraphics[width=\linewidth]{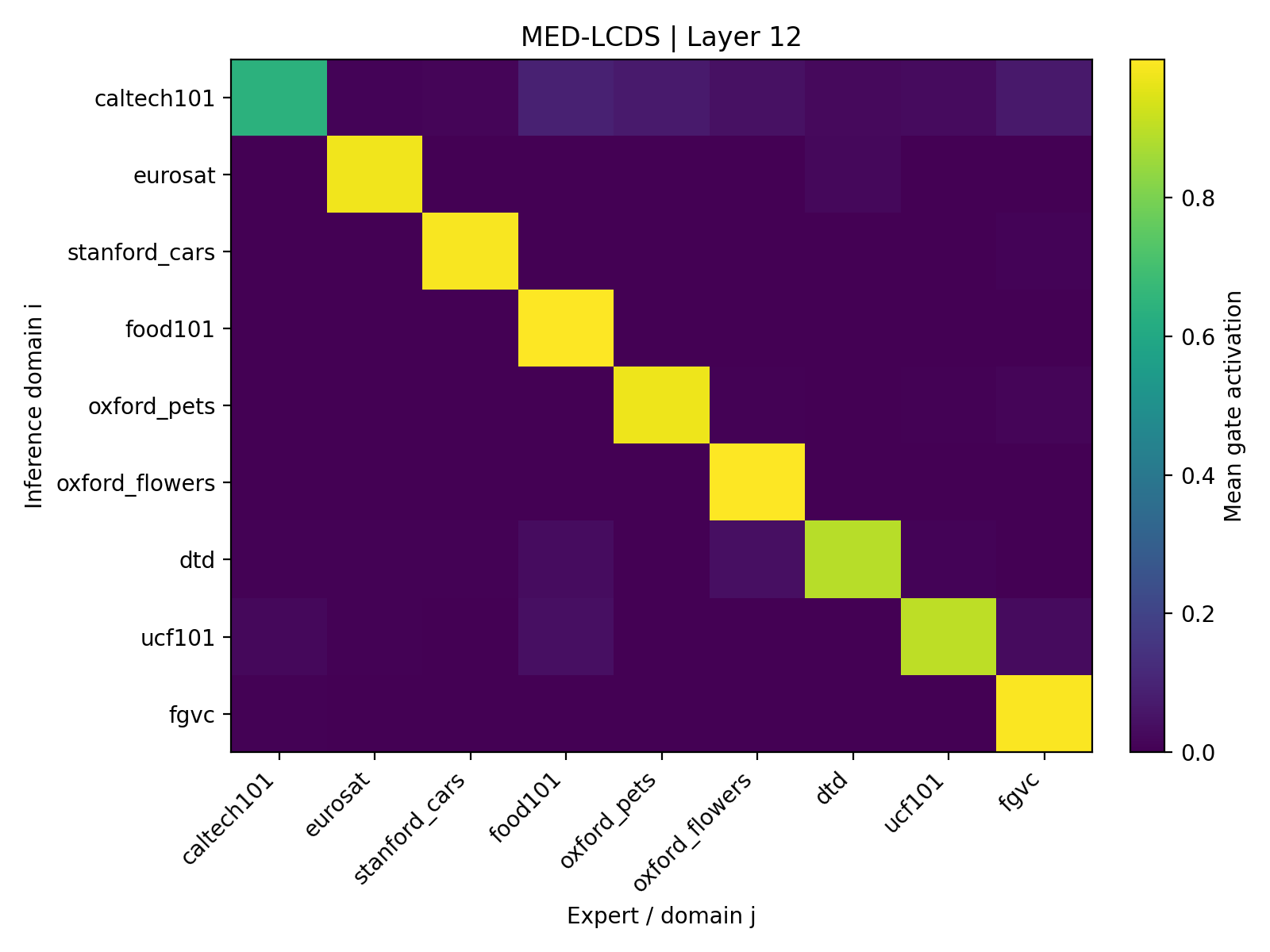}
        \caption{Layer 12 | \ours{}}
    \end{subfigure}

    \caption{Visualization of the gating at image encoder layers 9--12, where rows denote layers, columns denote methods, and color indicates the mean gating activation.}
    \label{fig:image_layers_9_12}
\end{figure*}

From Figure~\ref{fig:image_layers_1_4} to Figure~\ref{fig:image_layers_9_12}, we visualize the gating behavior of \ours{} using heatmaps. In these figures, rows correspond to the inference domain $i$ (i.e., the dataset being evaluated), while columns represent expert $j$. The color intensity indicates the mean gating activation, reflecting the degree to which each expert is selected for a particular domain. All heatmaps are generated by running the experiments on $\textit{all}$ split.

\textbf{\med vs. Phatgoose~\cite{muqeeth2024learning}.} As shown in Figures~\ref{fig:image_layers_1_4}--\ref{fig:image_layers_9_12}, Phatgoose~\cite{muqeeth2024learning} tends to activate multiple experts simultaneously during inference, whereas the \med family activates a smaller set of domain-relevant experts. This indicates that the routing signal in \med is more focused and selective due to the domain supervision, which also contributes to the improved performance reported in Table~\ref{tab:cross_domain_classes}.

\textbf{\ours{} vs. \dsours.} As illustrated in Figure~\ref{fig:image_layers_1_4}, \ours{} in the early layers consistently routes inputs to the appropriate domain expert, while \dsours occasionally activates experts that are not relevant to the current domain. For example, \dsours frequently selects the Oxford Flowers expert in layer 3, whereas \ours{} does not exhibit this behavior.

Overall, these results suggest that \ours{} produces more domain-consistent routing patterns, enabling clearer expert specialization and generalization leading to improved cross-domain performance even surpassing the Expert-LoRAs.

\section{Ablation Studies}
\paragraph{\bf \ours components.} Table 3 in the main paper showed the ablations of the components in \ours{} for cross-domain evaluation. Now, we present the in-domain ablation evaluation results.
Table~\ref{tab:ablation_in_domain_results} shows corresponding results for the in-domain setting. Again, the addition of domain supervision has a large gain and the softmax non-linearity clearly outperforms the sigmoid array. The only difference to the cross-domain results is that, for in domain evaluation, the logit scaling induces a small degradation of performance. This shows that logit scaling has a minor effect in terms of changing the relative magnitude of the logits for the classes of the true data domain. This gains of the cross-domain evaluation, which is the one that matters in practice where the true domain is unknown, show that this small loss is compensated by the larger gains of calibrating logits across domains. 

\end{document}